\documentclass[preprint]{article}

\PassOptionsToPackage{numbers, compress}{natbib}

\usepackage{neurips_2026}

\usepackage[utf8]{inputenc} 
\usepackage[T1]{fontenc}    
\usepackage{hyperref}       
\usepackage{url}            
\usepackage{booktabs}       
\usepackage{amsfonts}       
\usepackage{nicefrac}       
\usepackage{microtype}      
\usepackage{xcolor}         

\usepackage{microtype}
\usepackage{graphicx}
\usepackage{booktabs}

\usepackage{amsmath}
\usepackage{amssymb}
\usepackage{mathtools}
\usepackage{needspace}
\usepackage{fancyvrb}
\usepackage[capitalize,noabbrev]{cleveref}
\usepackage{url}
\usepackage{hyperref}
\usepackage{mathtools}
\usepackage{float}
\usepackage{placeins}
\usepackage{titletoc}
\usepackage{amsfonts}
\usepackage{algorithm}
\usepackage{algorithmicx}
\usepackage[noend]{algpseudocode}
\usepackage{tikz}
\usetikzlibrary{
  arrows.meta,
  positioning,
  calc,
  shapes.geometric,
  shapes.symbols
}

\usepackage{textcomp}
\usepackage{ dsfont } 
\usepackage{nicefrac}      
\usepackage{bbm}
\usepackage{color}
\usepackage{enumerate}
\usepackage{enumitem}
\usepackage{stmaryrd}
\usepackage{subcaption}
\usepackage{xspace}
\usepackage{comment}
\usepackage{xcolor}
\usepackage{multirow}
\usepackage{array}
\usepackage{titletoc}
\usepackage{wrapfig}
\usepackage{caption}

\usepackage{afterpage}
\usepackage{dblfloatfix}

\renewcommand{\arraystretch}{1.2}

\newtheorem{theorem}{Theorem}[section]

 \newtheorem{definition}[theorem]{Definition}

\newcommand{\code}[1]{{\texttt {#1}}}

\renewcommand{\P}{\mathcal{P}}

\newcommand{\M}{\mathcal{M}}

\newcommand{\p}{\phi}

\newcommand{\traj}{\mathcal{Z}}

\newcommand{\choice}[2]{#1 \; \code{or} \; #2}
\newcommand{\reach}[1]{\code{reach} \; #1}
\newcommand{\avoid}[1]{\code{avoid} \; #1}

\newcommand{\semantics}[1]{{\llbracket #1 \rrbracket}}

\newcommand{\eventually}[1]{\code{achieve} \; #1}
\newcommand{\always}[1]{~ \code{ensuring} \; #1}
\newcommand{\true}{\code{true}}
\newcommand{\false}{\code{false}}

\newcommand{\task}{\mathsf{R}}

\newcommand{\drill}{\textsf{DIBS}}

\newcommand{\Train}{\mathsf{Train}}
\newcommand{\Unseen}{\mathsf{Unseen}}

\newcommand{\spectrl}{\textsc{Spectrl}\xspace}

\newcommand{\genrl}{\mbox{\textnormal{\textsf{GenRL}}}}

\newcommand{\hp}[1]{%
  \ifnum#1>0
    \hphantom{\_}%
    \hp{\the\numexpr#1-1\relax}%
  \fi
}

\newcommand{\Goal}{\mathcal{G}}     
\newcommand{\Safe}{\mathcal{S}}     

\newcommand{\taskfam}{\mathcal{R}}
\newcommand{\spec}{\varphi}

\newcommand{\Init}{\mathsf{Init}}

\newcounter{rowfig}

\makeatletter
\renewcommand\p@subfigure{\thefigure.}
\makeatother

\usepackage{amsmath,bm}

\def\eqref#1{equation~\ref{#1}}

\def\1{\bm{1}}

\DeclareMathAlphabet{\mathsfit}{\encodingdefault}{\sfdefault}{m}{sl}
\SetMathAlphabet{\mathsfit}{bold}{\encodingdefault}{\sfdefault}{bx}{n}

\newcommand{\R}{\mathbb{R}}

\title{Decoupled Behavioral Cloning for Scalable Inductive Generalization in RL from Specifications}

\author{%
Vignesh Subramanian$^{1}$,
Subhajit Roy$^{2}$,
Suguman Bansal$^{1}$\\[0.5em]
$^{1}$School of Computer Science, Georgia Institute of Technology, USA\\
$^{2}$Department of Computer Science and Engineering, Indian Institute of Technology Kanpur, India\\[0.5em]
$^{1}$\texttt{\{vignesh,suguman\}@gatech.edu},
$^{2}$\texttt{subhajit@cse.iitk.ac.in}
}

\begin{document}
\maketitle

\begin{abstract}
Inductive generalization is a framework for reinforcement learning (RL)
generalization in which inductively related task instances admit inductively
related policies. Prior work captures this structure via a higher-order
\emph{policy-evolution function} learned directly with RL, but suffers from
poor \emph{training scalability}: as training tasks grow, aggregated reward
feedback becomes noisy and conflicting, destabilizing training and weakening
generalization. We propose \drill, a \emph{decoupled behavioral cloning}
approach that separates learning task-specific policies from learning the
evolution function. We first learn individual teacher policies per task via
standard RL, then fit the evolution function via behavioral cloning on
teacher-labeled state-action pairs. This replaces noisy reward aggregation
with dense, stable supervision. \drill\ achieves significant improvements in both training stability and zero-shot generalization against existing RL and meta-RL algorithms.

\end{abstract}

\section{Introduction}
\label{sec:intro}

The \textit{generalization} problem in \textit{reinforcement learning (RL)}~%
\cite{sutton1998reinforcement} asks
how agents can learn policies that transfer to new, unseen situations~%
\cite{KirkZGR23, cobbe2019quantifying, zhang2018study, cobbe2020leveraging}.
Despite remarkable successes, generalization remains a fundamental challenge.
RL agents that perform well on individual tasks often fail on even slightly
modified scenarios.

\textit{Inductive generalization} (\genrl)~\cite{subramanian2025inductive} is a
notion of generalization that leverages
\textit{inductive similarities} between tasks. The core intuition is that
inductively related tasks must have inductively related policies. 
 Inductive relationships arise naturally
whenever tasks exhibit recursion or iteration: once we understand how to
transform the policy for the $i-th$ task to the $i+1$ task by training on a few tasks, we can generalize to
arbitrarily many tasks. This structure is evident in robotics
and control, where physical constraints impose regular, repeating patterns.
Prior work formalizes this intuition by encoding {\em inductive families of tasks} using \emph{temporal logic
specifications}, where tasks share
identical logical structure and differ only in the instantiation of
low-level predicates.

\begin{figure*}[t]
    \centering

    \begin{minipage}[t]{0.30\textwidth}
        \centering

        \begin{subfigure}[t]{\linewidth}
            \centering
            \includegraphics[width=0.95\linewidth, trim={0 0 0 0.4in}, clip]{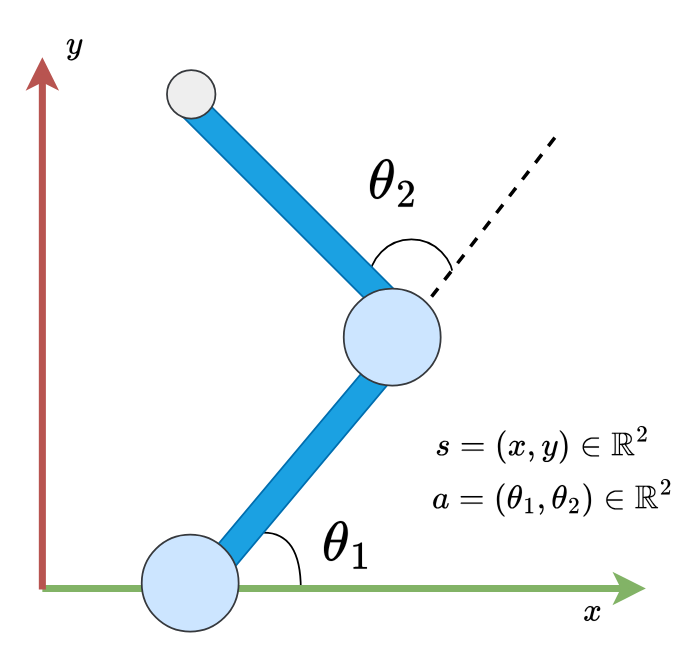}
            \caption{Reacher benchmark environment}
            \label{fig:reacherdyn}
        \end{subfigure}

        \vspace{1em}

        \begin{subfigure}[t]{\linewidth}
            \centering
            \includegraphics[width=0.95\linewidth, trim={0 0 0 0.1in}, clip]{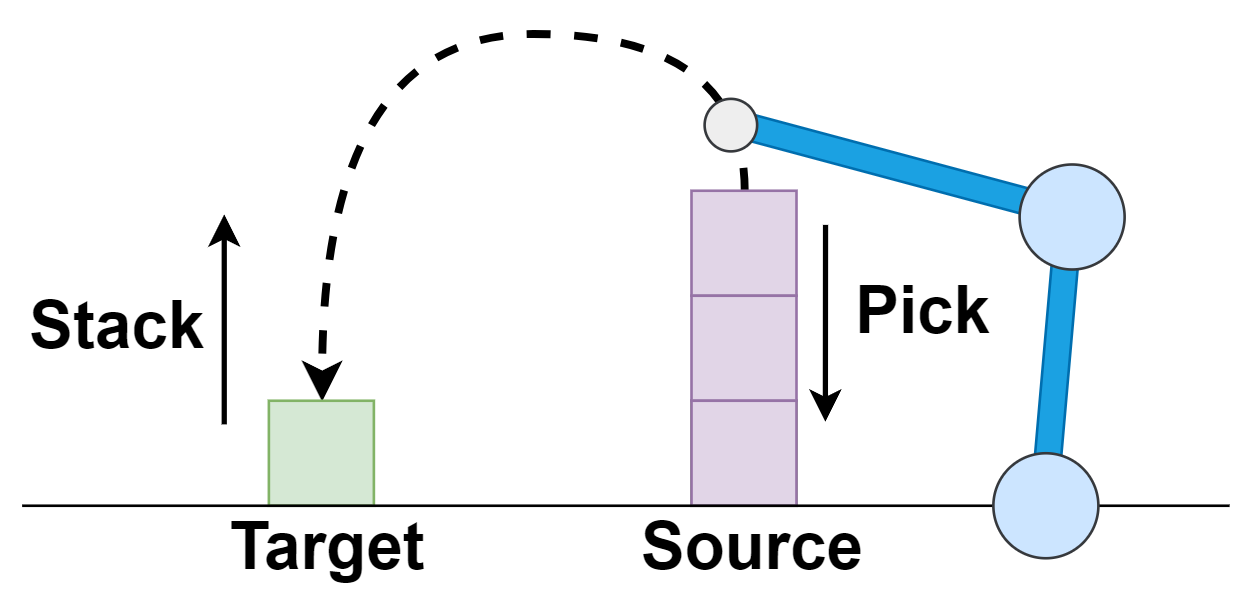}
            \caption{Inductive task family in the Reacher benchmark}
            \label{fig:r4illus}
        \end{subfigure}
    \end{minipage}
    \hfill
    \begin{minipage}[t]{0.66\textwidth}
        \centering

        \begin{subfigure}[t]{\linewidth}
            \centering
            \includegraphics[width=\linewidth]{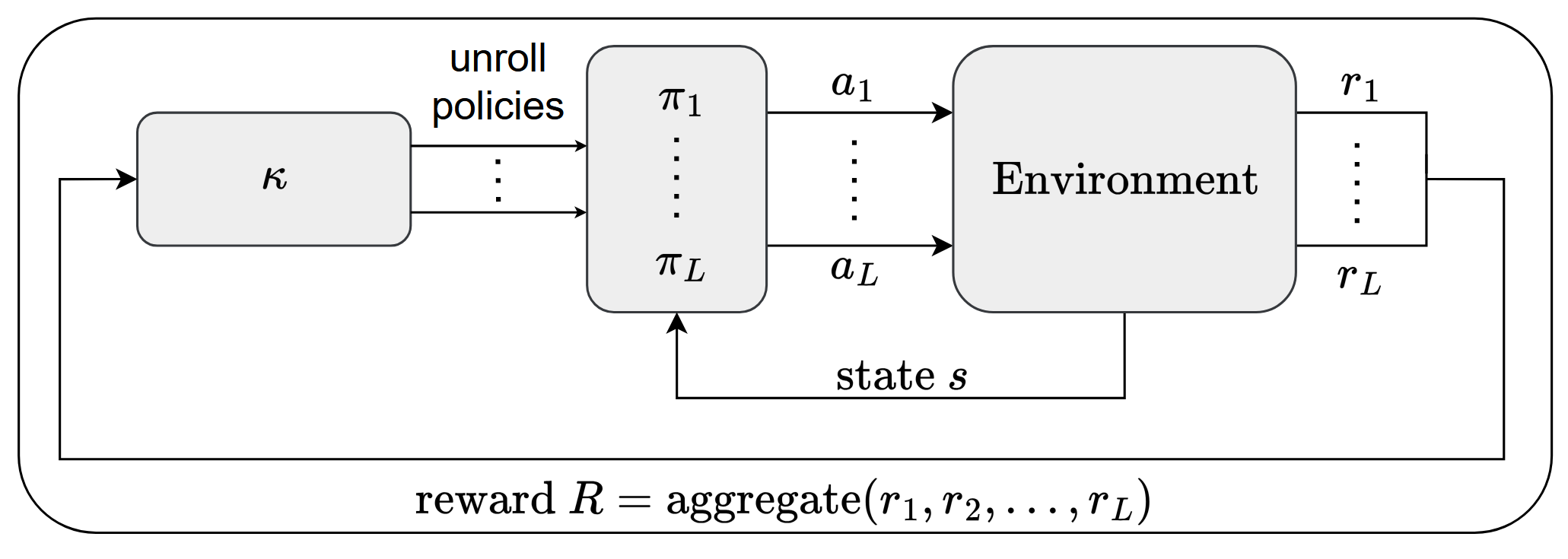}
            \caption{\genrl~flow diagram}
            \label{fig:flow-diagram-genrl}
        \end{subfigure}

        \vspace{1em}

        \begin{subfigure}[t]{\linewidth}
            \centering
            \includegraphics[width=\linewidth]{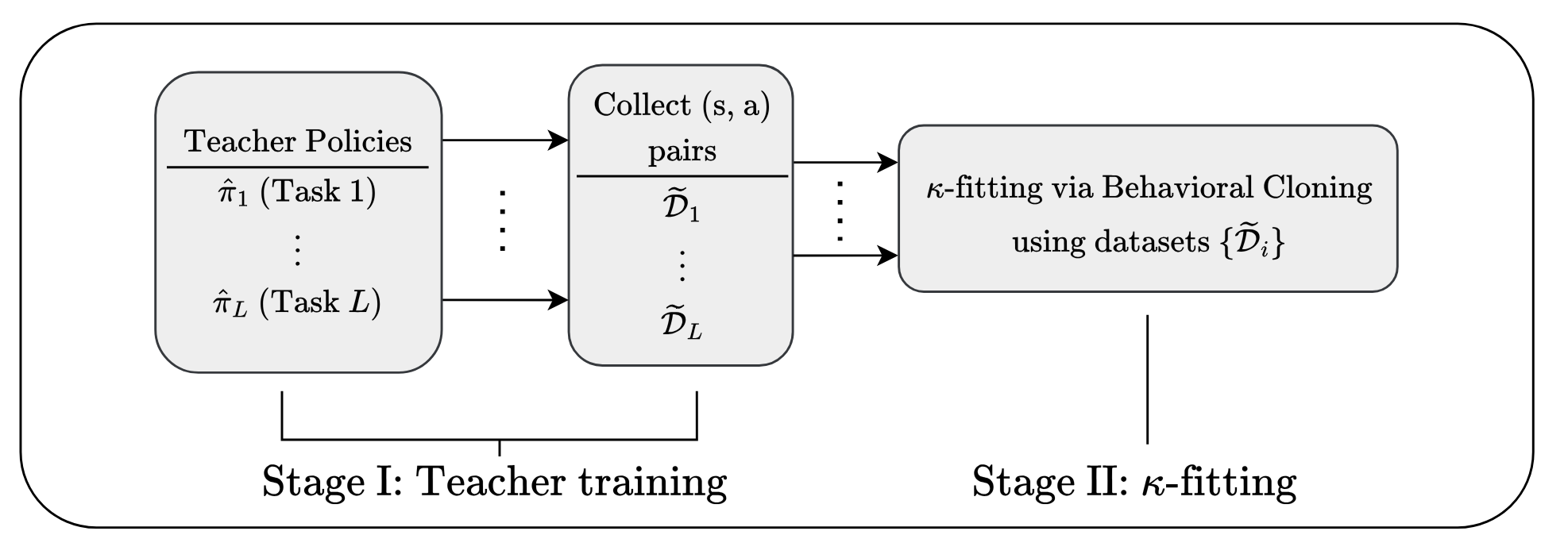}
            \caption{\drill~flow diagram}
            \label{fig:flow-diagram-ours}
        \end{subfigure}
    \end{minipage}

    \caption{
    \textbf{Left column: Benchmark illustrations.}
    \textbf{(a)} Reacher benchmark environment.
    \textbf{(b)} Inductive task family in the Reacher benchmark.
    \textbf{Right column: Contrasting flow diagrams.}
    \textbf{(c)} \genrl~flow diagram.
    \textbf{(d)} \drill~flow diagram.
    }
    \label{fig:combined_flow_reacher}
\end{figure*}

Consider the robotic arm in Fig.~\ref{fig:combined_flow_reacher}. It has
to displace a tower of blocks from a source to a target. This complex task
decomposes into a series of inductively related subtasks. In the $i$-th
subtask, the robot relocates the topmost block from the source at height $i$
to the  top of the target  at height $(h - i)$, where $h$
is the initial height of the source. The $0$-th task is
the base task, and successive tasks are obtained by updates to the 
locations of the topmost blocks in the source and target.

The inductive generalization algorithm
\genrl~captures this structure by learning
a higher-order \textit{policy-evolution function} $\Omega: \Pi \to \Pi$
such that $\pi_{i+1} = \Omega(\pi_i)$, approximated as an $m$-degree
polynomial over policy parameters. Learning reduces to inferring a small
set of $\kappa$-coefficients via an RL-style training loop over training
tasks (Fig.~\ref{fig:flow-diagram-genrl}). Despite stronger
generalization than several state-of-the-art methods, \genrl~suffers from a
fundamental \textit{training scalability} bottleneck: policy learning for
training tasks is tightly coupled with $\kappa$-coefficient training. As
the number of training tasks grows, aggregated reward feedback becomes
noisy and conflicting, making the training loop increasingly difficult to
stabilize, ultimately constraining its generalization
capability.

We propose \drill, a \emph{decoupled behavioral cloning} approach that
resolves this bottleneck by separating two phases that \genrl~conflates
(Fig.~\ref{fig:flow-diagram-ours}). In \textbf{Stage I}, we independently
learn \emph{teacher policies} for each training task using standard RL.
In \textbf{Stage II}, we fit the $\kappa$-template via \emph{behavioral
cloning} on teacher-labeled state-action
datasets~\cite{torabi2018behavioral, levine2016end, osa2018algorithmic},
replacing noisy multi-task reward aggregation with stable, dense
supervision. Since teacher policies are fixed during $\kappa$-training,
the destabilizing feedback loop in \genrl~is eliminated. A further
advantage is that teacher policies can be learned using \textit{any} RL
algorithm, unlike \genrl, which is tightly coupled to
ARS~\cite{mania2018simple}, fundamentally limiting the environments it
can handle.
This decoupling raises non-trivial challenges. One of them is that  independently
trained teachers can drift arbitrarily in parameter space across task indices,
making it difficult for a single $\kappa$ to fit all of them simultaneously. 
We address this through cross-index regularization during teacher training. In summary, our contributions are:

\begin{itemize}[leftmargin=10pt]
    \item \textbf{Algorithm.} We propose \drill, a decoupled behavioral
    cloning approach to inductive generalization that replaces \genrl's
    unstable multi-task RL loop with stable supervised learning
    (Section~\ref{sec:algorithm}).
    \item \textbf{Techniques.} We introduce cross-index regularization
    and confidence-based dataset filtering to address the challenges
    that decoupling introduces.
    \item \textbf{Empirics.} We conduct our evaluation on challenging continuous environments spanning increasing dynamical complexity, from simple planar motion (\textsc{Car2D}) to higher-dimensional 3D navigation and attitude control (\textsc{SimpleDrone}, \textsc{DroneAttitude}), and a manipulation-style benchmark (\textsc{Reacher}).
    \drill~achieves $3.82\times$ better
    training scalability and $6.19\times$ better zero-shot generalization
    over \genrl. We also compare {\drill} against four other RL and meta-RL baselines; {\drill} achieves 3.02$\times$ and 3.04$\times$ better training scalability and zero-shot generalization (resp.) against the closest of these baselines~(Section~\ref{sec:experiments}).
\end{itemize}







\paragraph{Related Work.}
Prior work tackles generalization in reinforcement learning through several
complementary ideas. Meta-learning~\cite{zintgraf2019varibad,finn2017model,%
hospedales2021meta} prepares an agent for rapid adaptation by learning an
initialization or latent task representation that can be updated with a
small amount of new experience; unlike \genrl~and \drill, meta-learning
approaches do not perform zero-shot generalization. Goal-conditioned
RL~\cite{sohn2018hierarchical,naderian2021c,liu2022goal} trains a single
policy conditioned on a desired outcome, enabling zero-shot generalization
by changing the goal input. Multi-task RL~\cite{sodhani2021multi,oh2017zero,%
teh2017distral,espeholt2018impala} learns shared representations or a
shared policy across a fixed training distribution, encouraging skill
reuse. Programmatic and logic-based policy
representations~\cite{bastani2018verifiable,verma2018programmatically,%
zhu2019inductive,cao2022galois} impose symbolic structure on the policy
class, improving systematic reuse and interpretability relative to purely
neural policies. While all these lines of work enable generalization by
adapting to a new task, conditioning on a goal, or sharing representations
across a fixed training set, none directly addresses the setting we study:
an indexed inductive family where task instances vary systematically with
the index, and the objective is to learn a single rule that predicts how
the policy should change as the index changes.
\genrl~\cite{subramanian2025inductive} is, to our knowledge, the only
prior method that explicitly targets this form of inductive generalization.
The use of temporal logics in RL has enabled long-horizon learning across
a range of
settings~\cite{taylor2024,aksaray2016q,brafman2018ltlf,de2019foundations,%
hasanbeig2018logically,hasanbeig2019,yuan2019modular,ijcai2019-0557,%
li2017reinforcement,jothimurugan2022specification,alur2022framework,%
yang2021tractability, liu2023constrained, svoboda2024reinforcement, guo2025one, majumdar2025regret, alur2026specification}; temporal logics are useful for
inductive generalization as they provide structured syntax to
capture task similarities.

\section{Preliminaries}
\label{sec:prelims}

\paragraph{Markov Decision Process (MDP).~}
The environment in RL is modeled as a {\em Markov Decision
Process} (MDP) $\M = (S, A, P, \eta)$, where $S \subseteq \R^n$ is a continuous
state space, $A \subseteq \R^m$ is a continuous action space,
$P(s,a,s') = p(s'\mid s,a)\in\R_{\geq 0}$ is the transition density of moving
to $s'$ after taking action $a$ in state $s$, and
$\eta: S \rightarrow \R_{\geq 0}$ is the initial-state distribution. 

A \emph{rollout}
$\zeta\in\traj$ is either an infinite sequence
$\zeta = s_0\xrightarrow{a_0}s_1\xrightarrow{a_1}\cdots$
or a finite sequence
$\zeta=s_0\xrightarrow{a_0}\cdots\xrightarrow{a_{t-1}} s_t$,
where $s_i \in S$ and $a_i \in A$. 
We let $\traj_f$ denote the set of finite rollouts. A (deterministic)
\emph{policy} $\pi:\traj_f \to A$ maps a finite rollout to an action.

In RL, the transition probabilities are assumed to be unknown, and the MDP is accessed
only through sampling. Given a policy $\pi$, a rollout is
generated by sampling an initial state $s_0\sim\eta(\cdot)$ and then iterating:
take $a_i=\pi(\zeta_{0:i})$ and sample the next state
$s_{i+1}\sim p(\cdot\mid s_i,a_i)$.

\paragraph{Specification Language.~}
We specify  long-horizon tasks using the temporal logic language $\spectrl$~\cite{jothimurugan2019composable}.
A \spectrl specification is built from a set of \emph{atomic predicates}
${\P}_0$ over environment states. Each predicate $p \in {\P}_0$ is associated
with a Boolean-valued semantics $\semantics{p}:S\to\mathbb{B}=\{\true,\false\}$.
We write $s\models p$ to mean $\semantics{p}(s)=\true$. Using these predicates,
the syntax of \spectrl is
$\p ~::=~ \eventually{b} \mid \p_1 \always{b} \mid \p_1; \p_2 \mid \choice{\p_1}{\p_2}$.

Every specification $\phi$ induces a satisfaction function
$\semantics{\phi}:\traj\to\mathbb{B}$; we write $\zeta\models\phi$ iff
$\semantics{\phi}(\zeta)=\true$. `$\eventually$' encodes a
reachability objective and `$\always$' encodes a safety constraint, while `;'
and `\texttt{or}' denote sequencing and disjunction, respectively. The formal semantics is  in Appendix~\ref{ap:spectrl}.

The probability with which a policy $\pi$ satisfies a specification $\varphi$ in an MDP $\M$ (with initial state distribution $\eta$) is given by $\Pr \big[\pi\models \M, \varphi \big] \triangleq \mathbb{E}_{\zeta \sim \pi,\eta}\big[\zeta \models \varphi \big]$ where $\zeta$ is a rollout of policy $\pi$ in $\M$ with initial state distribution $\eta$.




\section{Inductive Generalization and Problem Statement} \label{sec:ig}

\subsection{Inductive Task Families and Generalization}

An \emph{inductive task family} is an indexed collection
$\taskfam=\{\task_i\}_{i=0}^L$ of tasks where each task $\task_i$
pairs an MDP $\M_i$ with a \spectrl\ specification $\spec_i$. The
family is \emph{inductive} if all tasks share the same \spectrl\
syntactic structure and differ only in the instantiation of atomic
predicates or the initial state distribution. Formally, there exists
a fixed specification form $\spec(\cdot)$ and predicate instantiations
$\{b_i\}_{i=0}^L$ such that $\spec_i = \spec(b_i)$, where predicate
parameters evolve with the index via structured update operators
(e.g., translation, scaling, or parameter drift).

\begin{figure}[t]
  \centering
  \begin{subfigure}[t]{0.53\linewidth}
    \centering
    \includegraphics[width=\linewidth,height=4cm]{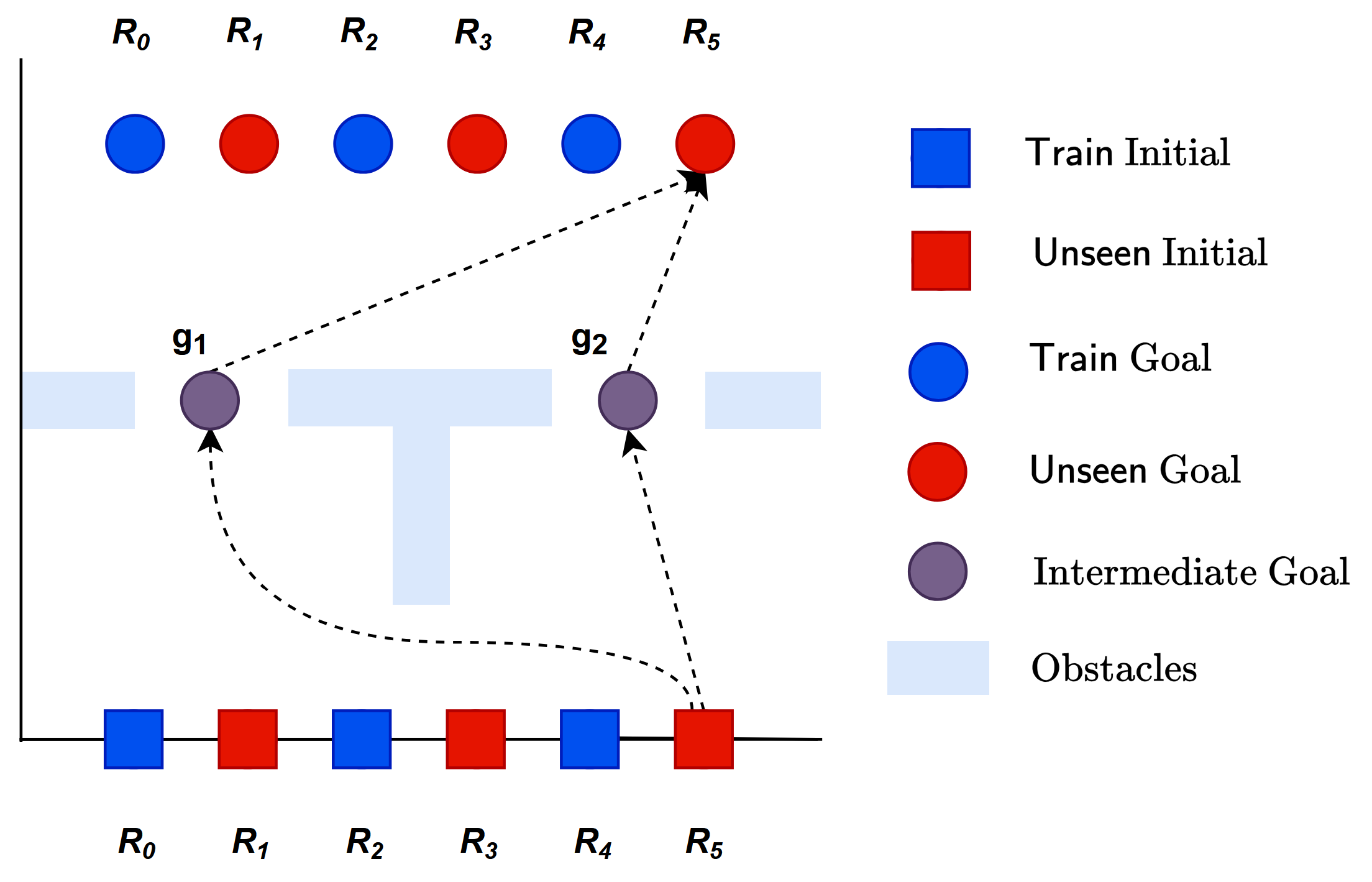}
    \caption{}
    \label{fig:choice_spec}
  \end{subfigure}
  \begin{subfigure}[t]{0.41\linewidth}
    \centering
    \includegraphics[width=\linewidth,height=4cm]{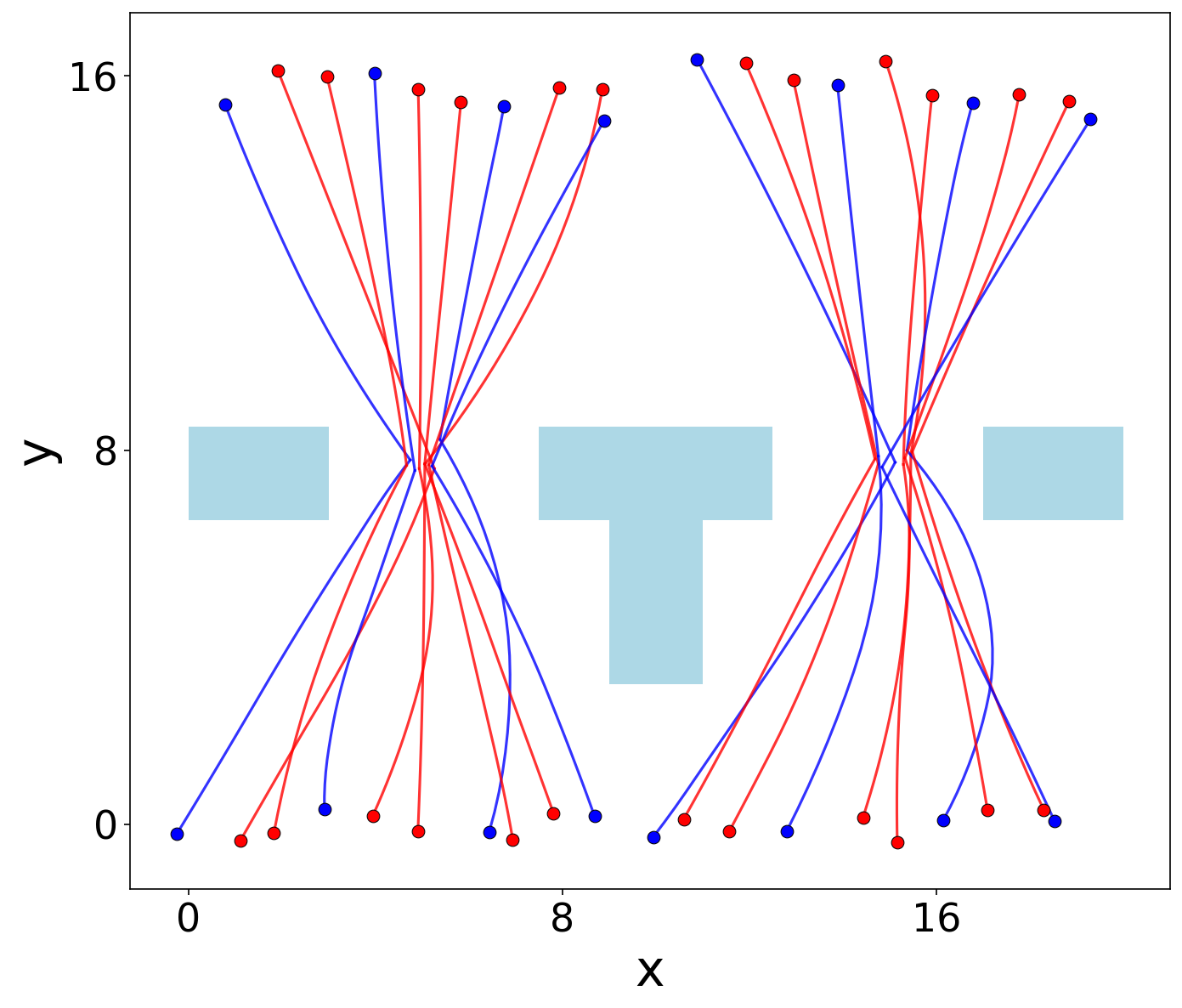}
    \caption{}
    \label{fig:choice_traj}
  \end{subfigure}
  \caption{\textbf{Choice Benchmark in Cartesian 2D Plane.}
  \textbf{(a)} Illustration of the Choice (1-level): from initial region, reach
  either $g_1$ or $g_2$ then reach the final goal; initial region and the  goal
  shift across indices.
  \textbf{(b)} Trajectories produced by \drill:
  \textcolor{blue}{blue} denotes $\Train$ tasks and
  \textcolor{red}{red} denotes $\Unseen$ tasks.}
  \label{fig:motivating_example}
\end{figure}

Fig.~\ref{fig:choice_spec} illustrates an inductive task family in
a 2D plane. In each task instance $\task_k$, the agent
starts from $\Init_k$ and must reach $\mathsf{goal}_k$ after visiting
either $\mathsf{g_1}$ or $\mathsf{g_2}$, while avoiding obstacles:
\[
\spec_k \;\triangleq\;
\Big(
\choice{\eventually{\reach(\mathsf{g_1})}}
       {\eventually{\reach(\mathsf{g_2})}}
;\ \eventually{\reach(\mathsf{goal}_k)}
\Big)\always{\Safe}.
\]
Here $\mathsf{g_1}$, $\mathsf{g_2}$ are fixed; only $\Init_k$ and
$\mathsf{goal}_k$ shift right with $k$. The family is inductive
because the same specification template is reused and only its
grounding evolves.

Given an inductive task family $\taskfam=\{\task_i\}_{i=0}^L$ and
training indices $\Train \subseteq \{0,\dots,L\}$ with $0,L\in\Train$,
the goal of \emph{inductive generalization} is to 
 produce \emph{zero-shot} policies for unseen indices
$\Unseen = \{0,\dots,L\}\setminus\Train$, i.e., policies
satisfying their tasks with high probability without further training:
$\Pr\!\big[\pi_j \models \task_j\big]
\text{ is high for many } j\in\Unseen$.
In Fig.~\ref{fig:motivating_example}, blue regions are training
indices and red regions are unseen; the algorithm trains on blue
tasks only and must generalize to both.

\subsection{Problem Statement: $\kappa$-Learning}
\label{sec:kappa}

The key insight of \genrl~ is that
inductively related tasks admit inductively related policies. If
$\pi_i$ denotes the policy for task $\task_i$, one can posit an
inductive relation $\Omega:\Pi\to\Pi$ such that
$\pi_{i+1}=\Omega(\pi_i)$. Since learning $\Omega$ directly is
intractable, \genrl~approximates it as an $m$-degree polynomial:
\[
\Omega(\pi_i)
= \kappa_m \odot [\pi_i]^m + \cdots + \kappa_1 \odot [\pi_i]
  + \kappa_0,
\]
where $\kappa_0,\dots,\kappa_m$ share the dimension of the neural
policy parameters, $\odot$ is the Hadamard product, and $[\pi_i]^r$
denotes elementwise power. Inductive generalization thus reduces to
learning these $\kappa$-coefficients from $\Train$ and unrolling
the learned template from $\pi_0$ to generate zero-shot policies.

A subtlety arises because policies need not evolve smoothly across
all indices even when tasks do. Consider  Fig.~\ref{fig:motivating_example} where policies on the far left route through $g_1$ and those on the far
right through $g_2$, with a discontinuous switch in between.
Following \genrl, we handle this via a \emph{compositional} reduction:
complex \spectrl\ specifications are decomposed into \emph{reach-avoid}
subspecifications of the form
$\varphi(\Goal,\Safe):=\big(\eventually{\Goal}\big)\always{\Safe}$,
$\kappa$-learning is applied to each, and the results are composed.
This ensures inductive assumptions are placed only on the simplest
specifications \spectrl\ can encode; see~\cite{subramanian2025inductive,
jothimurugan2021compositional} for details. The central learning
problem is:

\begin{definition}[$\kappa$-learning]
\label{def:kappalearning}
Given an inductive family of reach-avoid tasks
$\taskfam=\{\task_i\}_{i=0}^L$, training indices
$\Train=\{i_1<\cdots<i_n\}$ with $i_1=0$, $i_n=L$, polynomial
degree $m$, and base policy $\pi_0$, learn coefficients
$\kappa_0,\dots,\kappa_m$ such that
\[
\kappa^*_0, \cdots, \kappa^*_m \;\in\;
\operatorname*{arg\,max}_{\kappa_0,\cdots,\kappa_m}
\sum_{i\in\Train} \Pr[\pi_i \models \task_i],
\]
where $\pi_i$ is obtained by unrolling the $\kappa$-polynomial
from $\pi_0$.
\end{definition}

\paragraph{\genrl's approach and its limitation.}
\genrl~solves Def.~\ref{def:kappalearning} via a coupled RL
loop (Fig.~\ref{fig:flow-diagram-genrl}): it unrolls the current
$\kappa$ to generate $\{\pi_i\}_{i\in\Train}$, executes these
policies on their training tasks, and updates $\kappa$ from the
aggregated multi-task reward. The fundamental limitation is that
$\kappa$-updates depend on policies that are themselves still
learning --- errors compound across the loop, and instability
worsens as $|\Train|$ grows. \drill~resolves this by decoupling
the two phases, as described in Section~\ref{sec:algorithm}.


\section{Algorithm}
\label{sec:algorithm}

Algorithm~\ref{algo:bc-template} (Appendix~\ref{sec:supp:implementation}) describes our \emph{decoupled} training
procedure, \drill, which learns task-specific teacher policies $\hat{\pi}_i$ for their respective tasks $\task_i$ in Stage I, and then uses these teachers as oracles to learn a \textit{general} policy template in Stage II. Specifically, Stage II learns the template parameters $\kappa$ such that each task-specific policy instance obtained by unrolling the $\kappa$-template imitates the corresponding teacher.

\subsection{Stage I: Teacher-labeled dataset aggregation}

\label{alg:teacher_loss}

Stage~1 constructs, for each training index $i\in\Train$, a final dataset
$\widetilde{\mathcal{D}}_{i}$ of teacher-labeled state--action pairs.
This can be accomplished simply by learning teacher policies $\pi_i$ for all $i \in \Train$ using a standard RL approach independently.
This simple pipeline faces two critical issues:

\begin{itemize}[leftmargin=0.05cm]
\item \textbf{(C1)} \textbf{Cross-index teacher drift.}
A task may be successfully accomplished by a diverse set of policies. As Stage~II must fit {all} of these teacher policies using a
\textit{single} set of template coefficients $\kappa$, which may not be possible if the teachers are trained independently in an unconstrained manner---as adjacent teachers $\hat{\pi}_i$ 
and $\hat{\pi}_{i^-}$ can end up being very different, even when $\task_i$ and $\task_{i^-}$
are similar, making it hard for a single $\kappa$ to fit
all indices simultaneously.
We address this by adding \emph{cross-index regularization} during teacher training (see Section~\ref{cross_index}).


\item \textbf{(C2)} \textbf{Coverage and unreliable labels.}
While these test-time rollouts mostly visit \emph{high-quality states}, i.e., the subset of the state space where the teacher has been sufficiently trained and
consistently achieves high specification satisfaction, they provide limited
coverage of the set of states on which a template-generated policy may
need to act during unrolling. Hence, the candidate state set must be \textit{expanded} to include a wider range of states.  However, diversifying the sampling procedure also runs the risk of witnessing \emph{low-quality
states} that lie far outside the region that the teacher has explored, and hence, the teacher actions can be unreliable in such states.
We address this by \textit{training a confidence filter} (Section~\ref{alg:csa}).
\end{itemize}


\subsubsection{Learning teacher policies with cross-index regularization.\label{cross_index}}
For each $\task_i$ with $i\in\Train$, we train a teacher policy $\hat{\pi}_i$ s.t.,  (a) it achieves high reach-avoid success w.r.t its task $\task_i$, and (b) the policy $\hat{\pi_i}$ stays close to the teacher policy learned at the previous training index. 
The teacher $\hat{\pi}_i$ for task $\task_i$ is trained by minimizing 
\[
\mathcal{L}^{(i)}_{\mathrm{teacher}}
=
\mathcal{L}^{(i)}_{\mathrm{RL}}
+
\lambda_{\mathrm{x}}\,\mathcal{L}^{(i)}_{\mathrm{xreg}},
\quad\quad\quad\quad 
\mathcal{L}^{(i)}_{\mathrm{xreg}}
\;\triangleq\;
\|\hat{\pi}_i - \hat{\pi}_{i^-}\|_2^2,
\]
where $\mathcal{L}^{(i)}_{\mathrm{RL}}$ is the single-task RL objective and $\mathcal{L}^{(i)}_{\mathrm{xreg}}$ is the 
\emph{cross-index regularization} on reach-avoid task $\task_i$;
$\lambda_{\mathrm{x}}\ge 0$ controls the strength of the cross-index regularization.

The \emph{cross-index regularization} is given by the squared 
Euclidean distance between the teacher parameter vectors,
where $i^{-}$
denotes the previous element of $\Train$ before $i$ (so $i^{-}<i$ and there is no
training index between them) and  $\|\cdot\|_2$ is the Euclidean norm over policy parameter vectors (i.e., we
treat $\hat{\pi}_i$ and $\hat{\pi}_{i^-}$ as vectors of network weights). This term
provides a {soft continuity constraint} across indices such that it biases the optimizer
toward solutions for $\task_i$ that can be obtained by a small change from the previous
teacher, while still allowing the policy to deviate when the task requires it. 

The teacher policy for this loss function can be trained using a standard RL algorithm (Note: For the base policy $\pi_0$, $\mathcal{L}^{(0)}_{\mathrm{xreg}} = 0$ as there is no previous task). 

\subsubsection{Confidence-filtered state accumulation.} \label{alg:csa}
After training $\hat{\pi}_i$, we build a preliminary dataset
${\mathcal{D}}_{i}$ by sampling states from the teacher’s
on-policy distribution $d^{\hat{\pi}_i}$.
At the same time, to combat the \textit{coverage} problem (see challenge C2 above), we diversify our samples by also sampling from two other sources: the replay-buffer distribution
$\mathrm{Replay}_i$, and the reset distribution $\mathrm{Reset}_i$.
Here, $d^{\hat{\pi}_i}$ collects states visited in specification-satisfying \emph{on-policy}
rollouts of the \emph{converged} teacher $\hat{\pi}_i$ during test-time (i.e., after
training has converged). $\mathrm{Replay}_i$ samples states from the teacher’s replay
buffer accumulated over training (including early exploration states and other
perturbation-induced states), and $\mathrm{Reset}_i$ samples initial states produced by
environment resets or randomized initializations.



However, as discussed in the challenge C2 above, the data points from the replay/reset buffers may be of \textit{low quality} as the teacher may not have witnessed these samples during its exploration. Hence, to avoid learning our $\kappa$ on a noisy dataset, we train a {\em confidence filter}
that assigns a confidence probability $p_i^{\textrm{conf}}(s)$ to states $s$, indicating whether $s$
is a high- or a low-quality state. 

\textbf{Confidence filter training.} 
To train the confidence filter, we maintain two state buffers:
(i) a high-quality state buffer $\mathcal{B}^{\mathrm{high}}_i$ containing states visited
during teacher training in rollouts that satisfy the specification for task $i$, and (ii) a low-quality state buffer $\mathcal{B}^{\mathrm{low}}_i$ containing states from early
exploration, perturbed observations, and reset-induced states, and, more generally, states
from rollouts that do not satisfy (or have lower satisfaction of) the specification.
We assign labels $y=1$ for $s\sim\mathcal{B}^{\mathrm{high}}_i$ and $y=0$ for
$s\sim\mathcal{B}^{\mathrm{low}}_i$. The confidence filter is trained with binary
cross-entropy:
\[
\mathcal{L}^{(i)}_{\mathrm{conf}}
=
\mathbb{E}_{s \sim \mathcal{B}^{\mathrm{high}}_i}\!\left[-\log (p_i^{\mathrm{conf}}(s))\right] 
\quad+\quad
\mathbb{E}_{s \sim \mathcal{B}^{\mathrm{low}}_i}\!\left[-\log\!\big(1-p_i^{\mathrm{conf}}(s)\big)\right].
\]

\textbf{Aggregating the samples.} Once $p_i^{\textrm{conf}}(\cdot)$ is obtained, we convert the
preliminary dataset ${\mathcal{D}}_{i}$ into a labeled dataset $\widetilde{\mathcal{D}}_{i}$ by retaining only those
candidates that the confidence filter deems high-quality. Specifically, we keep a
candidate state $s\in{\mathcal{D}}_{i}$ only if
$p_i^{\mathrm{conf}}(s)\ \ge\ \tau_c$,
where $\tau_c\in(0,1)$ is a fixed threshold. For each retained state, we query the teacher
for an action label $a=\hat{\pi}_i(s)$ and add $(s,a)$ to the final dataset: $
\widetilde{\mathcal{D}}_{i}
=
\{(s,a)\ :\ s \in {\mathcal{D}}_{i},\ p_i^{\mathrm{conf}}(s)\ge\tau_c,\ a=\hat{\pi}_i(s)\}
$. This retains the coverage benefits of replay/reset while discarding candidates likely to
yield unreliable labels.


\paragraph{Implementation Details.}
Instead of training the teacher policy and the confidence filter separately,
we jointly optimize both via $
\mathcal{L}^{(i)}_{\mathrm{teacher-conf}}
=
\mathcal{L}^{(i)}_{\mathrm{teacher}}
+
\lambda_{\mathrm{conf}}\,\mathcal{L}^{(i)}_{\mathrm{conf}},
$ where $\lambda_{\mathrm{conf}} \in [0,1]$ controls how strongly we train the confidence filter.
We train a single neural network for $\mathcal{L}^{(i)}_{\mathrm{teacher-conf}}$ with a shared
trunk $h_i$ and two heads: an action (teacher policy) head and a confidence-filter head.
Therefore, for a state $s$,
$z = h_i(s)$, $\quad a = \mathrm{ActHead}_i(z)$, $\quad c_i(s) = \mathrm{ConfHead}_i(z)$.
Here $a$ is the action output by the teacher policy at state $s$, and $c_i(s)\in\R$ is a
confidence value indicating whether $s$ is a high-quality state. We convert this confidence value to a
probability via
$p_i^{\mathrm{conf}}(s)\triangleq \mathrm{Sigmoid}(c_i(s))$. Thus, larger $p_i^{\mathrm{conf}}(s)$ indicates higher confidence that $s$ is a high-quality state.

\subsection{Stage II: Learning $\kappa$ coefficients via Behavioral Cloning}
\label{sec:stage2-bc}


We learn the template coefficients $\kappa$ by fitting the final
datasets $\bigcup_{i\in \Train}\widetilde{\mathcal{D}}_i$. We fit $\kappa$ using
\emph{behavioral cloning (BC)}, a standard imitation-learning objective that trains a policy to match the demonstrated action on a dataset of state--action pairs. In our case, the
demonstrations come from the dataset accumulated in Stage I: for each training index $i\in\Train$, we
construct a final dataset $\widetilde{\mathcal{D}}_i$ of state--action pairs, where each
action is labeled by the teacher policy $\hat{\pi}_i$. Unlike standalone BC, which fits a single policy to a single dataset,
we use BC to fit a \emph{shared} $\kappa$-template whose unrolling generates a policy for
each training index, using all $\{\widetilde{\mathcal{D}}_i\}_{i\in\Train}$ as supervision.
BC fits $\kappa$ by minimizing the discrepancy between the template-generated
actions (from $\pi_i$) and the teacher actions on the teacher-labeled datasets.
Using the squared-error behavioral cloning loss, we minimize
\[
\mathcal{L}_{\mathrm{BC}}(\kappa)
=
\sum_{i\in \Train}
\frac{1}{|\widetilde{\mathcal{D}}_{i}|}
\sum_{(s,a)\in \widetilde{\mathcal{D}}_{i}}
\left\|\pi_i(s)-a\right\|_2^2.
\]
Here, $\pi_i$ is the policy for task $\task_i$ obtained by unrolling the $\kappa$-polynomial. For unrolling formalism, see Appendix~\ref{sec:supp:index_unrolling}. We update $\kappa$ with a first-order optimizer:
\[
\kappa \leftarrow \kappa - \eta\,\nabla_{\kappa}\mathcal{L}_{\mathrm{BC}}(\kappa),
\]
alternating between (i) unrolling the current $\kappa$ to obtain the
template-generated policies $\{\pi_i\}_{i\in\Train}$ and (ii) taking gradient steps
on $\mathcal{L}_{\mathrm{BC}}$ until the loss plateaus. The output is a set of
learned template coefficients $\kappa$ that can be unrolled from base policy
$\pi_0$ to produce policies for any index in the family.

\section{Empirical Evaluation}
\label{sec:experiments}


Our empirical analysis is designed to evaluate three metrics: (a) training scalability, (b) zero-shot generalization, and (c) the correlation between training scalability and zero-shot generalization. 





\subsection{Experimental Setup}

\noindent\textbf{Baselines.}
We compare our method \drill \ to five other methods. \textbf{ARS}~\cite{mania2018simple} trains a single policy using aggregated rollouts across
all training indices in $\Train$, with no explicit index-evolution template. \textbf{BC} \cite{torabi2018behavioral, levine2016end, osa2018algorithmic} 
is a standalone imitation learning approach that trains a single policy on the pooled teacher-labeled dataset
$\bigcup_{i\in\Train}\widetilde{\mathcal{D}}_i$, with no explicit template. 
While neither ARS nor BC is a generalization method, we also compare with 
\textbf{MAML}~\cite{finn2017model}
and \textbf{VariBAD}~\cite{zintgraf2019varibad}, which are strong meta-RL generalization baselines. Finally,  \textbf{\genrl}
is an inductive generalization approach that learns the $\kappa$-coefficients using an ARS-style RL loop. \drill\  borrows the high-level architecture of  \genrl \ but uses the procedure in Section~\ref{sec:algorithm} to train $\kappa$-coefficients.

\noindent\textbf{Environments.}
We evaluate on four continuous-control environments: \textsc{Car2D}, \textsc{Reacher},
\textsc{SimpleDrone}, and \textsc{DroneAttitude}. \textsc{Car2D} is a planar car-like navigation model. The state is the agent
position $s=(x,y)\in\mathbb{R}^2$, and the action is $a=(v_f,\theta)\in\mathbb{R}^2$,
where $v_f$ denotes the forward speed and $\theta$ denotes the heading direction.
\textsc{Reacher} is a two-link planar arm where the state is the end-effector position
$s=(x,y)\in\mathbb{R}^2$ and the action specifies the two joint angles
$a=(\theta_1,\theta_2)\in\mathbb{R}^2$ (Fig.~\ref{fig:reacherdyn}). \textsc{SimpleDrone}
models holonomic 3D navigation with state $s=(x,y,z)\in\mathbb{R}^3$ and continuous
velocity actions $a=(v_x,v_y,v_z)\in\mathbb{R}^3$.
\textsc{DroneAttitude} augments 3D position with attitude variables, using
$s=(x,y,z,\psi,\theta)\in\mathbb{R}^5$, and controls the agent through body-frame velocity
commands and angular rates,
$a=(v_f,v_s,v_u,\omega_\psi,\omega_\theta)\in\mathbb{R}^5$. See Figure~\ref{fig:envs} in Appendix for illustrations.

Overall, these environments span increasing dynamical complexity, from simple planar motion
(\textsc{Car2D}) to higher-dimensional 3D navigation and attitude control
(\textsc{SimpleDrone}, \textsc{DroneAttitude}), and a manipulation-style benchmark
(\textsc{Reacher}).

\noindent\textbf{Inductive Task Families.}
We create several inductive task families over the environments (see Appendix~\ref{sec:supp:specs}). For \textsc{Reacher}, the inductive task families consist of variations of the tower-destacking problem (Fig.~\ref{fig:reacher_all}) with the tower varying in height from 8 to 10 blocks.  
For the other environments, we use
\textsc{$k$-reachability} (long-horizon sequencing) where the initial state distribution and all the $k$ intermediate points shift to the right as the task index increases (Appendix Fig.~\ref{fig:spec_nreach}). 
In \textsc{Car2D} we also consider \textsc{Choice} (branching via
disjunction) with 1-level (Fig.~\ref{fig:motivating_example}) and 2-level branching variants (Appendix Fig.~\ref{fig:spec_choice_2}).

\definecolor{arsblue}{HTML}{1F77B4}
\definecolor{bcgreen}{HTML}{2CA02C}
\definecolor{mamlorange}{HTML}{FF7F0E}
\definecolor{varibadbrown}{HTML}{8C564B}
\definecolor{genrlpurple}{HTML}{9467BD}
\definecolor{dipsred}{HTML}{D62728}

\begin{figure*}[t]
\centering
Legend:
{\small {\color{arsblue}\rule[0.6ex]{1.2em}{1pt}$\,\bullet$} ARS,
{\color{bcgreen}\rule[0.6ex]{1.2em}{1pt}$\,\blacksquare$} BC,
{\color{mamlorange}\rule[0.6ex]{1.2em}{1pt}$\,\blacklozenge$} MAML,
{\color{varibadbrown}\rule[0.6ex]{1.2em}{1pt}$\,\blacktriangle$} VariBAD,
{\color{genrlpurple}\rule[0.6ex]{1.2em}{1pt}$\,\blacktriangledown$} GenRL, and
{\color{dipsred}\rule[0.6ex]{1.2em}{1pt}{\large $\times$}} \drill~(Ours).}

\setlength{\abovecaptionskip}{1pt}
\setlength{\belowcaptionskip}{1pt}

\setcounter{subfigure}{0}
\par\medskip
{\centering\small\textbf{(i) \textsc{Car2D} $k$-Reachability}\par}
\vspace{0.2em}

\begin{subfigure}[t]{0.33\textwidth}
  \centering
  \includegraphics[width=\linewidth]{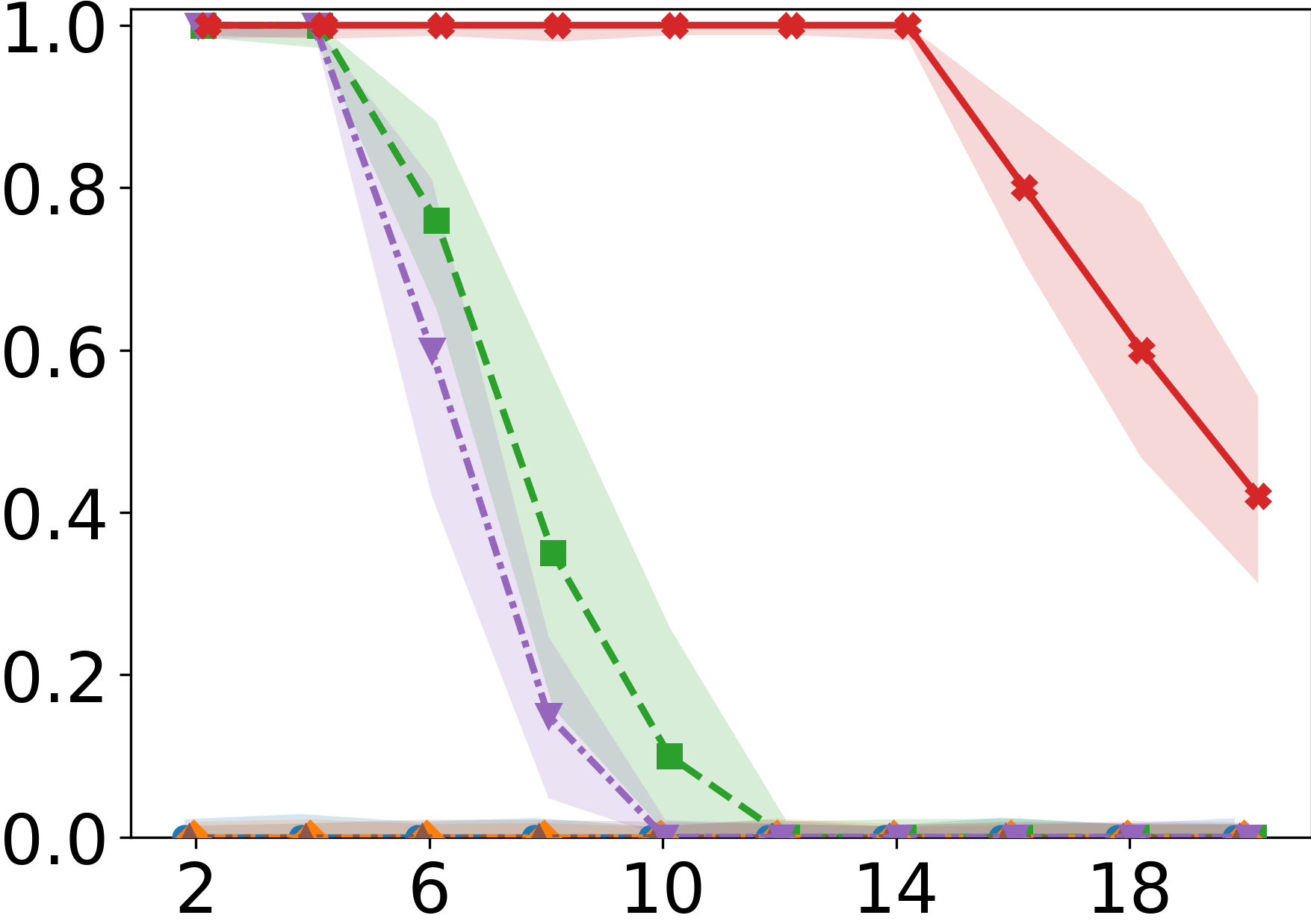}
  \caption{}\label{fig:r1a}
\end{subfigure}
\begin{subfigure}[t]{0.33\textwidth}
  \centering
  \includegraphics[width=\linewidth]{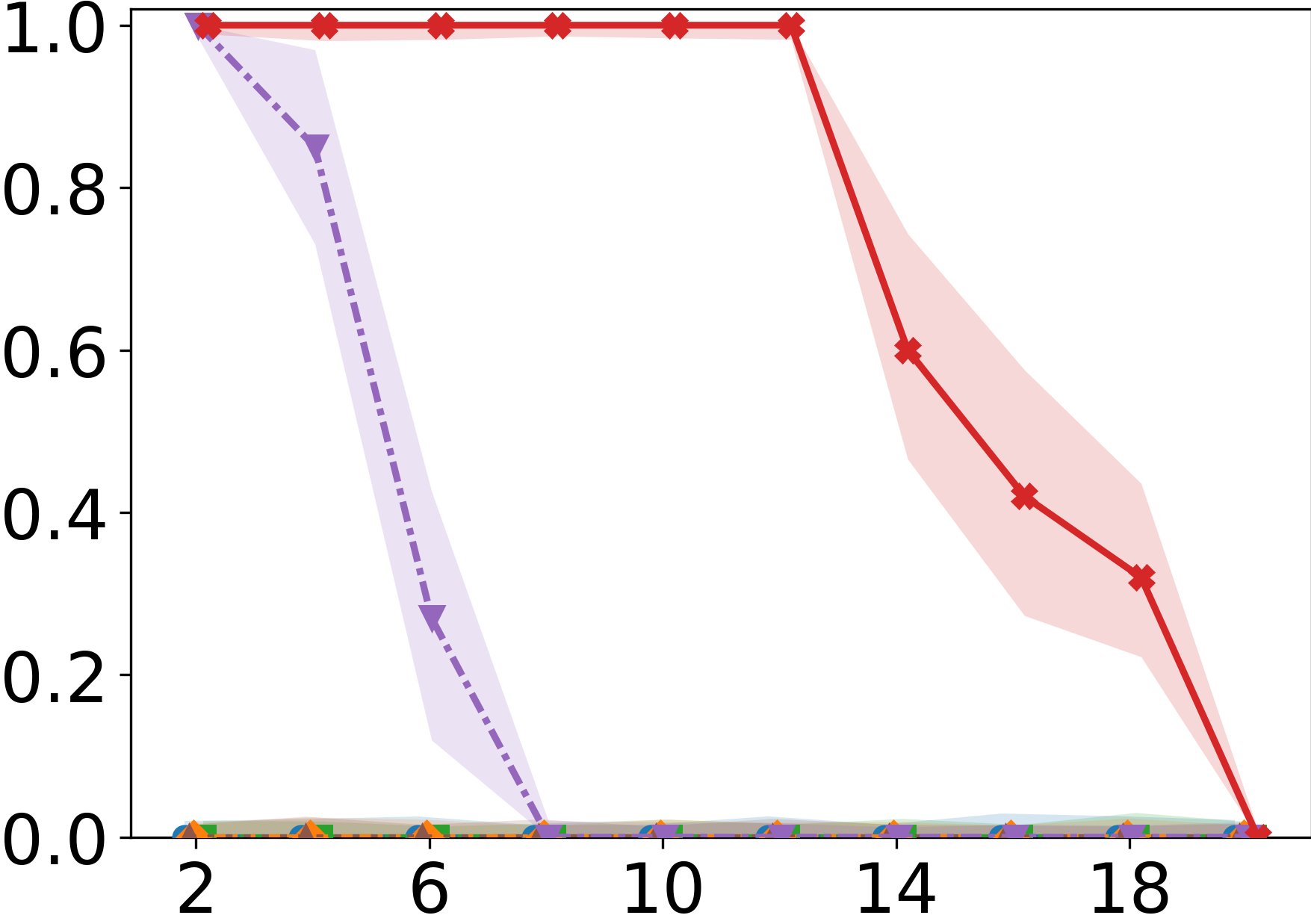}
  \caption{}\label{fig:r1b}
\end{subfigure}
\begin{subfigure}[t]{0.32\textwidth}
  \centering
  \includegraphics[width=\linewidth]{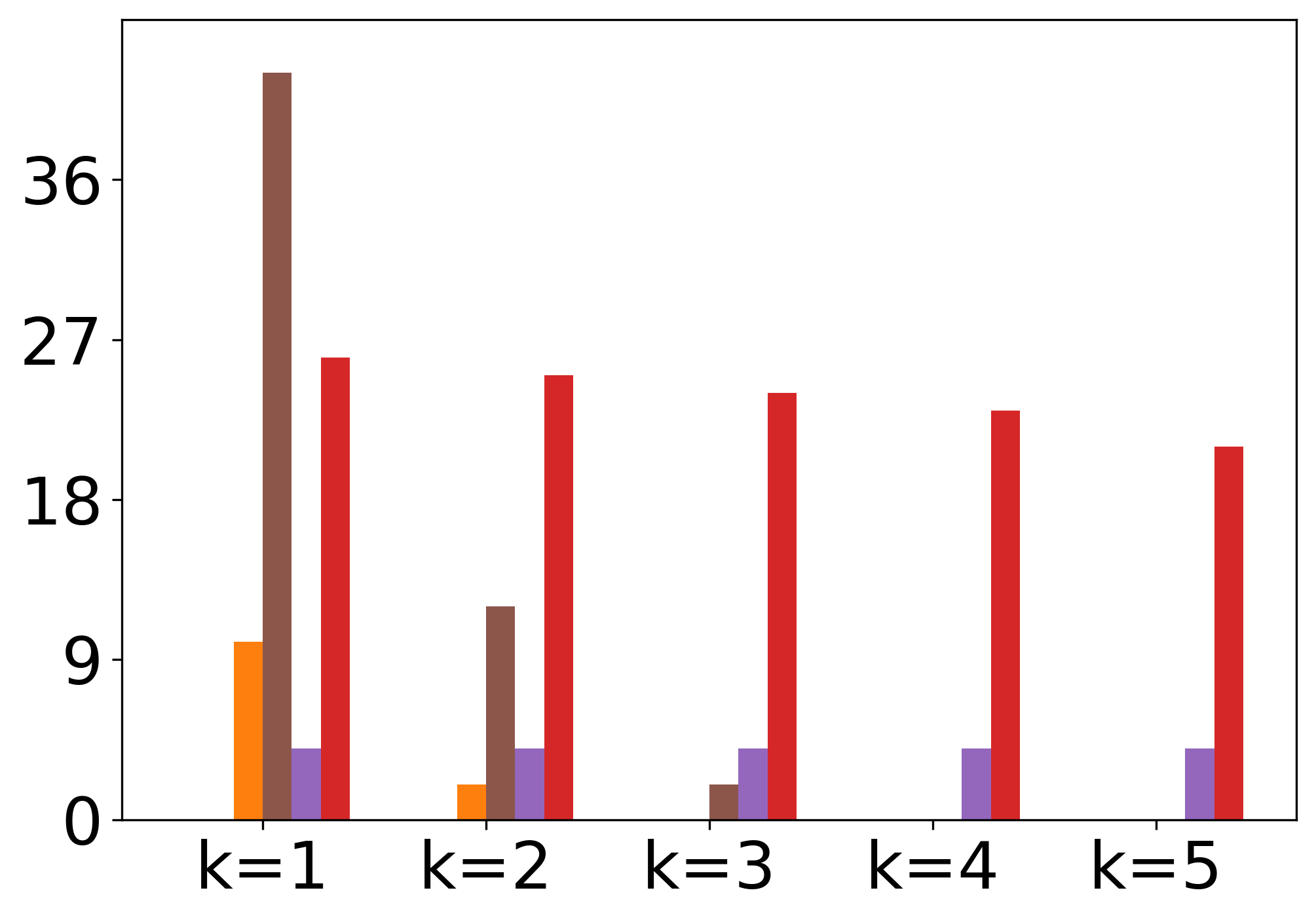}
  \caption{}\label{fig:r1c}
\end{subfigure}

\setcounter{subfigure}{0}
\par\medskip
{\centering\small\textbf{(ii) \textsc{SimpleDrone} $k$-Reachability}\par}
\vspace{0.2em}

\begin{subfigure}[t]{0.33\textwidth}
  \centering
  \includegraphics[width=\linewidth]{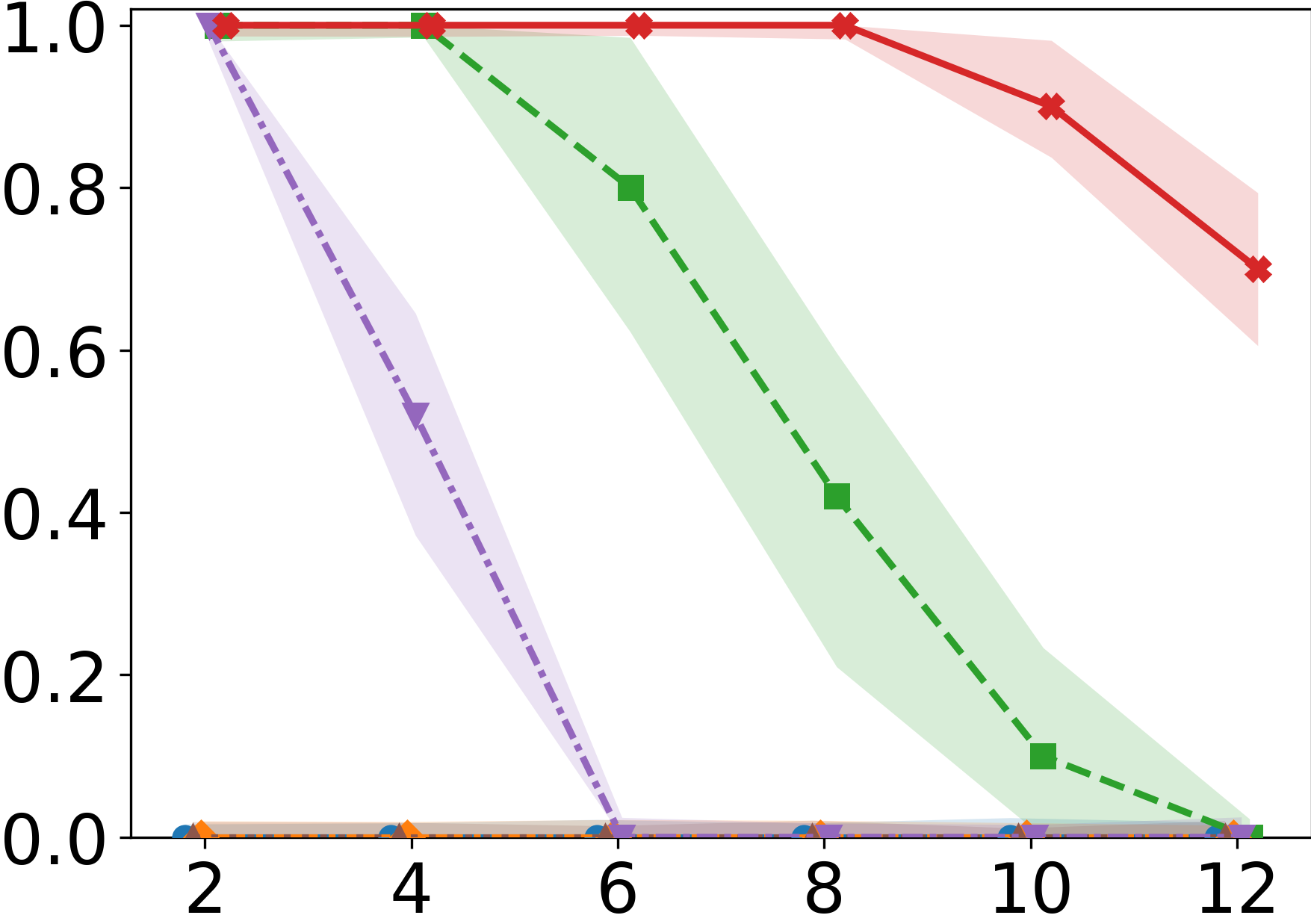}
  \caption{}\label{fig:r2a}
\end{subfigure}
\begin{subfigure}[t]{0.33\textwidth}
  \centering
  \includegraphics[width=\linewidth]{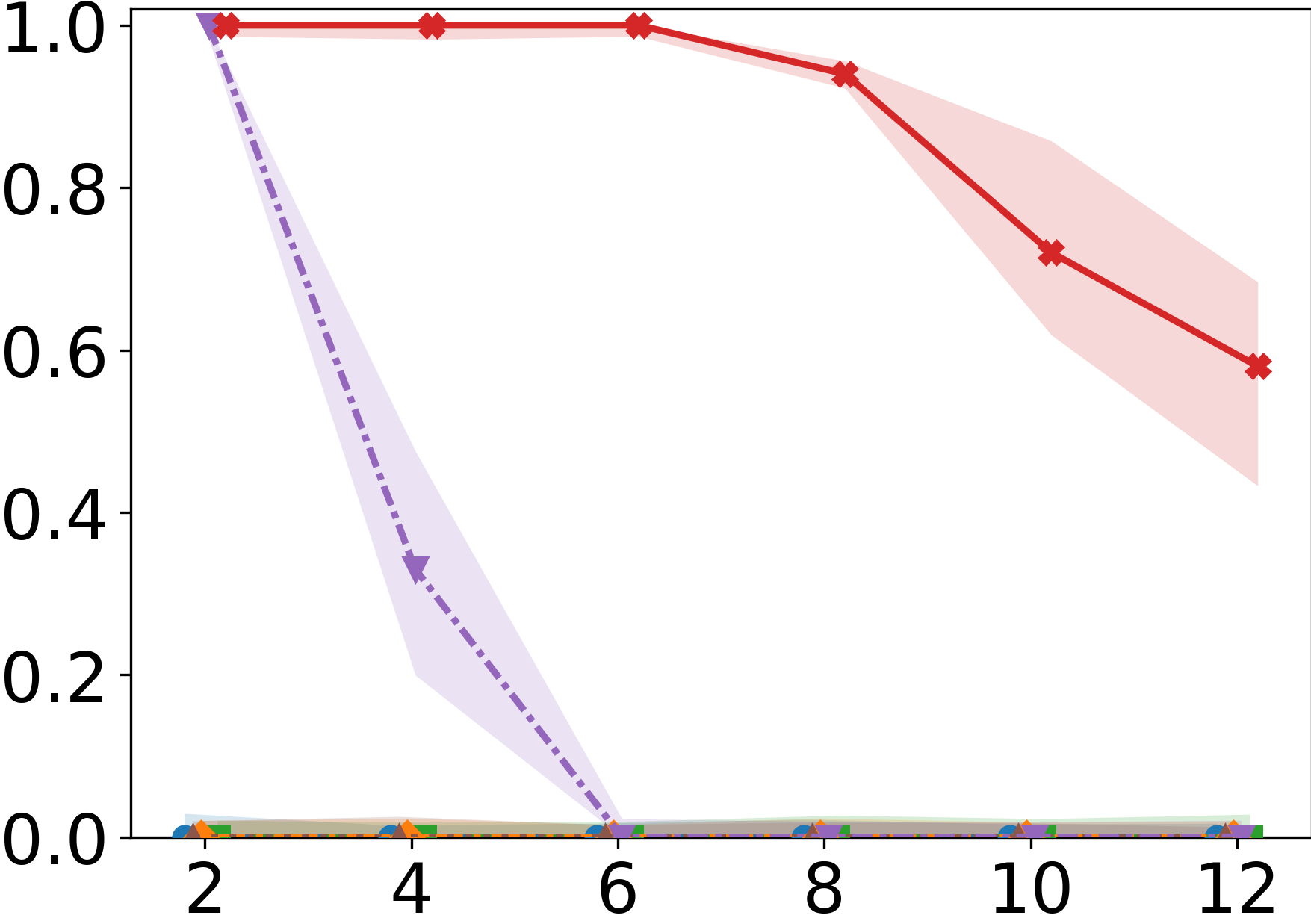}
  \caption{}\label{fig:r2b}
\end{subfigure}
\begin{subfigure}[t]{0.32\textwidth}
  \centering
  \includegraphics[width=\linewidth]{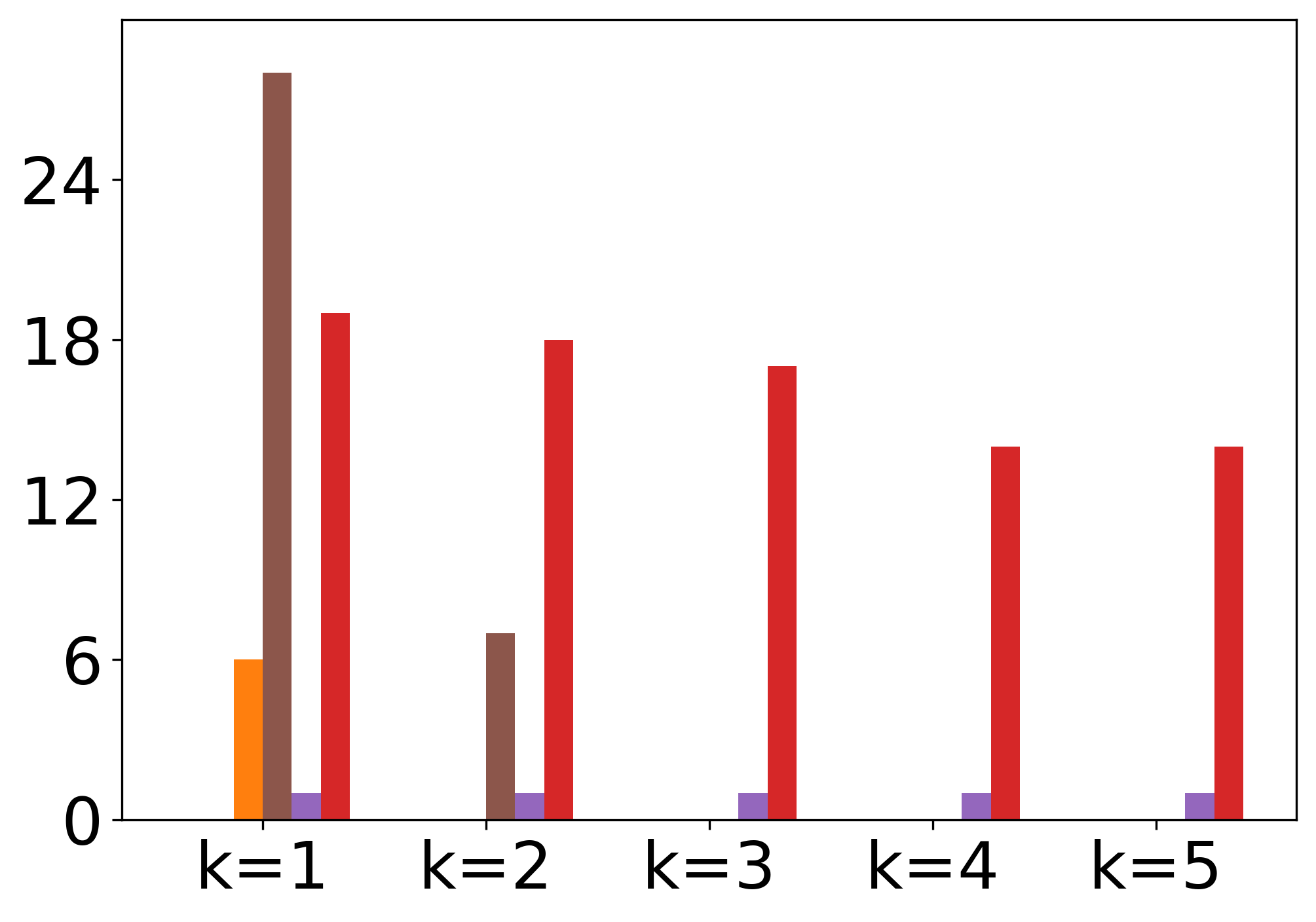}
  \caption{}\label{fig:r2c}
\end{subfigure}

\setcounter{subfigure}{0}
\par\medskip
{\centering\small\textbf{(iii) \textsc{DroneAttitude} $k$-Reachability}\par}
\vspace{0.2em}

\begin{subfigure}[t]{0.33\textwidth}
  \centering
  \includegraphics[width=\linewidth]{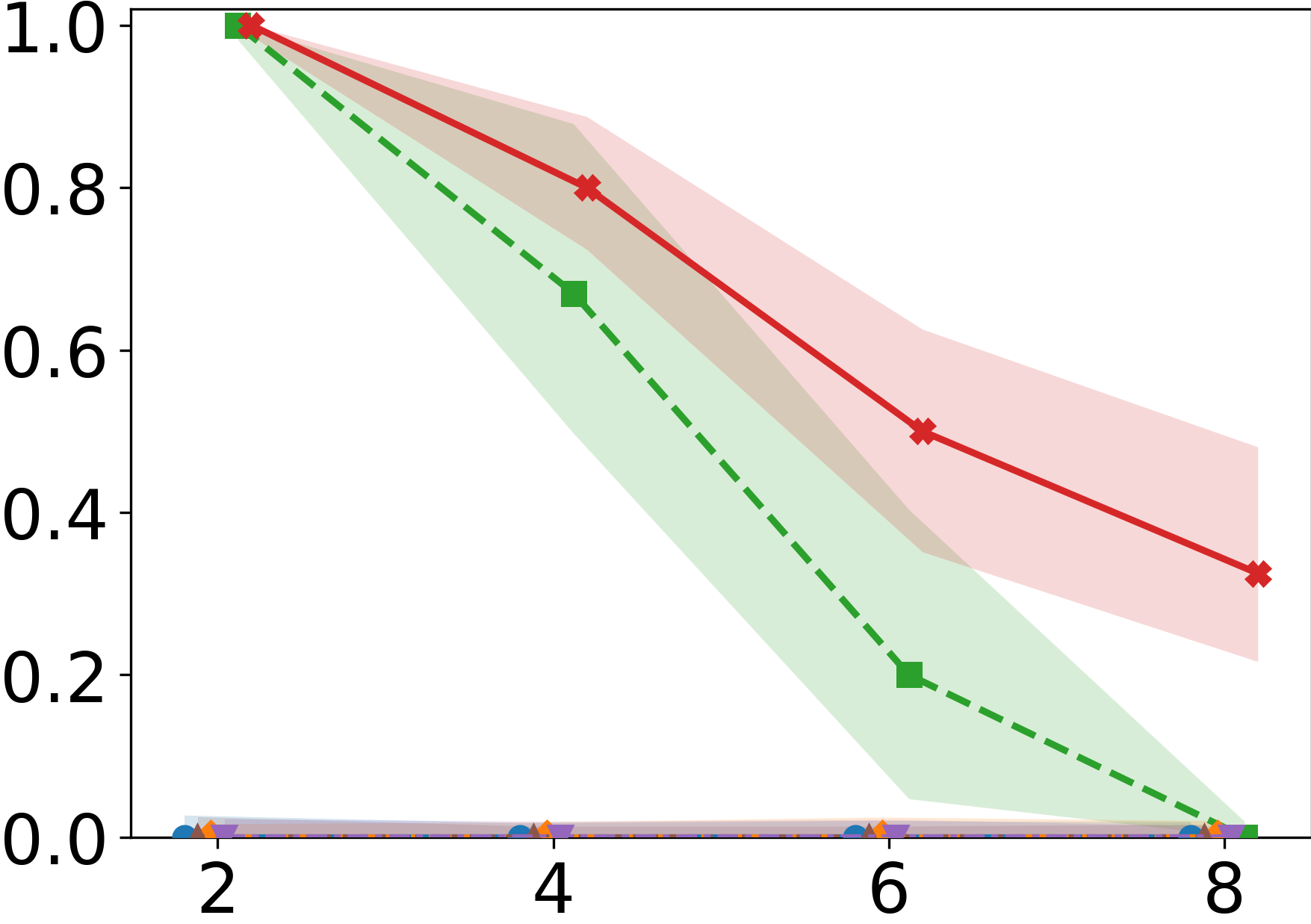}
  \caption{}\label{fig:r3a}
\end{subfigure}
\begin{subfigure}[t]{0.33\textwidth}
  \centering
  \includegraphics[width=\linewidth]{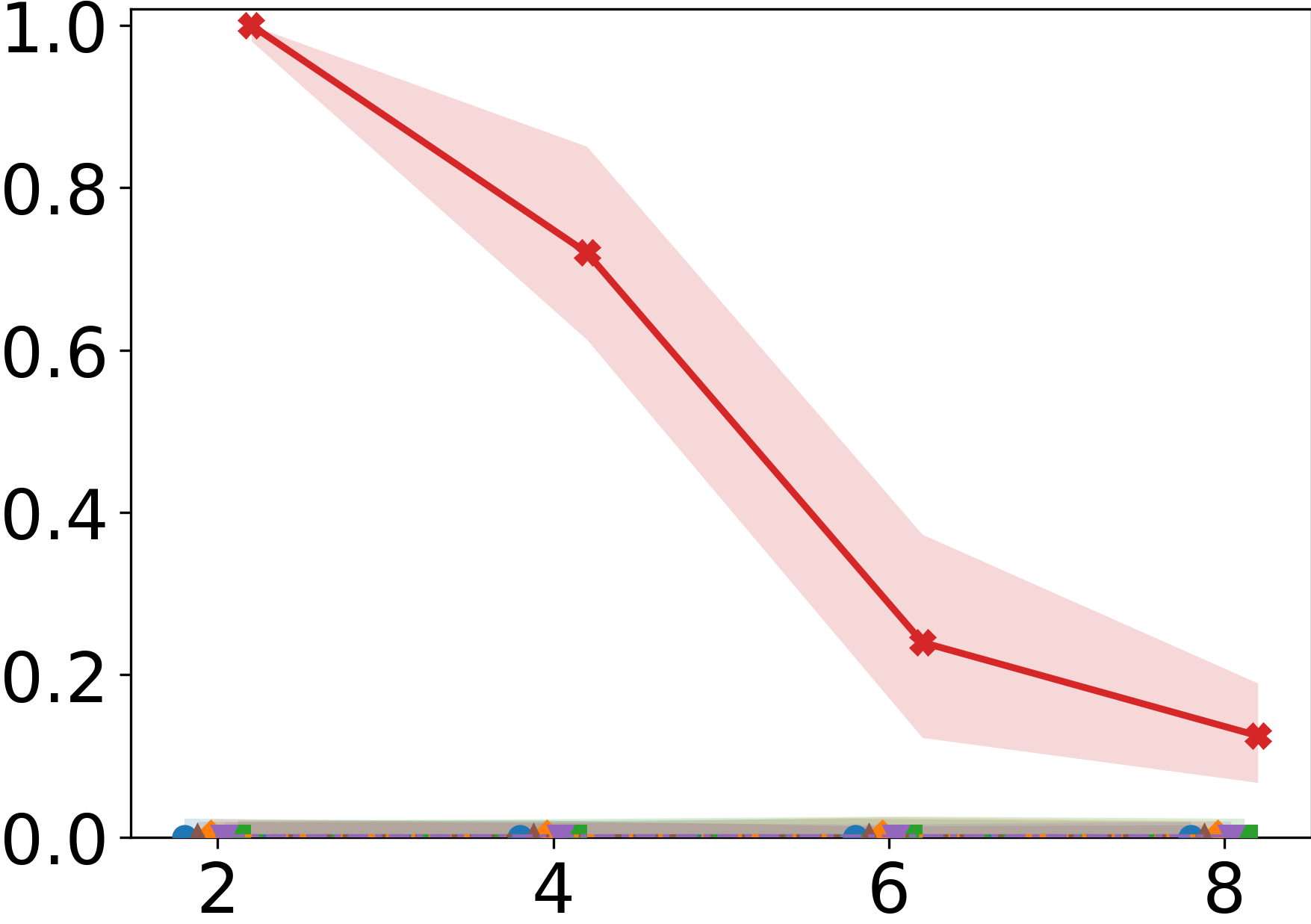}
  \caption{}\label{fig:r3b}
\end{subfigure}
\begin{subfigure}[t]{0.32\textwidth}
  \centering
  \includegraphics[width=\linewidth]{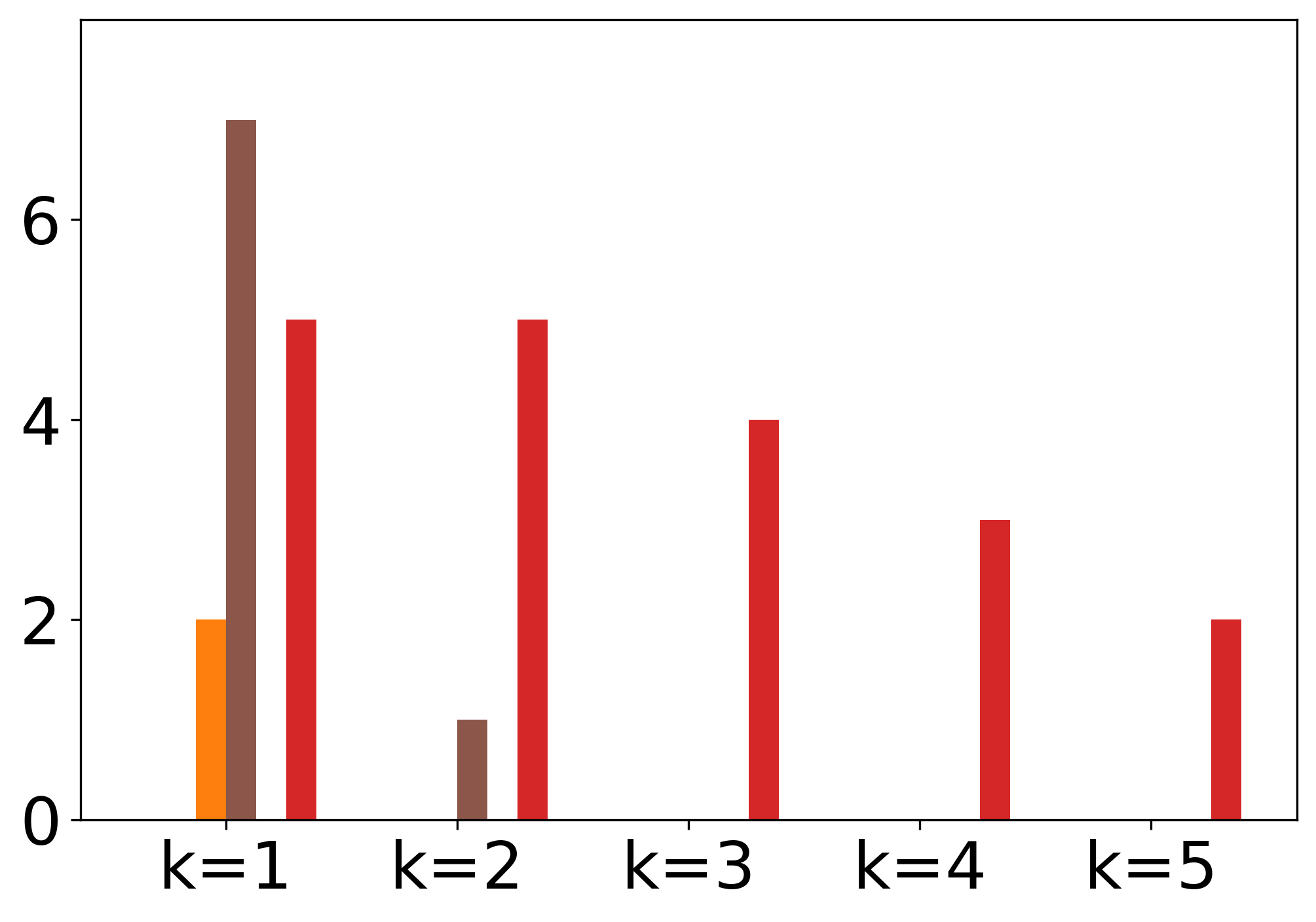}
  \caption{}\label{fig:r3c}
\end{subfigure}

\caption{\textbf{Scalability plots}. Each row corresponds to one environment on $k$-reachability tasks. \textbf{Column (a)} shows the successful training ratio ($y$-axis) versus number of training tasks $|\Train|$ ($x$-axis) for $k=5$. \textbf{Column (b)} shows the zero-shot generalization ratio ($y$-axis) versus number of training tasks $|\Train|$ ($x$-axis) for $k=5$. \textbf{Column (c)} shows absolute zero-shot generalization ($y$-axis) for $k \in \{1,\ldots,5\}$ ($x$-axis). We report mean ($\pm$ standard deviation) performance across \textbf{ten} random seeds.}

\label{fig:successvstasks_all}
\end{figure*}

\noindent\textbf{Train/unseen splits and index spacing.} Each benchmark defines task instances indexed by $\{0,\dots,L\}$. A training run chooses an
ordered subset of training tasks $\Train=\{i_1<\cdots<i_n\}$ with $i_1=0$ and $i_n=L$, and the remaining unseen tasks $\Unseen=\{0,\dots,L\}\setminus\Train$.

\noindent\textbf{Metrics.} Given a policy and a specification (a single task), we estimate the probability of specification satisfaction via repeated rollouts of the policy. A policy is said to {\em succeed} if this estimate exceeds a given threshold value $\delta$. For all our experiments, we use $1000$ rollouts
and $\delta=0.9$.

For a generalization baseline and an inductive task family, we define two metrics: (a) successful training ratio, and (b) zero-shot generalization ratio. Given a training set $\Train$, the {\em successful training ratio} is given by $\frac{|{ i \in \Train : \pi_i \text{ succeeds}}|}{|\Train|}$, where $\pi_i$ is the policy obtained by the baseline for the training task indexed by $i$. In our training plots, we report the successful training ratio on the $y$-axis against the number of training tasks $|\Train|$ on the $x$-axis. Given a testing set $\Unseen$, the {\em absolute zero-shot generalization} is given by $|{ i \in \Unseen : \pi_i \text{ succeeds}}|$. The {\em zero-shot generalization ratio} is given by $\frac{|{ i \in \Unseen : \pi_i \text{ succeeds}}|}{|\Unseen|}$, i.e., the fraction of unseen tasks solved. In our zero-shot generalization plots, we report this ratio on the $y$-axis against the number of training tasks $|\Train|$ on the $x$-axis.

\subsection{Results and Observations}

Fig.~\ref{fig:successvstasks_all} summarizes results for some \textsc{$k$-Reachability} tasks across multiple inductive task families:
\textsc{Car2D}, 
\textsc{SimpleDrone},
\textsc{DroneAttitude}; plots for remaining benchmarks are in the Appendix~\ref{app:more-results}.

\noindent\textbf{Training scalability.}
Training scalability is evaluated as successful training ratio vs. $|\Train|$, as demonstrated in Fig.~\ref{fig:successvstasks_all}a. Higher values indicate better scalability. Fig.~\ref{fig:successvstasks_all}a shows that our approach \drill~maintains the strongest training scalability. 
Aggregated across all benchmarks and
training configurations, \drill~improves the fraction of training indices solved by
$3.82\times$ relative to the other inductive generalization approach \genrl. 
We attribute this success to our decoupled approach that reduces $\kappa$-learning to a simple supervised learning objective (in the form of behavioral cloning), as opposed to a complex multi-reward aggregation RL-style loop in \genrl.

\noindent\textbf{Zero-Shot Generalization.}
\drill~also achieves the strongest zero-shot generalization across benchmarks,
both in the ratio (Fig.~\ref{fig:successvstasks_all}b) and in the absolute value (Fig.~\ref{fig:successvstasks_all}c). Because template fitting
remains stable across a wider range of training configurations, the learned
$\kappa$ can be unrolled to solve substantially more unseen indices. Aggregated across all benchmarks and training configurations, \drill~improves the fraction of unseen indices solved by $6.19\times$ relative to \genrl.

Notably, \textbf{BC}, which is behavioral cloning without  the template scales well in $|\Train|$ (Fig.~\ref{fig:successvstasks_all}a); however, it does not generalize well (Fig.~\ref{fig:successvstasks_all}b--c). This is not unexpected, since in the absence of a template \textbf{BC} has no context to generalize beyond its supervised training data. This also demonstrates the strength of the template-based generalization approach pursued in inductive generalization. 

\noindent\textbf{Correlation between training scalability and zero-shot generalization.}
We observe a consistent and strong correlation between training
success and generalization, i.e., when a method fails to solve its training indices at larger
$|\Train|$, its zero-shot generalization typically degrades as well
(Fig.~\ref{fig:successvstasks_all}a--b). Though this is an observed trend rather than a causal claim, it suggests that scalable training dynamics
are closely tied to extracting generalizable structure from larger
training index sets.

MAML and VariBAD can perform competitively in simpler settings, but they do not scale reliably to long-horizon or branching tasks. As the task horizon increases, for example, for larger $k$ in the \textsc{$k$-Reachability} experiments in Fig.~\ref{fig:successvstasks_all}c, their training reliability and zero-shot generalization degrade relative to the template-based methods. A similar pattern appears in the \textsc{Choice} benchmarks, where branching decisions are required and both methods struggle because they do not model the branching structure; see Appendix Fig.~\ref{fig:successvstasks_choice}. These results show that while MAML and VariBAD can work in easier regimes, their performance deteriorates as the task structure becomes longer-horizon or more compositional. This trend is consistent with the behavior observed for \genrl, although \genrl~benefits from an explicit template structure.



\begin{figure*}[t]
  \centering

  \begin{subfigure}[t]{0.24\textwidth}
    \centering
    \includegraphics[width=\linewidth]{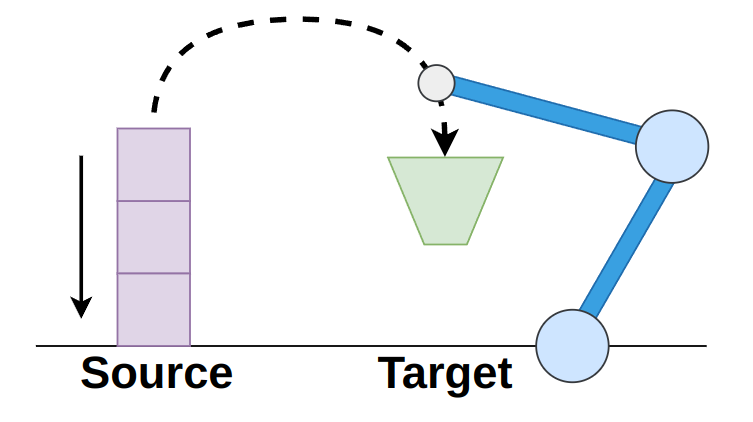}
    \caption{Pick-and-drop (same side) --- Variant 1}
    \label{fig:r1}
  \end{subfigure}\hfill
  \begin{subfigure}[t]{0.24\textwidth}
    \centering
    \includegraphics[width=\linewidth]{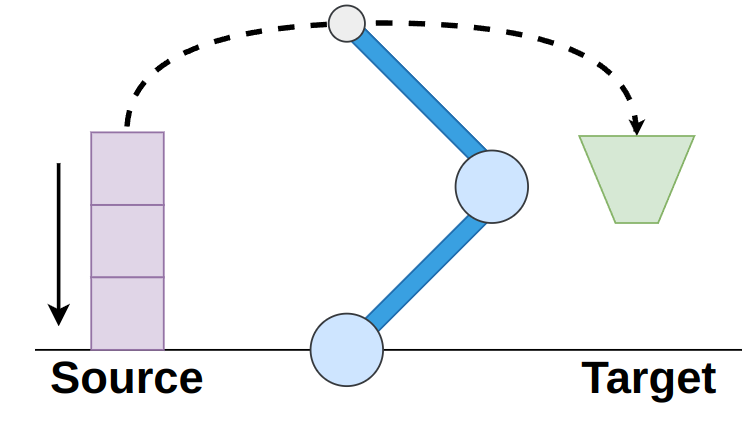}
    \caption{Pick-and-drop (opposite side) --- Variant 2}
    \label{fig:r3}
  \end{subfigure}\hfill
  \begin{subfigure}[t]{0.24\textwidth}
    \centering
    \includegraphics[width=\linewidth]{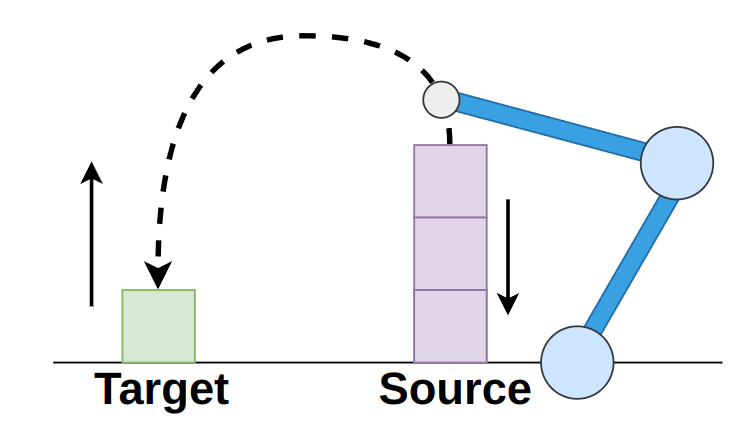}
    \caption{Pick-and-stack (same side) --- Variant 3}
    \label{fig:r2}
  \end{subfigure}\hfill
  \begin{subfigure}[t]{0.24\textwidth}
    \centering
    \includegraphics[width=\linewidth]{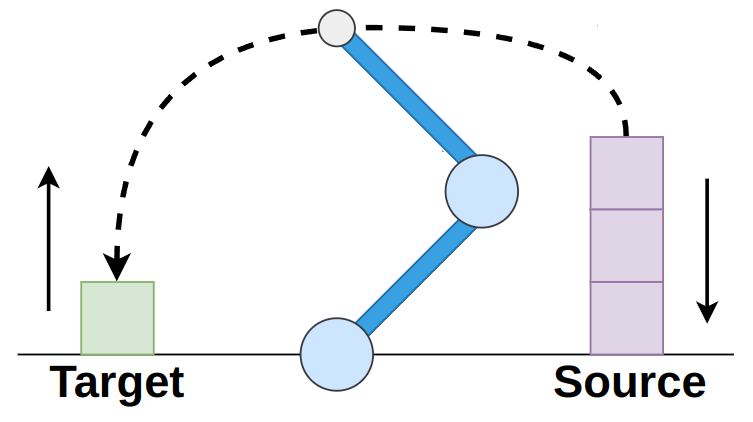}
    \caption{Pick-and-stack (opposite side) --- Variant 4}
    \label{fig:r4}
  \end{subfigure}

  \vspace{0.8em}

  Legend:
  {\small
  {\color{arsblue}\rule{0.9em}{0.9em}} ARS,
  {\color{bcgreen}\rule{0.9em}{0.9em}} BC,
  {\color{mamlorange}\rule{0.9em}{0.9em}} MAML,
  {\color{varibadbrown}\rule{0.9em}{0.9em}} VariBAD,
  {\color{genrlpurple}\rule{0.9em}{0.9em}} GenRL, and
  {\color{dipsred}\rule{0.9em}{0.9em}} \drill~(Ours).}

  \vspace{0.2em}

  \begin{subfigure}[t]{0.48\linewidth}
    \centering
    \includegraphics[width=\linewidth]{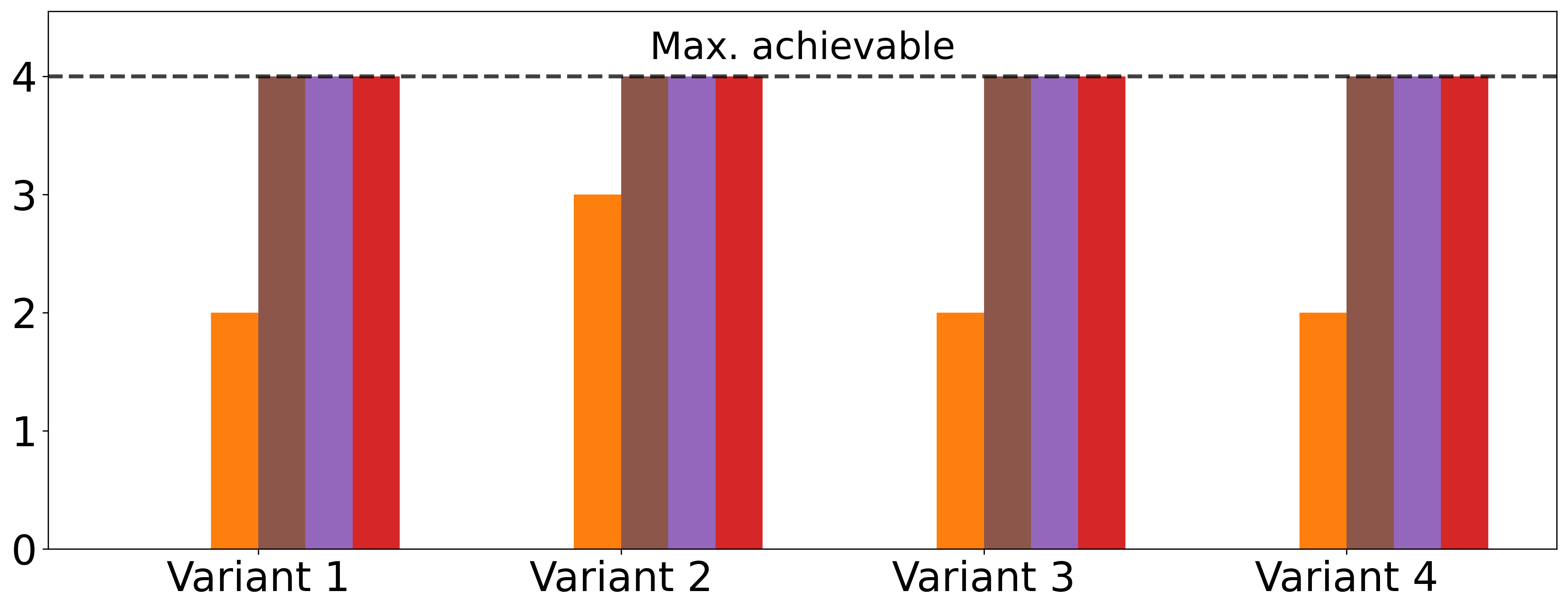}
    \caption{8 blocks}
    \label{fig:reacher_8_blocks}
  \end{subfigure}\hfill
  \begin{subfigure}[t]{0.48\linewidth}
    \centering
    \includegraphics[width=\linewidth]{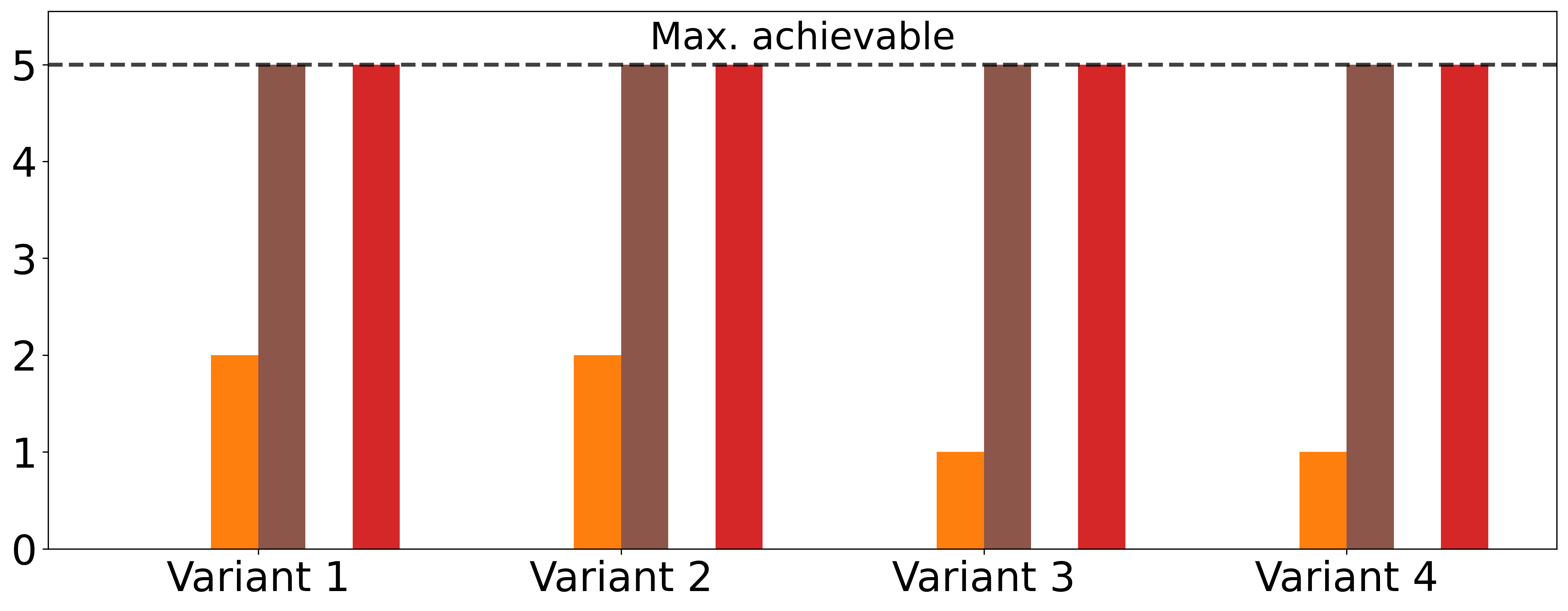}
    \caption{10 blocks}
    \label{fig:reacher_10_blocks}
  \end{subfigure}

  \caption{\textbf{\textsc{Reacher} task variants and zero-shot generalization results.}
  Top row: specification illustrations for the four \textsc{Reacher} tower pick-and-place variants. 
  Bottom row: absolute zero-shot generalization results for the 8-block and 10-block tower settings. 
  The $y$-axis in the bottom-row plots shows absolute zero-shot generalization, and the $x$-axis denotes the four Reacher variants shown in the top row.}
  \label{fig:reacher_all}
\end{figure*}

\noindent\textbf{Ablation Study: Decoupling teacher learning from $\kappa$-learning.} Since the $\kappa$ optimization is embedded within this ARS loop, \genrl~cannot
easily benefit from stronger single-task RL methods. Consequently, when ARS degrades, \genrl's template learning also breaks down
(Fig.~\ref{fig:reacher_all}).
For example, \textsc{Reacher} 
becomes harder as the number of displaced blocks grows due to
longer horizons and tighter manipulation. In such cases, ARS frequently fails to learn a strong controller, which impacts \genrl's $\kappa$-template learning due to poor reward signals and unstable $\kappa$ updates.

In contrast, \drill~separates teacher learning from template fitting. We train each
per-index teacher $\hat{\pi}_i$ with any suitable RL algorithm (e.g., PPO), then fit the
same template class via behavioral cloning on teacher-labeled datasets. Because template
optimization is independent of the teacher RL backbone, \drill~is capable of using stronger teachers and even domain-specific teachers out of the box, which directly
translates into better supervision, thereby enabling stable learning even in harder settings.
Empirically, this produces a usable template where \genrl~fails
(Fig.~\ref{fig:reacher_all}), and the same trend appears in
\textsc{SimpleDrone} and \textsc{DroneAttitude}
(Fig.~\ref{fig:r2a}--\ref{fig:r2c}).

\noindent\textbf{Ablation Study: Cross-index drift regularization.} We ablate the cross-index drift regularizer used in Stage~I teacher training on
\textsc{SimpleDrone}. Removing the regularizer ($\lambda_{\mathrm{x}}=0$) causes a sharp
drop in zero-shot generalization: the best run decreases from $26$ solved unseen tasks
(with regularization) to $3$ solved unseen tasks (without regularization). This
highlights that cross-index drift regularization is important for producing a teacher
sequence that can be fit smoothly by a single template, which in turn improves
generalization to unseen task indices.


\vspace{-0.4em}
\section{Conclusion}
\vspace{-0.4em}
We introduce a novel inductive 
generalization approach to RL that, by decoupling teacher policy learning from template 
training, (a) addresses a fundamental instability in prior work and (b) eliminates dependence on a specific RL algorithm, both of which limit scalability to larger task families. Experiments on complex temporal logic 
benchmarks confirm that our approach significantly outperforms existing methods 
on both training stability and zero-shot generalization. 

\subsection*{Acknowledgment} Research reported in this publication was partly supported by an Amazon Research Award, Fall 2024

\clearpage

\bibliographystyle{bib}
\bibliography{references}

\newpage
\clearpage
\appendix
\section*{Appendix}

\section{Limitations}
\label{sec:supp:limitations}

Our approach improves inductive generalization by learning a shared policy-evolution template, but it comes with a few limitations. First, the method still depends on having sufficiently strong per-index teachers; if a teacher fails on a hard index, the supervised signal used to fit the template can degrade. Second, as the number of training indices grows, the aggregated supervision pool grows and the template (and the policies it generates) may require higher-capacity networks to avoid underfitting, which increases optimization difficulty and can introduce a tradeoff between expressiveness and smooth template fitting. In addition, the added design components (cross-index regularization, candidate-state mixture, confidence threshold) introduce hyperparameters that may require tuning across environments.

\section{Societal Impact}
\label{sec:supp:impact}

This work can help make real-world robotics more efficient by reducing how often systems must be retrained from scratch when tasks change in a structured way, such as shifting goals, initial conditions, or obstacle layouts. By learning a reusable policy-evolution rule from a small set of trained task policies, the approach can cut repeated engineering effort and enable faster deployment across related task variants, which is practical for domains like navigation, manipulation, and autonomy. We also explicitly account for safety through reach-avoid task formulations, where success is defined jointly by reaching the goal and maintaining safety constraints. That said, any deployment in physical systems still requires careful validation under distribution shift and conservative safety checks, since real environments can differ from the assumed task family.

\section{SPECTRL Specification Language}
\label{ap:spectrl}

Let a subrollout of $\zeta$ be the subsequence
$\zeta_{\ell:k} = s_\ell\xrightarrow{a_\ell}\cdots\xrightarrow{a_{k-1}} s_k$.

For a finite rollout $\zeta$ of length $t$, satisfaction is defined as:
\begin{align*}
\zeta&\models\eventually{b} ~&&\text{if}~ \exists\ \tau \leq t,~s_\tau\models b \\
\zeta&\models \p \always{b} ~&&\text{if}~ \zeta\models\p ~\text{and}~ \forall\ \tau\leq t, ~ s_\tau\models b \\
\zeta&\models\p_1; \p_2 ~&&\text{if}~ \exists\ \tau < t, ~\zeta_{0:\tau}\models \p_1 ~\text{and}~ \zeta_{\tau+1:t}\models\p_2 \\
\zeta&\models\choice{\p_1}{\p_2} ~&&\text{if}~ \zeta\models\p_1 ~\text{or}~ \zeta\models\p_2.
\end{align*}

\section{More Algorithm Details}

The complete algorithm is given in Algorithm~\ref{algo:bc-template}.

\subsection{Template unrolling}
\label{sec:supp:index_unrolling}

\paragraph{Unrolling the $\kappa$-template across an index gap.}
Given a policy at index $i$, applying the $\kappa$-parameterized update once produces a
policy for the next index; applying it repeatedly produces a policy for a larger index
gap. We write this procedure as
\[
\texttt{UnrollPolicy}(\kappa,\pi,\Delta,m),
\]
which takes an input policy $\pi$, applies the $\kappa$-parameterized update $\Delta$
times (with degree $m$), and returns the resulting policy.

\paragraph{Unrolling across the training indices.}
Starting from the base policy $\pi_0$ at index $i_1=0$, we obtain template-generated
policies at the training indices by unrolling across successive gaps. Let
$\Train=\{i_1<\cdots<i_n\}$ with $i_1=0$. For each $k=2,\dots,n$, define the gap
$\Delta_k \triangleq i_k - i_{k-1}$ and set
\[
\pi_{i_k}
=
\texttt{UnrollPolicy}(\kappa,\pi_{i_{k-1}},\Delta_k,m).
\]
Thus, $\pi_{i_k}$ denotes the template-generated policy for the training task at index
$i_k$.

\section{Implementation Details}
\label{sec:supp:implementation}

\paragraph{Teacher training algorithms.}
For each training index $i\in\Train$, we train a task-specific teacher policy
$\hat{\pi}_i$ using standard RL on $\task_i$. We use ARS for \textsc{Car2D}, and PPO
for \textsc{Reacher}, \textsc{SimpleDrone}, and \textsc{DroneAttitude}. These
choices yield stable expert-level teachers for subsequent dataset labeling. To parameterize how \emph{densely} training indices cover the index line, we use a 
spacing parameter $\Delta\in\mathbb{N}$ between training indices to construct
$\Train=\{0,\Delta,2\Delta,\dots,L\}$ (truncated to lie in $\{0,\dots,L\}$). Larger $\Delta$
yields sparser supervision along the index family and typically makes zero-shot
generalization harder because neighboring indices are less constrained by training data.

\paragraph{Teacher and confidence-head architectures.}
We train a single neural network per index $i$ with a shared MLP trunk $h_i$ and two
heads: an action (teacher policy) head and a confidence head. For \textsc{Car2D} and
\textsc{Reacher}, the trunk is a 1-layer MLP with 8 hidden units. For the
higher-dimensional benchmarks, we scale capacity while keeping the architecture class
consistent: for \textsc{SimpleDrone} we use a 2-layer MLP with 64 hidden units per
layer, and for \textsc{DroneAttitude} we use a 2-layer MLP with 128 hidden units per
layer.

\begin{algorithm}[t]
\caption{\drill: Decoupled $\kappa$-learning via Behavioral Cloning}
\label{algo:bc-template}
\begin{algorithmic}[1]
\Require Task indices $\{0,\dots,L\}$; training index set $\Train=\{i_1<\cdots<i_n\}$ with $i_1=0$ and $i_n=L$;
base policy $\pi_0$; learning rate $\eta$; aggregation budget $M$; polynomial degree $m$;
teacher policies $\{\hat{\pi}_i\}_{i\in \Train}$;
preliminary datasets $\{\mathcal{D}_i\}_{i\in \Train}$;
confidence probability $\{p_i^{\mathrm{conf}}(\cdot)\}_{i\in \Train}$;
threshold $\tau_c$ and index gap $\Delta$.
\Statex

\State \textbf{Stage I: Teacher-labeled, confidence-filtered dataset creation}
\Statex \Comment{Assume teachers $\{\hat{\pi}_i\}$ are trained using Section~\ref{alg:teacher_loss}, and preliminary datasets $\{\mathcal{D}_i\}$ are collected as in Section~\ref{alg:csa}.}
\Statex
\ForAll{$i \in \Train$}
    \State $\widetilde{\mathcal D}_i \gets \emptyset$
    \While{$|\widetilde{\mathcal D}_i| < M$}
        \State Sample $s \sim \mathcal{D}_i$
        \If{$p_i^{\mathrm{conf}}(s)\ge \tau_c$} \Comment{retain high-quality states}
            \State $a \gets \hat{\pi}_i(s)$
            \State $\widetilde{\mathcal D}_i \gets \widetilde{\mathcal D}_i \cup \{(s,a)\}$
        \EndIf
    \EndWhile
\EndFor
\Statex

\State \textbf{Stage II: Fit $\kappa$ by Behavioral Cloning}
\State Initialize $\kappa$ (degree $m$)
\Repeat
    \State $\pi_{i_1} \gets \pi_0$ \Comment{$i_1=0$}
    \For{$k=2$ to $n$}
        \State Unroll $\kappa$ from $\pi_{i_{k-1}}$, $\Delta$ times to obtain $\pi_{i_k}$
    \EndFor

    \State $\mathcal L_{\mathrm{BC}} \gets 0$
    \For{$k=1$ to $n$}
        \State $\mathcal L_{\mathrm{BC}} \gets \mathcal L_{\mathrm{BC}}
        + \frac{1}{|\widetilde{\mathcal D}_{i_k}|}\sum_{(s,a)\in \widetilde{\mathcal D}_{i_k}}
        \|\pi_{i_k}(s)-a\|_2^2$
    \EndFor

    \State $\kappa \gets \kappa - \eta\,\nabla_{\kappa}\mathcal L_{\mathrm{BC}}$
\Until{$\mathcal L_{\mathrm{BC}}$ is below a threshold or stagnates}
\State \Return $\kappa$
\end{algorithmic}
\end{algorithm}

\paragraph{Hyperparameter selection (grid sweeps).}
We choose key hyperparameters by sweeping a small grid on the training indices in
$\Train$ and selecting values that yield (i) high teacher specification satisfaction and
(ii) stable Stage~2 template fitting. In particular, we sweep the candidate-pool sampling
rates $(\alpha,\beta,\gamma)$ used to build $\mathcal{D}_i$, the confidence threshold
$\tau_c$ used to filter candidates, and the confidence-loss weight
$\lambda_{\mathrm{conf}}$ used in the joint objective
$\mathcal{L}^{(i)}_{\mathrm{teacher-conf}}=\mathcal{L}^{(i)}_{\mathrm{teacher}}+
\lambda_{\mathrm{conf}}\mathcal{L}^{(i)}_{\mathrm{conf}}$. Concretely, we sweep
$\alpha\in\{0.6,0.7,0.8,0.9,1.0\}$, $\beta\in\{0.1,0.3,0.5,0.7,0.9\}$, and
$\gamma\in\{0.6,0.7,0.8,0.9,1.0\}$, along with $\tau_c\in\{0.7,0.8,0.9\}$ and
$\lambda_{\mathrm{conf}}\in\{0.1, 0.3,0.5, 0.7, 0.9\}$. We select the configuration that
produces strong teachers and reliable filtering while retaining sufficient coverage for
Stage~2. We similarly choose the cross-index regularization weight by sweeping
$\lambda_{\mathrm{x}}\in\{0,10^{-4},10^{-3},10^{-2}\}$ and selecting the largest value
that reduces cross-index teacher drift without degrading teacher specification
satisfaction.

\paragraph{Dataset aggregation settings.}
For each $i\in\Train$, we construct a preliminary candidate set $\mathcal{D}_i$ by
sampling states from the converged teacher’s on-policy rollouts, the teacher replay
buffer, and environment resets using the sampling rates $(\alpha,\beta,\gamma)$. We then
keep a candidate state $s\in\mathcal{D}_i$ only if $p_i^{\mathrm{conf}}(s)\ge \tau_c$,
query the teacher for an action label $a=\hat{\pi}_i(s)$, and add $(s,a)$ to the final
dataset $\widetilde{\mathcal{D}}_i$ until reaching an aggregation budget of $M$ labeled
state--action pairs per training index. We select $M$ via a sweep
$M\in\{5\times 10^{2},\,5\times 10^{3},\,5\times 10^{4},\,5\times 10^{5},\,5\times 10^{6}\}$.

\paragraph{Stage~2 behavioral cloning settings.}
We fit the template coefficients $\kappa$ using the squared-error BC loss over
$\bigcup_{i\in\Train}\widetilde{\mathcal{D}}_i$ with a first-order optimizer. We use Adam
with an annealed learning rate initialized at $\eta=1\times 10^{-1}$, batch size 512,
and train until the BC loss stagnates. At each outer iteration, we unroll the current
$\kappa$ from the base policy $\pi_0$ to obtain the template-generated policies
$\{\pi_i\}_{i\in\Train}$ used to evaluate $\mathcal{L}_{\mathrm{BC}}$.

\paragraph{Training compute.}
All experiments were run on a SLURM cluster with Intel Xeon Gold 6226 CPUs
(2.7\,GHz, 24 cores per node) and 192\,GB RAM per node. The codebase is available at
https://anonymous.4open.science/r/Imitation-Kappa-2094/.

\section{Environment Description}
\label{sec:environments}

\begin{figure}[t]
    \centering
    \begin{subfigure}{0.29\linewidth}
        \centering
        \includegraphics[width=\linewidth]{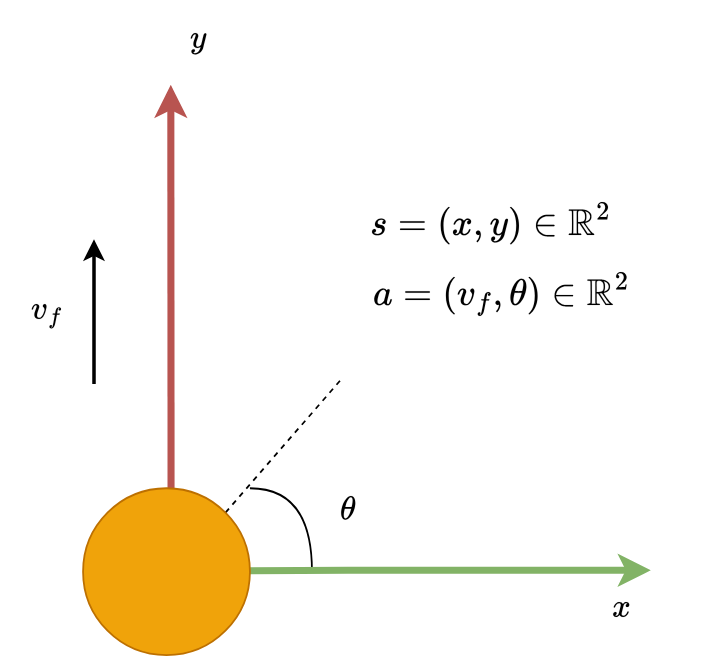}
        \caption{\textsc{Car2D}}
        \label{fig:env_car2d}
    \end{subfigure}
    \begin{subfigure}{0.29\linewidth}
        \centering
        \includegraphics[width=\linewidth]{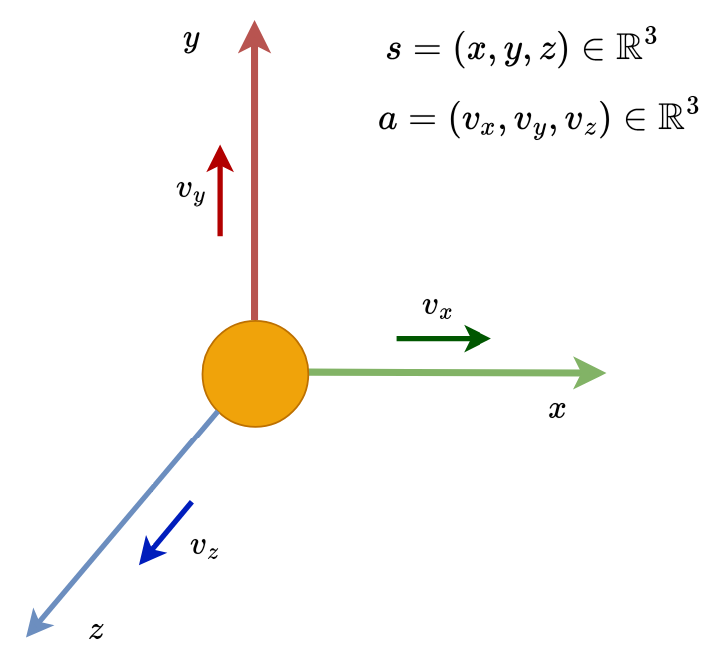}
        \caption{\textsc{SimpleDrone}}
        \label{fig:env_simple_drone}
    \end{subfigure}
    \begin{subfigure}{0.28\linewidth}
        \centering
        \includegraphics[width=\linewidth]{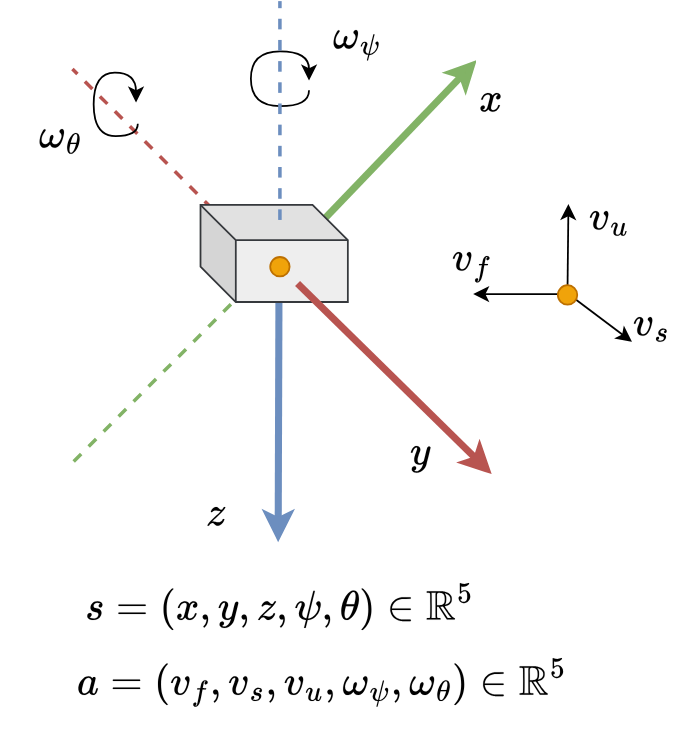}
        \caption{\textsc{DroneAttitude}}
        \label{fig:env_complex_drone}
    \end{subfigure}

    \caption{Agent models and their dynamics.}
    \label{fig:envs}
\end{figure}

Illustrations of the environments are given in Figure~\ref{fig:envs}.

\section{Specifications}
\label{sec:supp:specs}

\subsection{\textsc{$n$-Reachability} (Figure~\ref{fig:spec_nreach})}

\[
\begin{array}{c}
\eventually\big(\reach(g_1)\big)\ ;\ \eventually\big(\reach(g_2)\big)\ ;\ \\\cdots\ ;\ \eventually\big(\reach(g_n)\big)
\end{array}
\]
This specification requires the agent to reach a sequence of $n$ target regions in a
fixed order. The operator $\eventually(\cdot)$ means the corresponding region must be
reached at some point in the rollout, and the sequential composition ``$;$'' enforces
ordering, i.e., the agent must reach $g_1$ first, then (after that) reach $g_2$, and so
on, until it eventually reaches $g_n$.

\begin{figure}[t]
    \centering

    \begin{subfigure}[t]{0.48\linewidth}
        \centering
        \includegraphics[width=\linewidth]{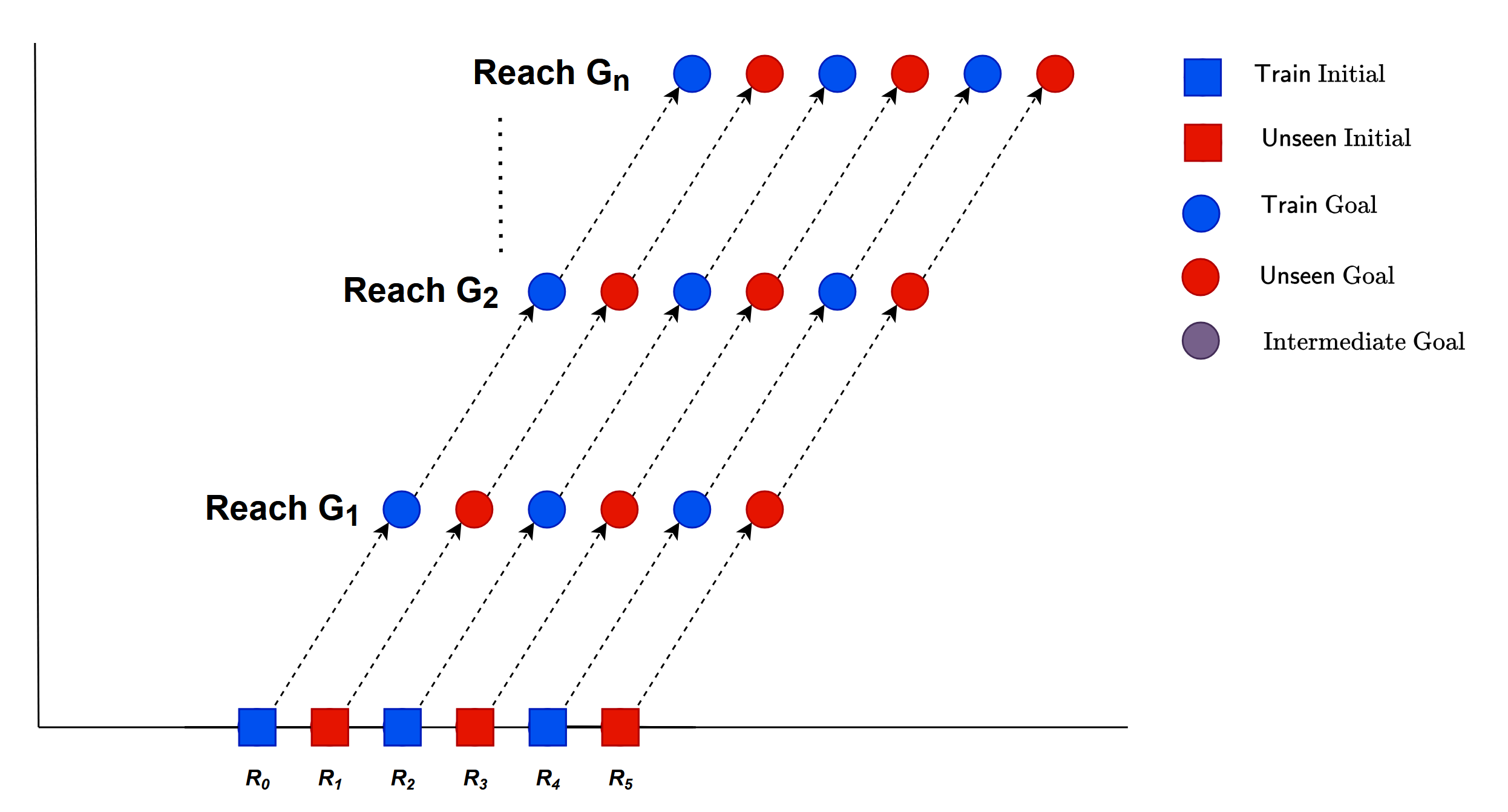}
        \caption{\textsc{$n$-Reachability} specification illustration.}
        \label{fig:spec_nreach}
    \end{subfigure}
    \hfill
    \begin{subfigure}[t]{0.48\linewidth}
        \centering
        \includegraphics[width=\linewidth]{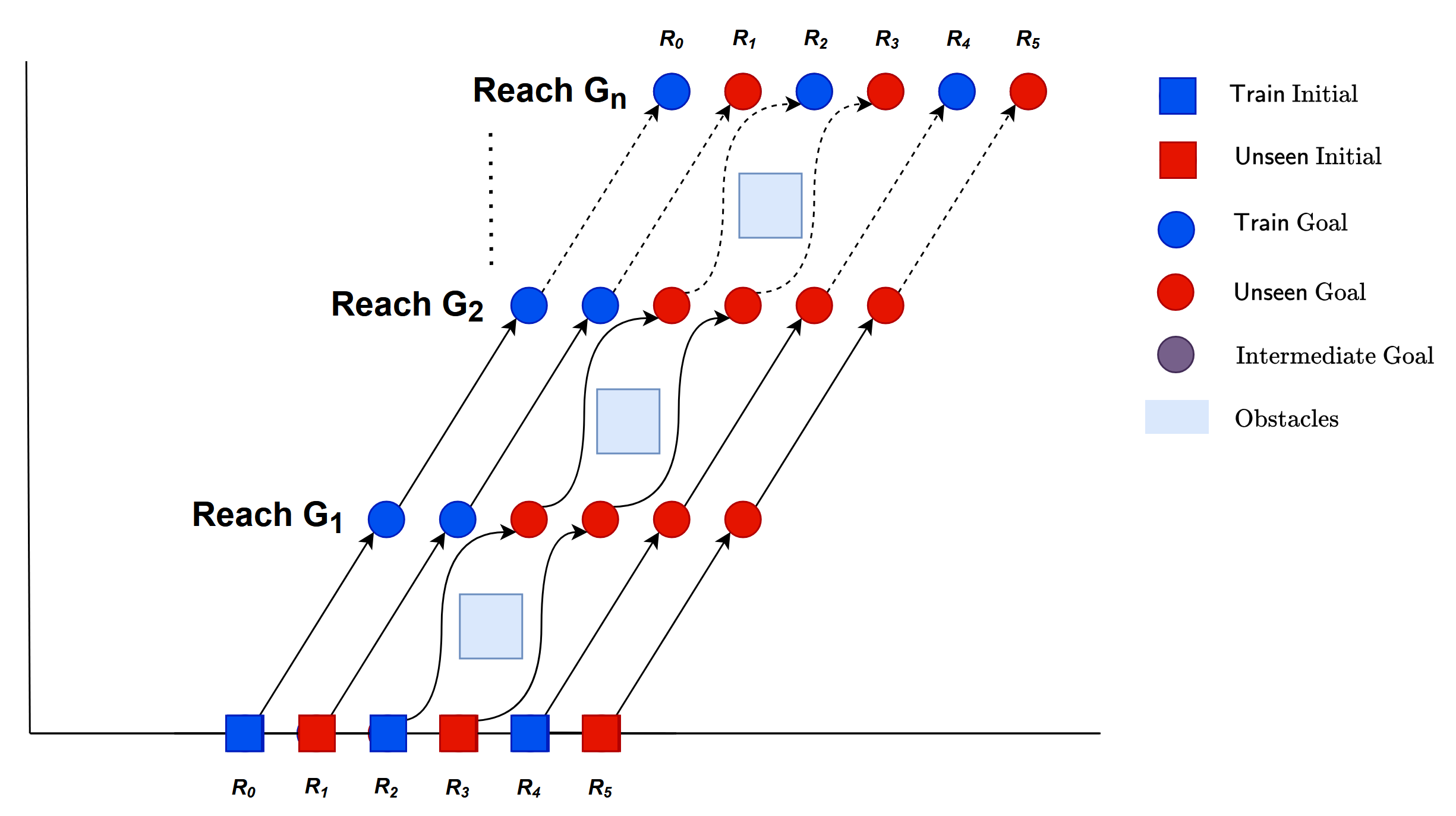}
        \caption{\textsc{$n$-Reachability+Obs} specification illustration.}
        \label{fig:spec_nreach_obs}
    \end{subfigure}

    \caption{Specification illustrations for \textsc{$n$-Reachability} and \textsc{$n$-Reachability+Obs}.}
    \label{fig:specs_nreach_combined}
\end{figure}

\subsection{\textsc{$n$-Reachability+Obs} (Figure~\ref{fig:spec_nreach_obs})}

\[
\begin{array}{c}
\eventually\big(\reach(g_1)\big)\ ;\ \eventually\big(\reach(g_2)\big)\ ;\ \\\cdots\ ;\ \eventually\big(\reach(g_n)\big)\\[0.25em]
\always\big(\avoid(obs)\big)
\end{array}
\]
This specification adds a safety requirement to \textsc{$n$-Reachability}. The first line
is the same ordered sequence of reach goals. The second line, $\always(\avoid(obs))$,
requires that the agent avoid the obstacle region $obs$ at \emph{all} timesteps during
execution.

\begin{figure}[t]
  \centering

  \begin{subfigure}[t]{0.48\linewidth}
    \centering
    \includegraphics[width=\linewidth]{images/specs/choice_1level.png}
    \caption{\textsc{Choice(l)} specification illustration where $l=1$}
    \label{fig:spec_choice_1}
  \end{subfigure}
  \hfill
  \begin{subfigure}[t]{0.48\linewidth}
    \centering
    \includegraphics[width=\linewidth]{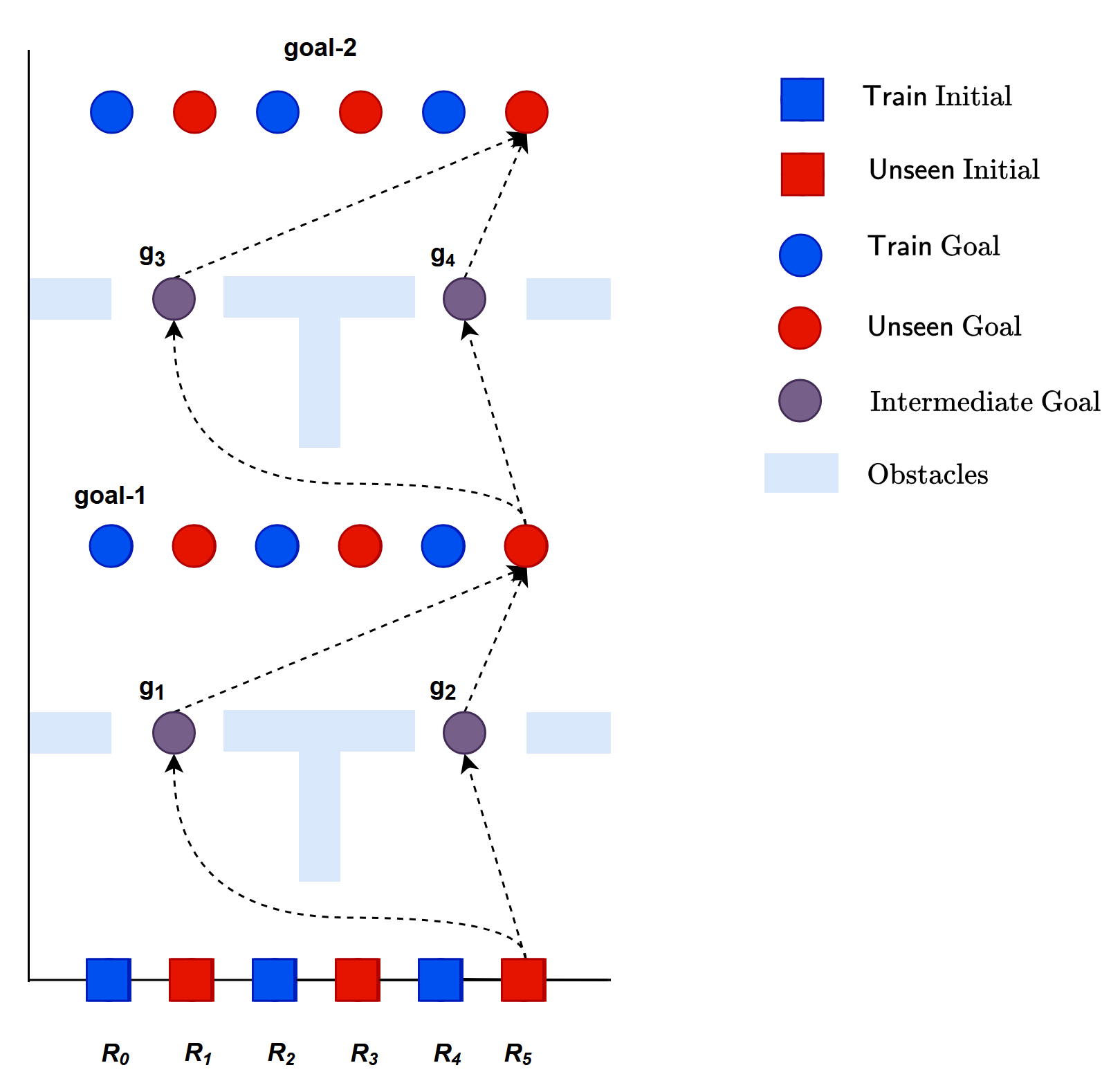}
    \caption{\textsc{Choice(l)} specification illustration where $l=2$}
    \label{fig:spec_choice_2}
  \end{subfigure}

  \caption{\textsc{Choice(l)} specification illustrations for two levels.}
  \label{fig:spec_choice}
\end{figure}

\subsection{\textsc{Choice}$(l)$ (Figure~\ref{fig:spec_choice_2})}

\[
\begin{array}{c}
\left(
\begin{array}{c}
\eventually\big(\reach(g_{i1}) \ \lor\ \reach(g_{i2})\big)\ ;\\[0.25em]
\eventually\big(\reach(goal_i)\big)
\end{array}
\right)^{l}\\[0.25em]
\always\big(\avoid(obs)\big)
\end{array}
\]
This specification stacks $l$ levels of a branching subtask. At each level $i$, the
agent must first reach \emph{either} $g_{i1}$ or $g_{i2}$ (the disjunction $\lor$), and
then reach the corresponding level goal $goal_i$. $(\cdot)^l$ indicates that this
two-step pattern is repeated for $i=1,\dots,l$ in sequence. Finally,
$\always(\avoid(obs))$ enforces obstacle avoidance throughout the entire rollout.

\subsection{\textsc{Reacher} Specifications}
\label{sec:supp:reacher_specs}


Each task is a sequential reach specification $\eventually(\reach(\cdot));\eventually(\reach(\cdot))$:
in which the arm first reaches a designated ``pick'' region (typically the top block of the source tower),
then reaches a designated ``place'' region (drop box or target tower).

\paragraph{Pick and drop (same side; Fig.~\ref{fig:r1}).}
Pick the top block from the source tower, then place it in the target drop box.
\[
\begin{array}{c}
\eventually\big(\reach(\text{top block of source tower})\big)\ ;\\[0.25em]
\eventually\big(\reach(\text{target drop box})\big)
\end{array}
\]

\paragraph{Pick and drop (opposite side; Fig.~\ref{fig:r3}).}
Pick the top block from the source tower, then place it in the target drop box on the opposite side.
\[
\begin{array}{c}
\eventually\big(\reach(\text{top block of source tower})\big)\ ;\\[0.25em]
\eventually\big(\reach(\text{target drop box})\big)
\end{array}
\]

\paragraph{Pick and vertical stack (same side; Fig.~\ref{fig:r2}).}
Pick the top block from the source tower, then stack it onto the top of the target tower.
\[
\begin{array}{c}
\eventually\big(\reach(\text{top block of source tower})\big)\ ;\\[0.25em]
\eventually\big(\reach(\text{top block of target tower})\big)
\end{array}
\]

\paragraph{Pick and vertical stack (opposite side; Fig.~\ref{fig:r4}).}
Pick the top block from the source tower, then stack it onto the top of the target tower on the opposite side.
\[
\begin{array}{c}
\eventually\big(\reach(\text{top block of source tower})\big)\ ;\\[0.25em]
\eventually\big(\reach(\text{top block of target tower})\big)
\end{array}
\]

\section{Additional Results}
\label{app:more-results}

\begin{table}[t]
\centering
\small
\setlength{\tabcolsep}{10pt}
\renewcommand{\arraystretch}{1.08}
\begin{tabular}{|l|c|c|}
\hline
\textbf{Baseline} & \textbf{Training tasks} & \textbf{Unseen tasks} \\
\hline
ARS      & --             & --             \\
BC       & $3.94\times$   & --             \\
MAML     & $10.83\times$  & $14.04\times$  \\
VariBAD  & $3.02\times$   & $3.04\times$   \\
GenRL    & $3.82\times$   & $6.19\times$   \\
\hline
\end{tabular}

\vspace{0.4em}

\caption{\textbf{Relative aggregate performance of \drill.}
Ratios compare the aggregate number of tasks across all benchmarks solved by \drill~against each baseline, using mean performance over \textbf{ten} random seeds. A dash indicates that the baseline solves zero tasks in the aggregate, so the ratio is undefined.}
\label{tab:relative-performance}
\end{table}

Table~\ref{tab:relative-performance} summarizes the aggregate advantage of \drill~over each baseline across all benchmarks. For each baseline, we report the factor by which \drill~solves more tasks, separately for training tasks and unseen tasks, using mean performance over ten random seeds. Overall, \drill~achieves substantially higher training scalability than all baselines. On unseen tasks, \drill~also shows strong zero-shot generalization gains, solving $6.19\times$ as many tasks as \genrl, $3.04\times$ as many as \textsc{VariBAD}, and $14.04\times$ as many as MAML. The dash entries indicate cases where the corresponding baseline solves zero tasks in the aggregate, so the ratio is undefined.

\subsection{Additional $k$-reachability results}

\begin{figure*}[t]
\centering

\setlength{\abovecaptionskip}{2pt}
\setlength{\belowcaptionskip}{2pt}

{\centering
Legend:
{\small {\color{arsblue}\rule[0.6ex]{1.2em}{1pt}$\,\bullet$} ARS,
{\color{bcgreen}\rule[0.6ex]{1.2em}{1pt}$\,\blacksquare$} BC,
{\color{mamlorange}\rule[0.6ex]{1.2em}{1pt}$\,\blacklozenge$} MAML,
{\color{varibadbrown}\rule[0.6ex]{1.2em}{1pt}$\,\blacktriangle$} VariBAD,
{\color{genrlpurple}\rule[0.6ex]{1.2em}{1pt}$\,\blacktriangledown$} GenRL, and
{\color{dipsred}\rule[0.6ex]{1.2em}{1pt}{\large $\times$}} \drill~(Ours).}
\vspace{0.2em}
\small\textbf{\textsc{Car2D} $k$-Reachability + obstacles}\par}

\begin{subfigure}[t]{0.3\textwidth}
  \centering
  \includegraphics[width=\linewidth]{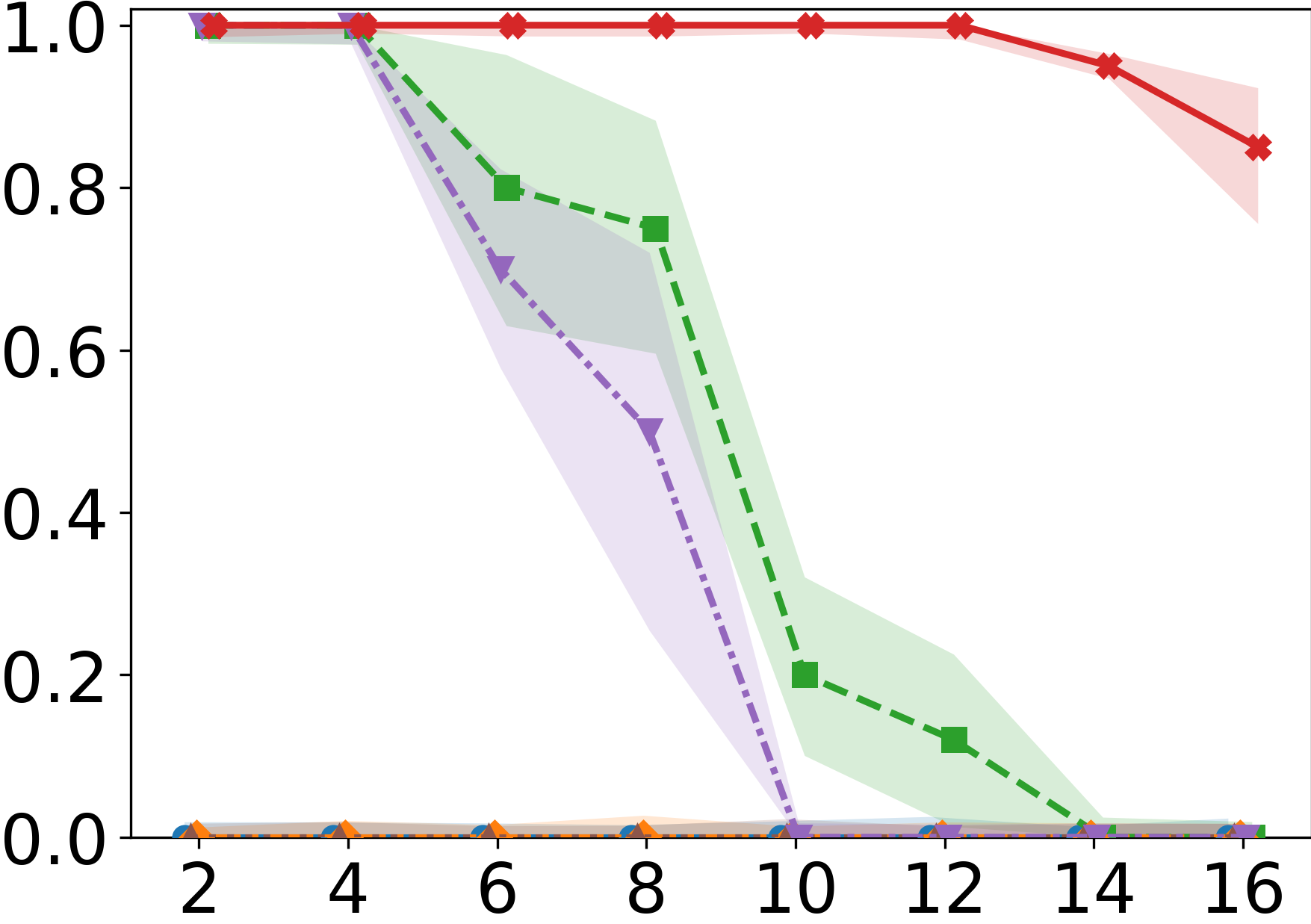}
  \caption{}\label{fig:obsa}
\end{subfigure}
\begin{subfigure}[t]{0.3\textwidth}
  \centering
  \includegraphics[width=\linewidth]{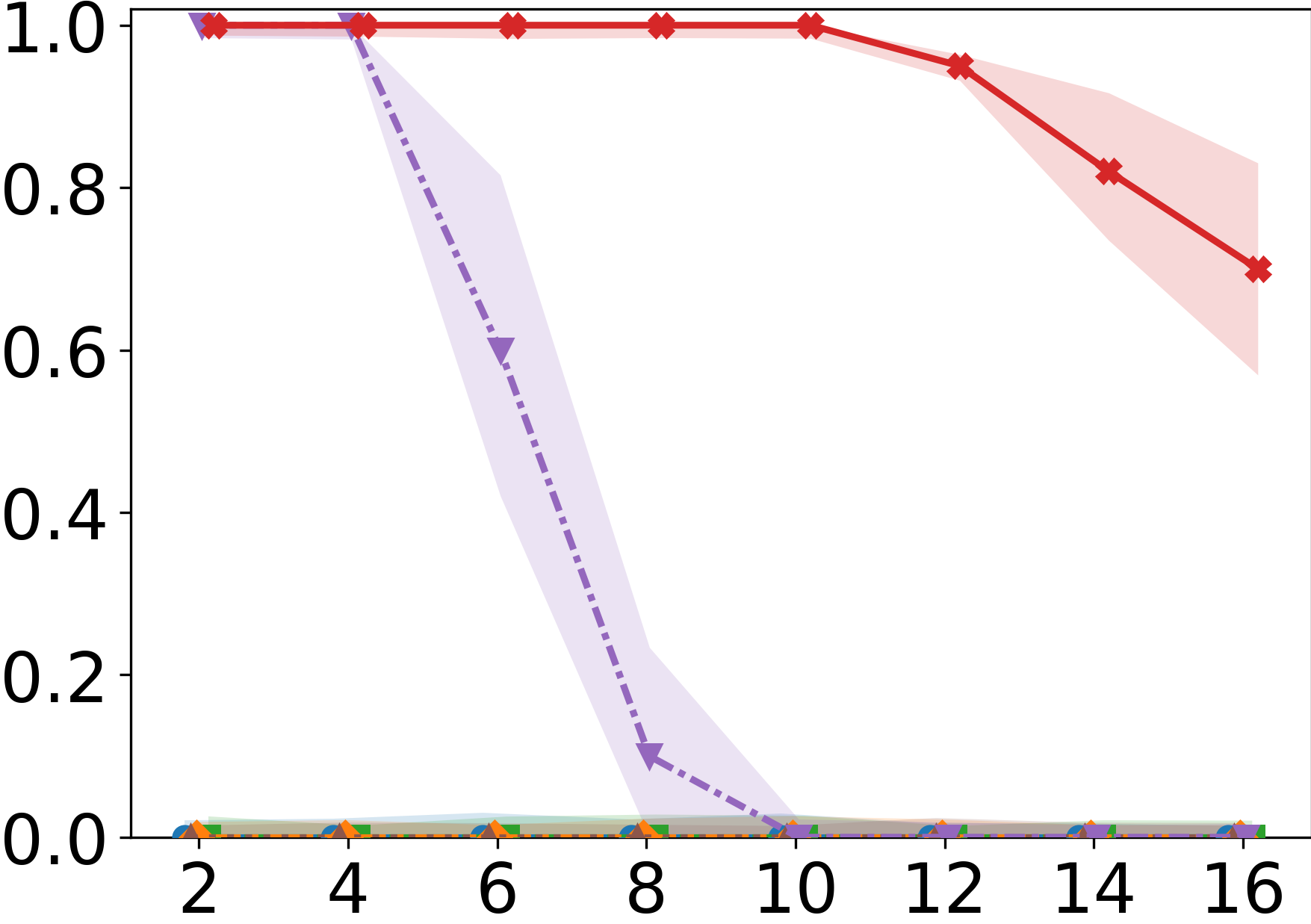}
  \caption{}\label{fig:obsb}
\end{subfigure}
\begin{subfigure}[t]{0.32\textwidth}
  \centering
  \includegraphics[width=\linewidth]{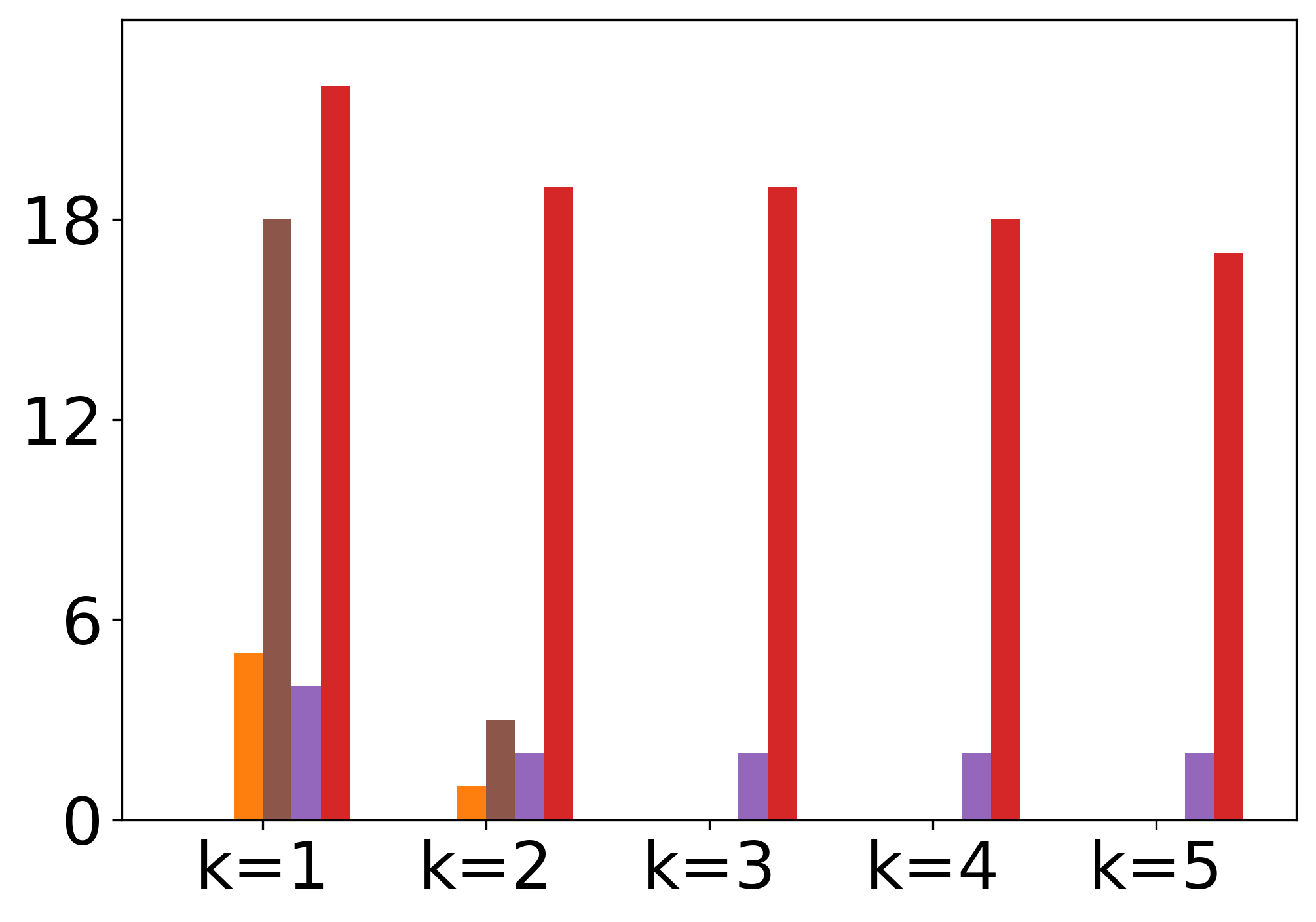}
  \caption{}\label{fig:obsc}
\end{subfigure}

\caption{\textbf{Scalability plots for \textsc{Car2D} $k$-Reachability with obstacles.}
\textbf{Column (a)} shows the successful training ratio ($y$-axis) versus the number of training tasks $|\Train|$ ($x$-axis) for $k=5$. 
\textbf{Column (b)} shows the zero-shot generalization ratio ($y$-axis) versus the number of training tasks $|\Train|$ ($x$-axis) for $k=5$. 
\textbf{Column (c)} shows absolute zero-shot generalization ($y$-axis) for $k \in \{1,\ldots,5\}$ ($x$-axis). 
We report mean ($\pm$ standard deviation) performance across \textbf{ten} random seeds.}

\label{fig:successvstasks_obstacles}
\end{figure*}

\begin{figure*}[t]
\centering
Legend:
{\small {\color{arsblue}\rule[0.6ex]{1.2em}{1pt}$\,\bullet$} ARS,
{\color{bcgreen}\rule[0.6ex]{1.2em}{1pt}$\,\blacksquare$} BC,
{\color{mamlorange}\rule[0.6ex]{1.2em}{1pt}$\,\blacklozenge$} MAML,
{\color{varibadbrown}\rule[0.6ex]{1.2em}{1pt}$\,\blacktriangle$} VariBAD,
{\color{genrlpurple}\rule[0.6ex]{1.2em}{1pt}$\,\blacktriangledown$} GenRL, and
{\color{dipsred}\rule[0.6ex]{1.2em}{1pt}{\large $\times$}} \drill~(Ours).}

\setlength{\abovecaptionskip}{2pt}
\setlength{\belowcaptionskip}{2pt}

\setcounter{subfigure}{0}
\par\medskip
{\centering\small\textbf{(i) \textsc{Car2D} $k$-Reachability}\par}
\vspace{0.2em}

\begin{subfigure}[t]{0.38\textwidth}
  \centering
  \includegraphics[width=\linewidth]{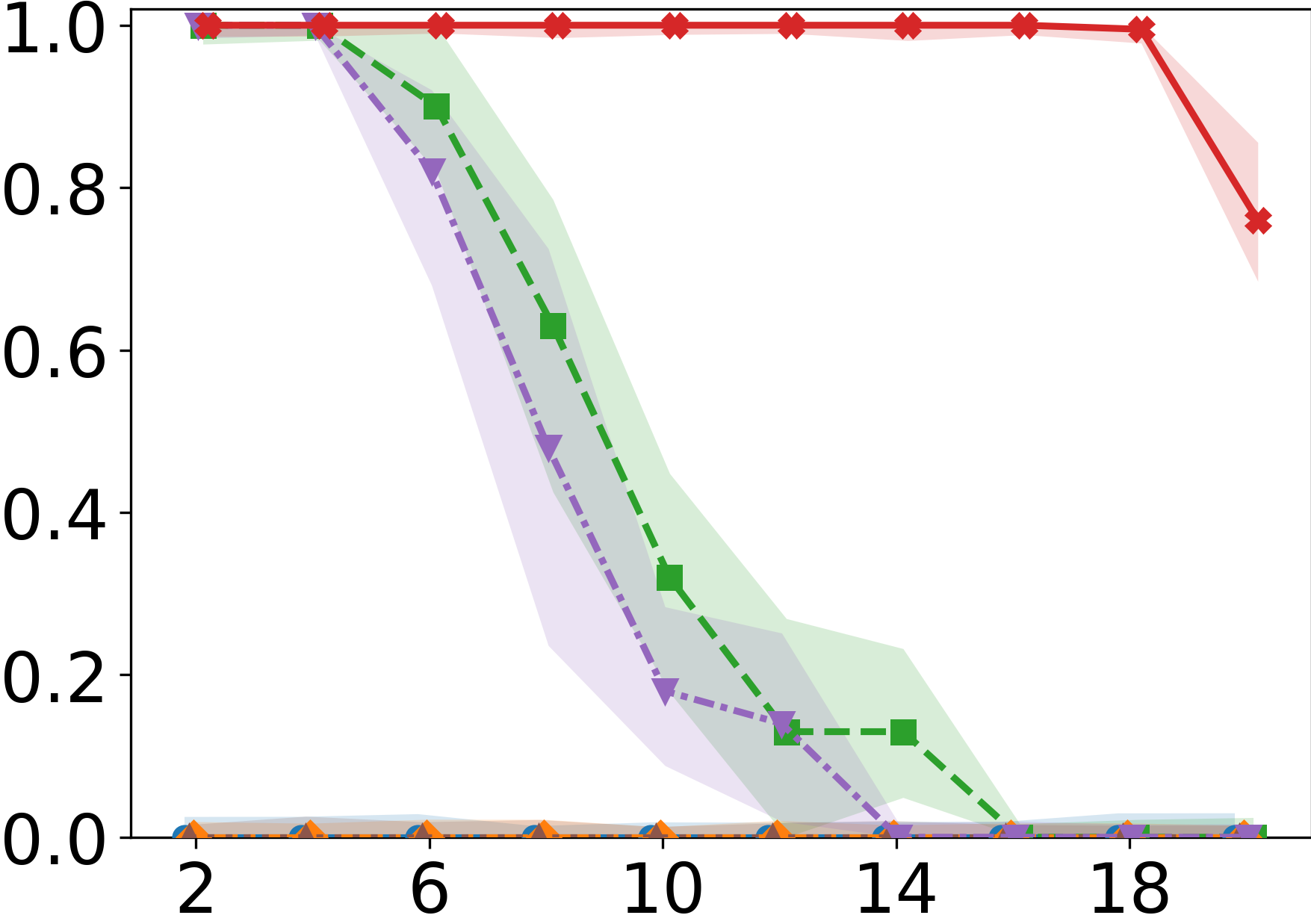}
  \caption{}\label{fig:k4_r1a}
\end{subfigure}\hspace{0.1\textwidth}
\begin{subfigure}[t]{0.38\textwidth}
  \centering
  \includegraphics[width=\linewidth]{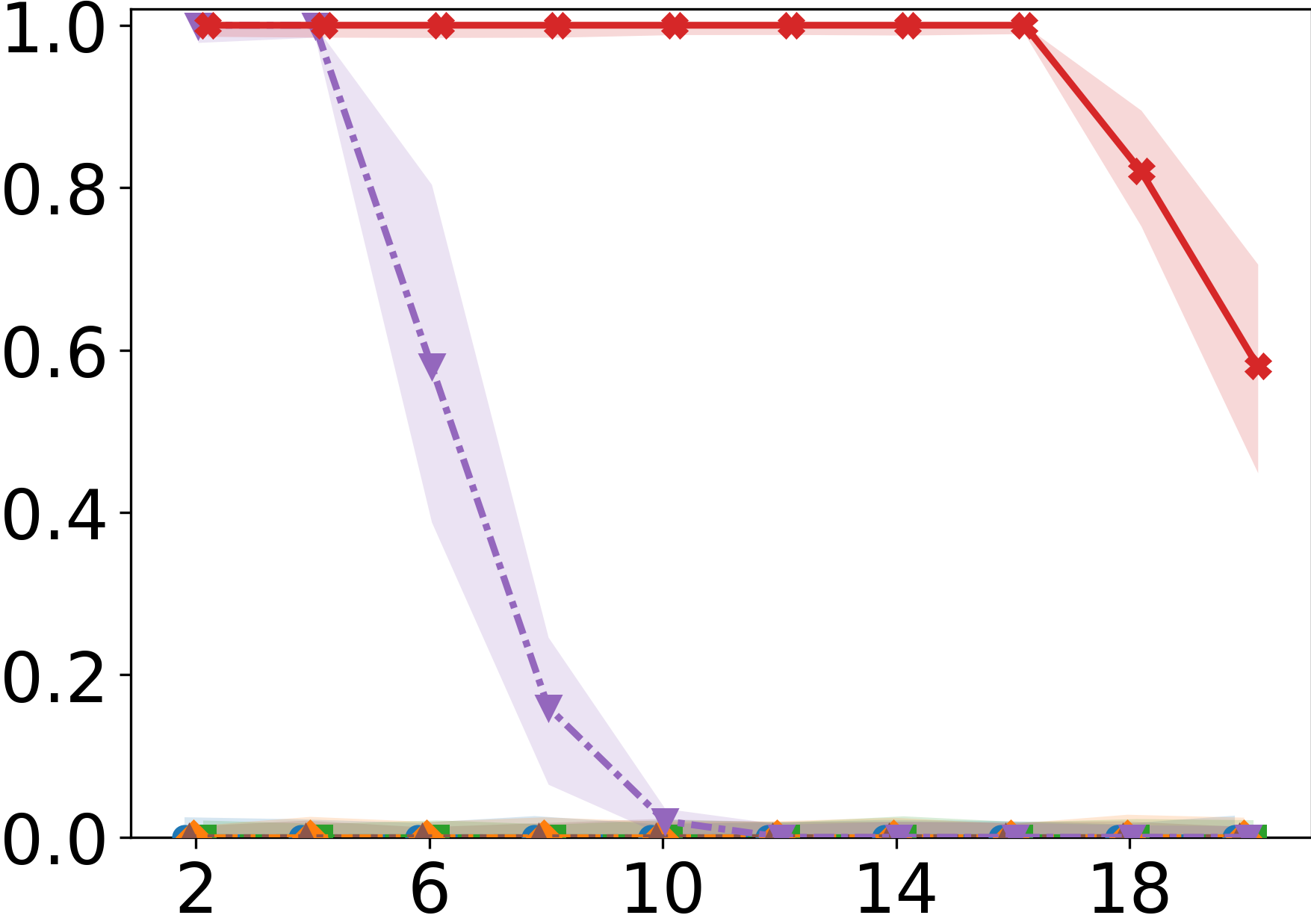}
  \caption{}\label{fig:k4_r1b}
\end{subfigure}

\setcounter{subfigure}{0}
\par\medskip
{\centering\small\textbf{(ii) \textsc{SimpleDrone} $k$-Reachability}\par}
\vspace{0.2em}

\begin{subfigure}[t]{0.38\textwidth}
  \centering
  \includegraphics[width=\linewidth]{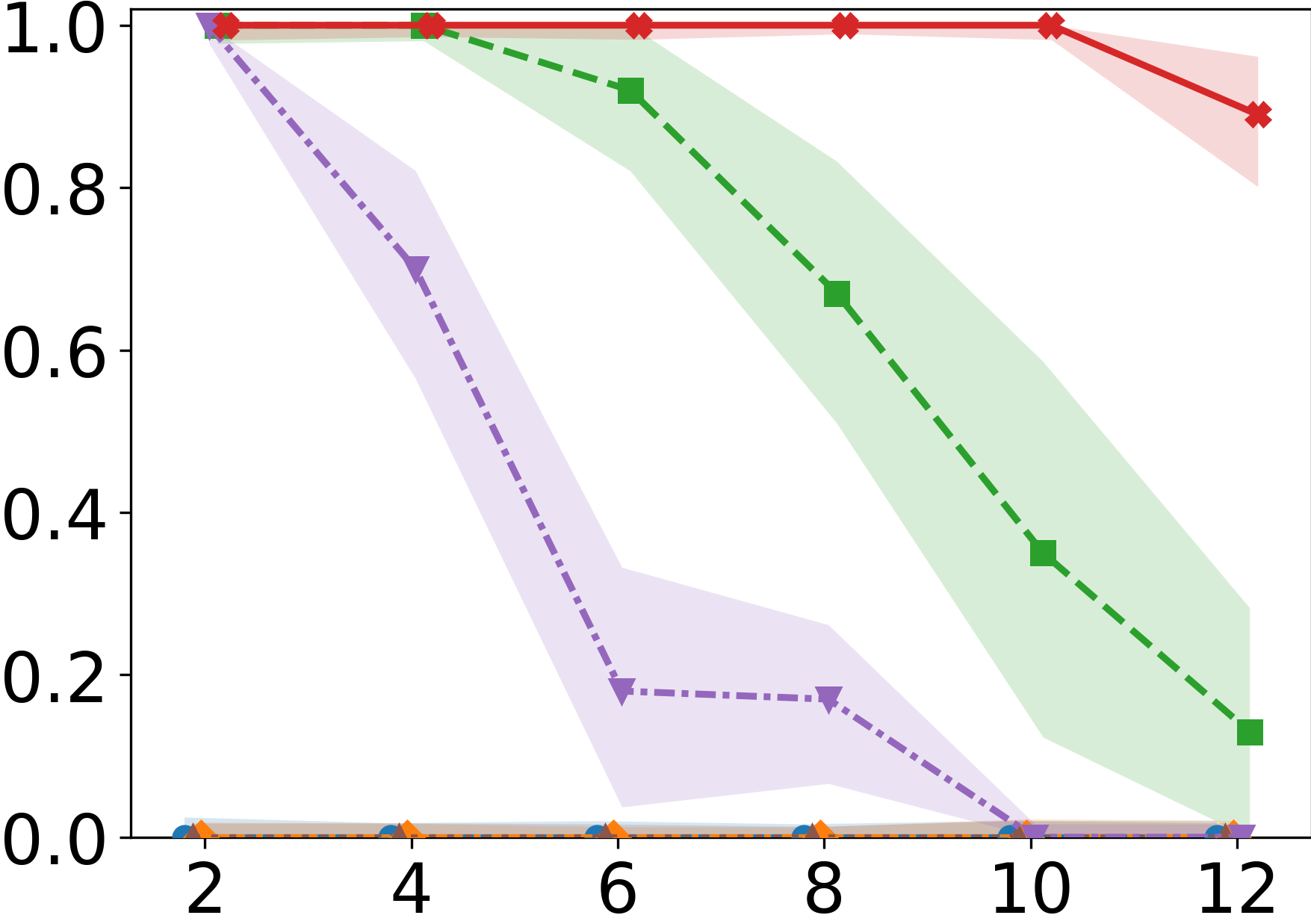}
  \caption{}\label{fig:k4_r2a}
\end{subfigure}\hspace{0.1\textwidth}
\begin{subfigure}[t]{0.38\textwidth}
  \centering
  \includegraphics[width=\linewidth]{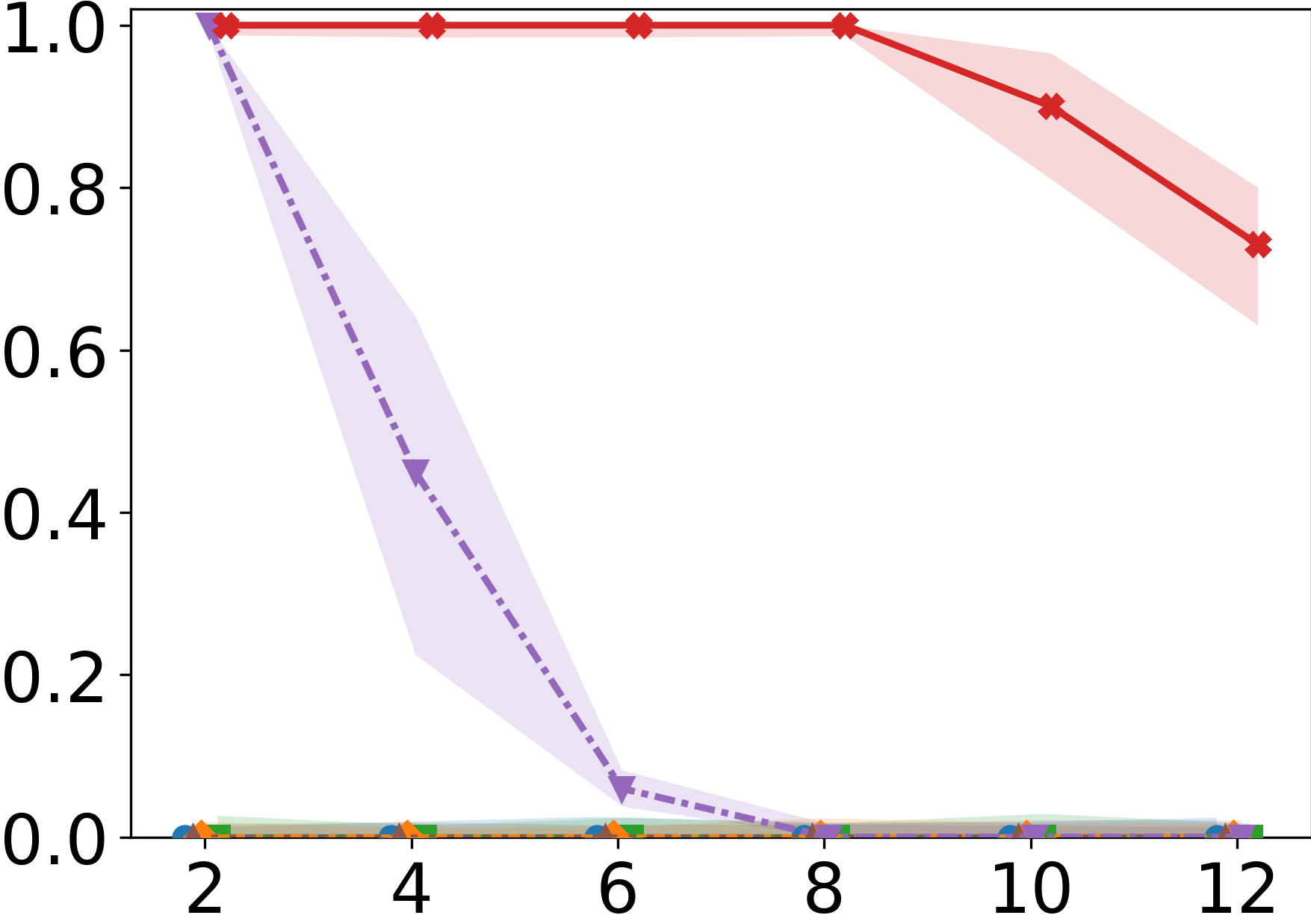}
  \caption{}\label{fig:k4_r2b}
\end{subfigure}

\setcounter{subfigure}{0}
\par\medskip
{\centering\small\textbf{(iii) \textsc{DroneAttitude} $k$-Reachability}\par}
\vspace{0.2em}

\begin{subfigure}[t]{0.38\textwidth}
  \centering
  \includegraphics[width=\linewidth]{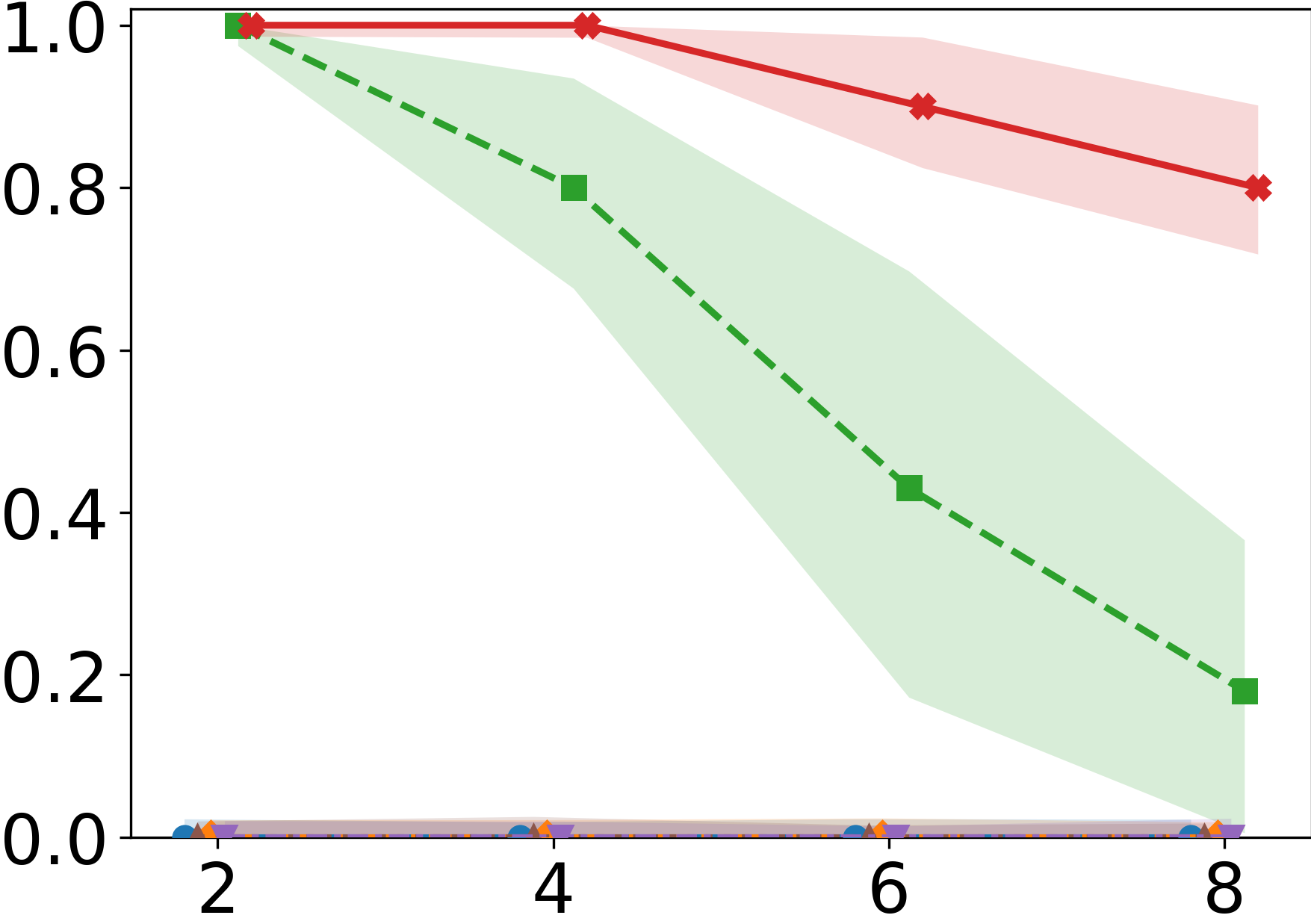}
  \caption{}\label{fig:k4_r3a}
\end{subfigure}\hspace{0.1\textwidth}
\begin{subfigure}[t]{0.38\textwidth}
  \centering
  \includegraphics[width=\linewidth]{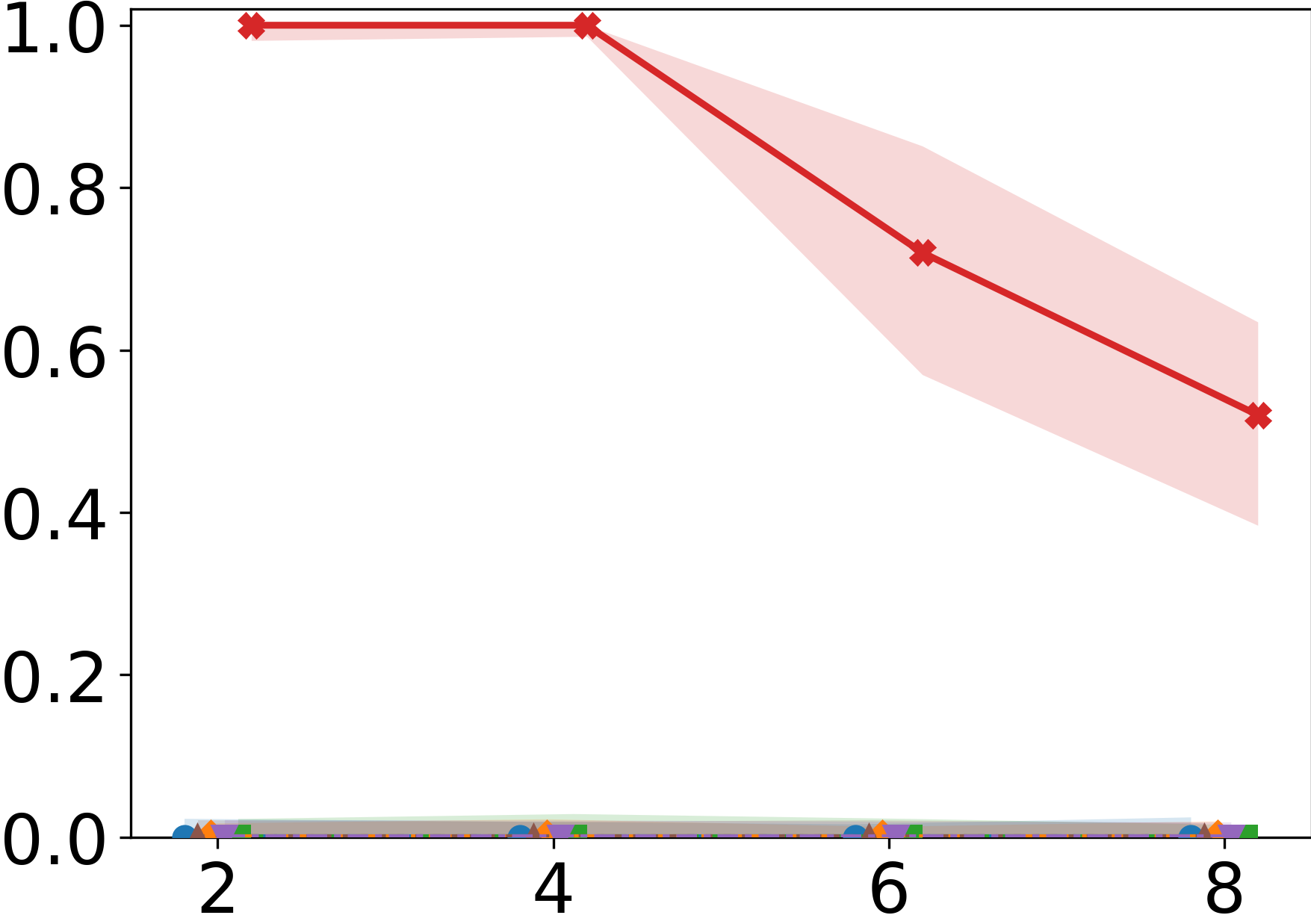}
  \caption{}\label{fig:k4_r3b}
\end{subfigure}

\setcounter{subfigure}{0}
\par\medskip
{\centering\small\textbf{(iv) \textsc{Car2D} $k$-Reachability + obstacles}\par}
\vspace{0.2em}

\begin{subfigure}[t]{0.38\textwidth}
  \centering
  \includegraphics[width=\linewidth]{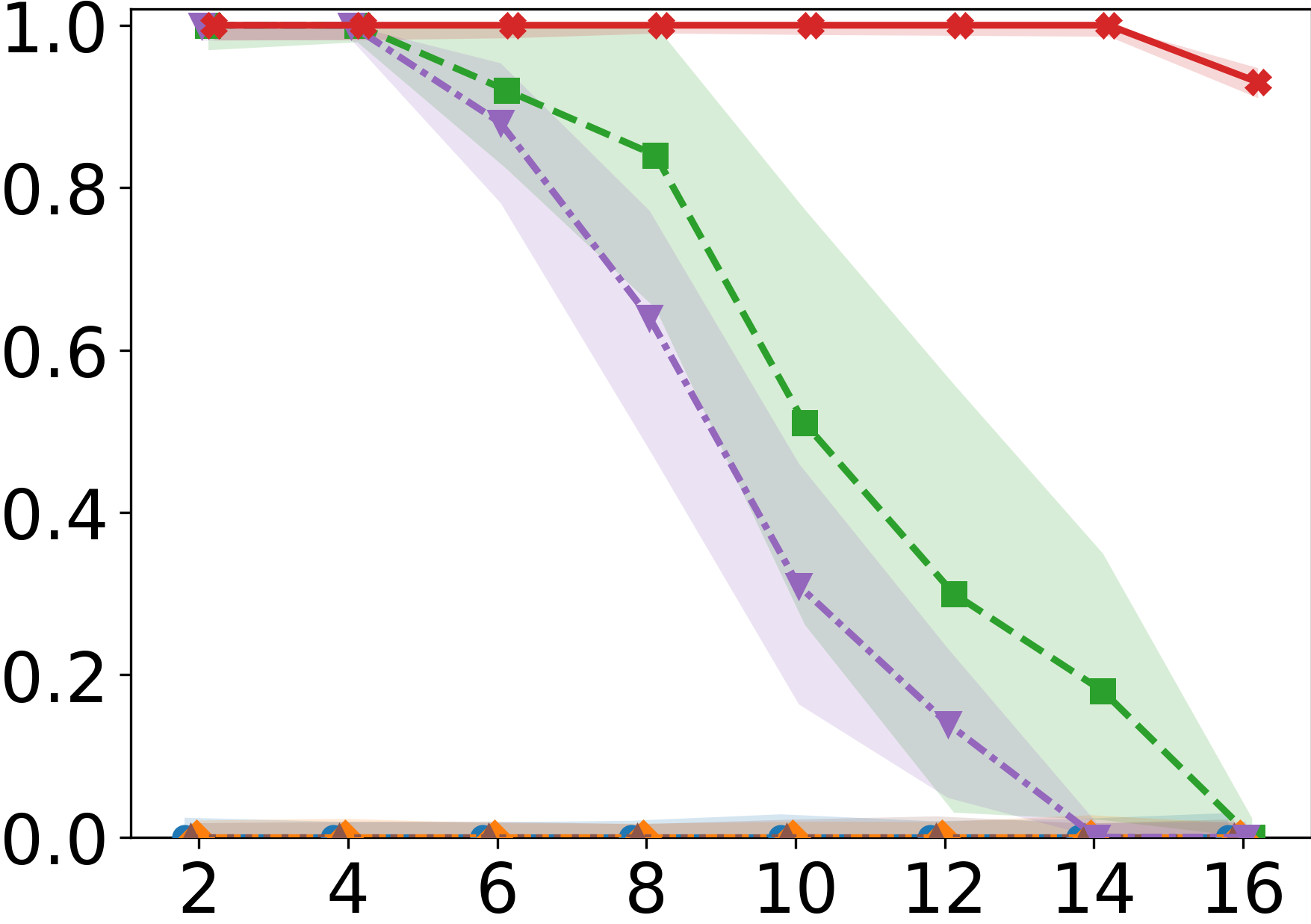}
  \caption{}\label{fig:k4_r4a}
\end{subfigure}\hspace{0.1\textwidth}
\begin{subfigure}[t]{0.38\textwidth}
  \centering
  \includegraphics[width=\linewidth]{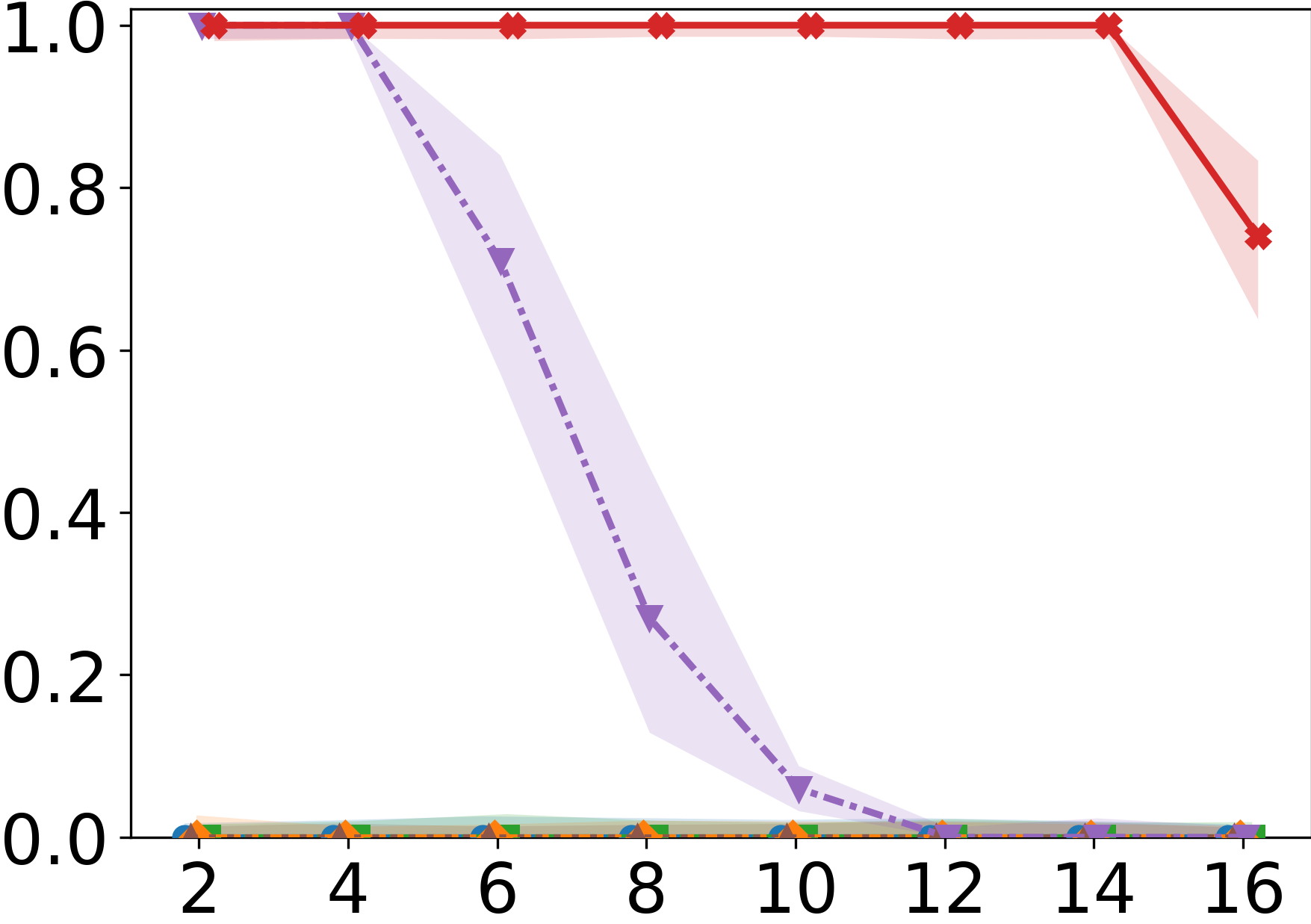}
  \caption{}\label{fig:k4_r4b}
\end{subfigure}

\caption{\textbf{Scalability plots for $k=4$ reachability.}
Each row corresponds to one environment on $k$-reachability tasks.
\textbf{Column (a)} shows the successful training ratio ($y$-axis) versus the number of training tasks $|\Train|$ ($x$-axis) for $k=4$.
\textbf{Column (b)} shows the zero-shot generalization ratio ($y$-axis) versus the number of training tasks $|\Train|$ ($x$-axis) for $k=4$.
We report mean ($\pm$ standard deviation) performance across \textbf{ten} random seeds.}
\label{fig:successvstasks_k4_all}
\end{figure*}

\begin{figure*}[t]
\centering
Legend:
{\small {\color{arsblue}\rule[0.6ex]{1.2em}{1pt}$\,\bullet$} ARS,
{\color{bcgreen}\rule[0.6ex]{1.2em}{1pt}$\,\blacksquare$} BC,
{\color{mamlorange}\rule[0.6ex]{1.2em}{1pt}$\,\blacklozenge$} MAML,
{\color{varibadbrown}\rule[0.6ex]{1.2em}{1pt}$\,\blacktriangle$} VariBAD,
{\color{genrlpurple}\rule[0.6ex]{1.2em}{1pt}$\,\blacktriangledown$} GenRL, and
{\color{dipsred}\rule[0.6ex]{1.2em}{1pt}{\large $\times$}} \drill~(Ours).}

\setlength{\abovecaptionskip}{2pt}
\setlength{\belowcaptionskip}{2pt}

\setcounter{subfigure}{0}
\par\medskip
{\centering\small\textbf{(i) \textsc{Car2D} $k$-Reachability}\par}
\vspace{0.2em}

\begin{subfigure}[t]{0.38\textwidth}
  \centering
  \includegraphics[width=\linewidth]{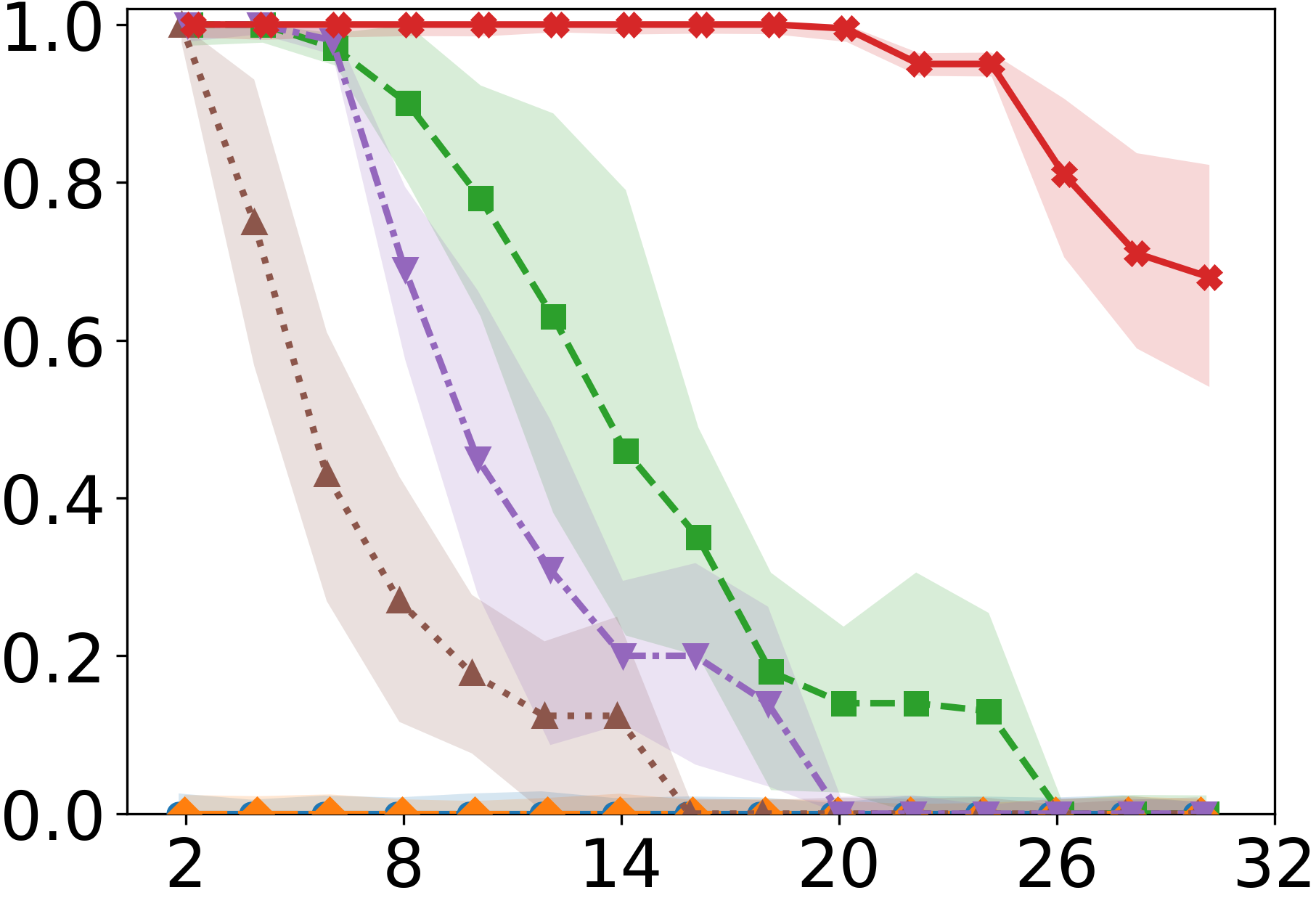}
  \caption{}\label{fig:k3_r1a}
\end{subfigure}\hspace{0.1\textwidth}
\begin{subfigure}[t]{0.38\textwidth}
  \centering
  \includegraphics[width=\linewidth]{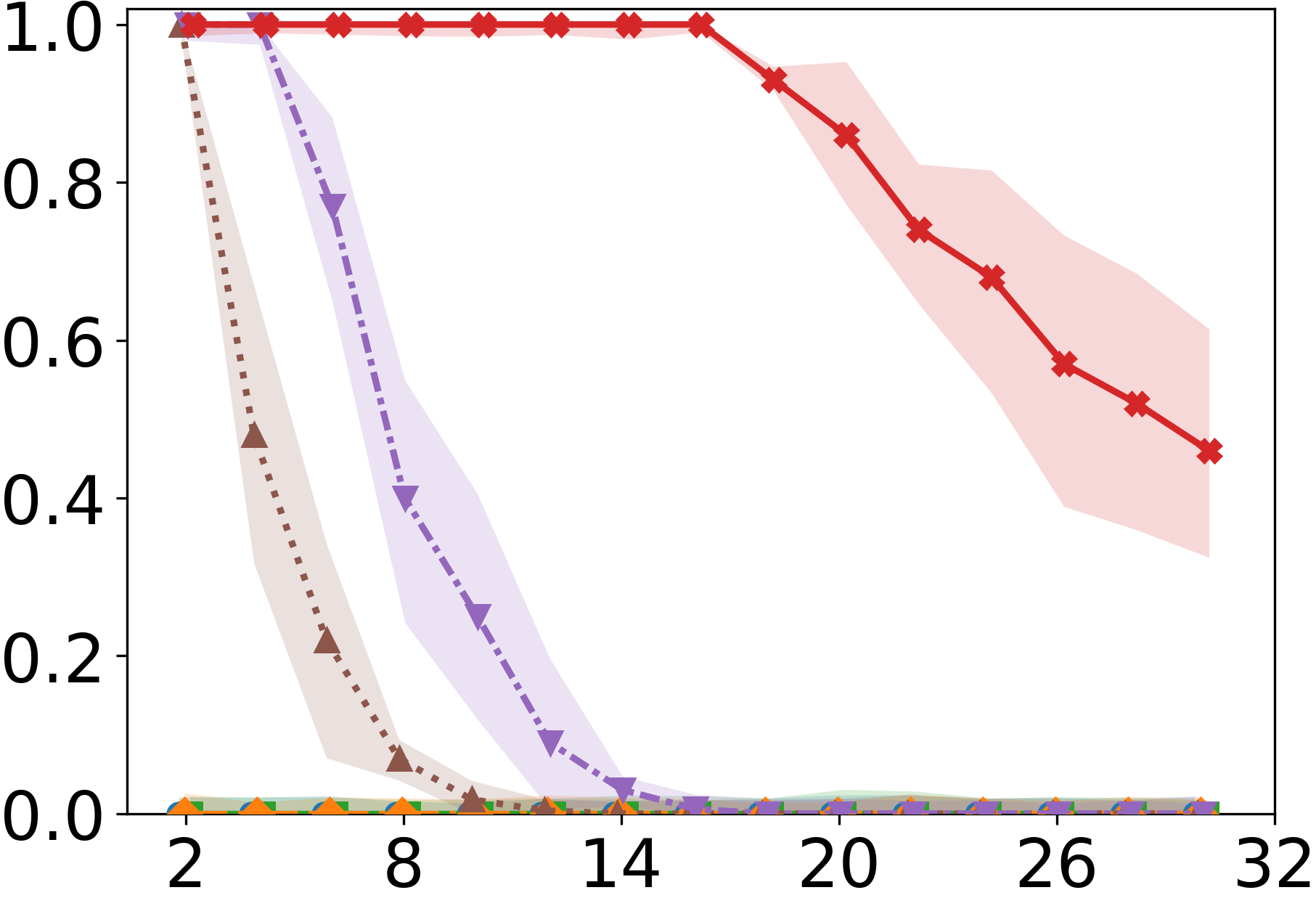}
  \caption{}\label{fig:k3_r1b}
\end{subfigure}

\setcounter{subfigure}{0}
\par\medskip
{\centering\small\textbf{(ii) \textsc{SimpleDrone} $k$-Reachability}\par}
\vspace{0.2em}

\begin{subfigure}[t]{0.38\textwidth}
  \centering
  \includegraphics[width=\linewidth]{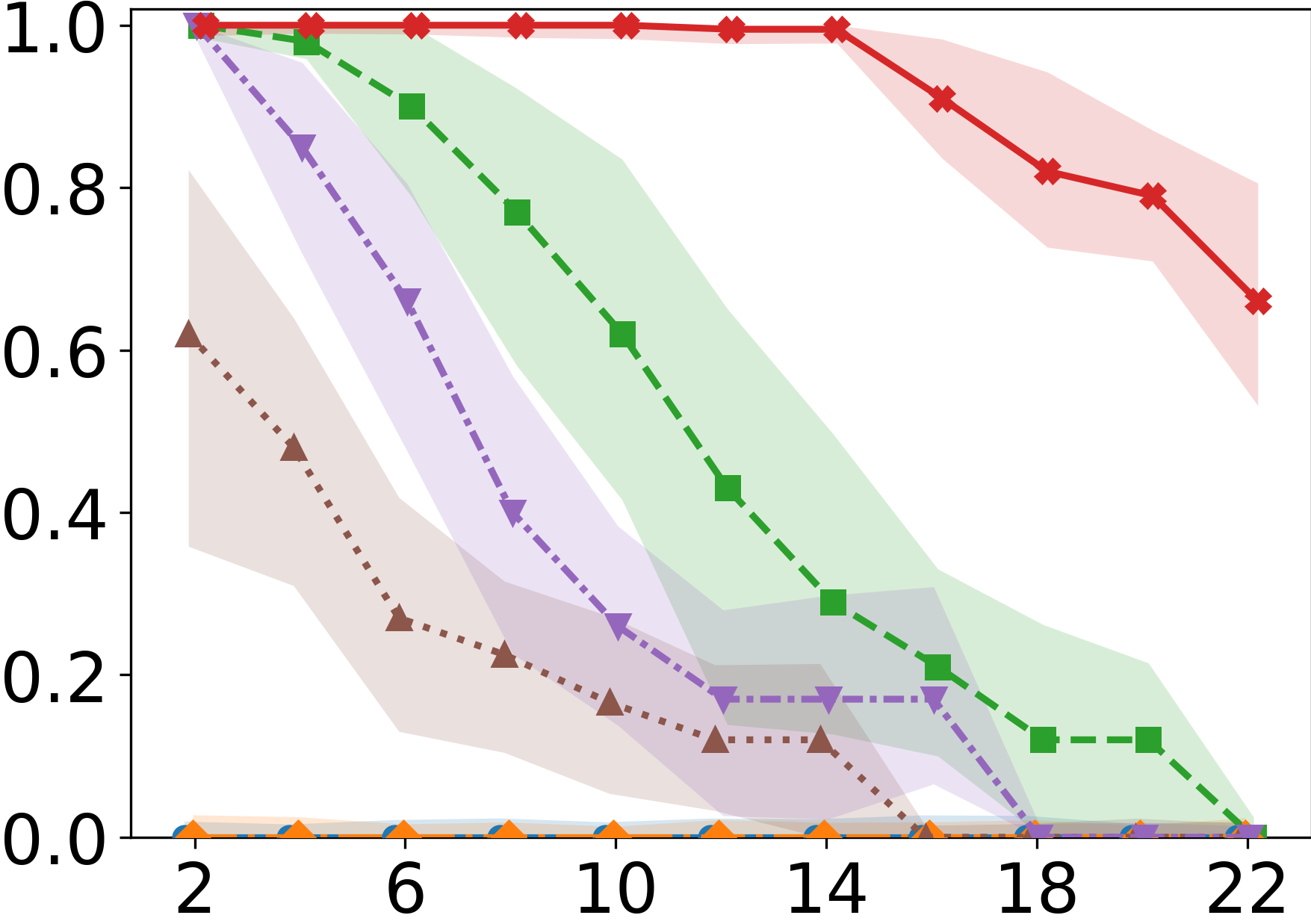}
  \caption{}\label{fig:k3_r2a}
\end{subfigure}\hspace{0.1\textwidth}
\begin{subfigure}[t]{0.38\textwidth}
  \centering
  \includegraphics[width=\linewidth]{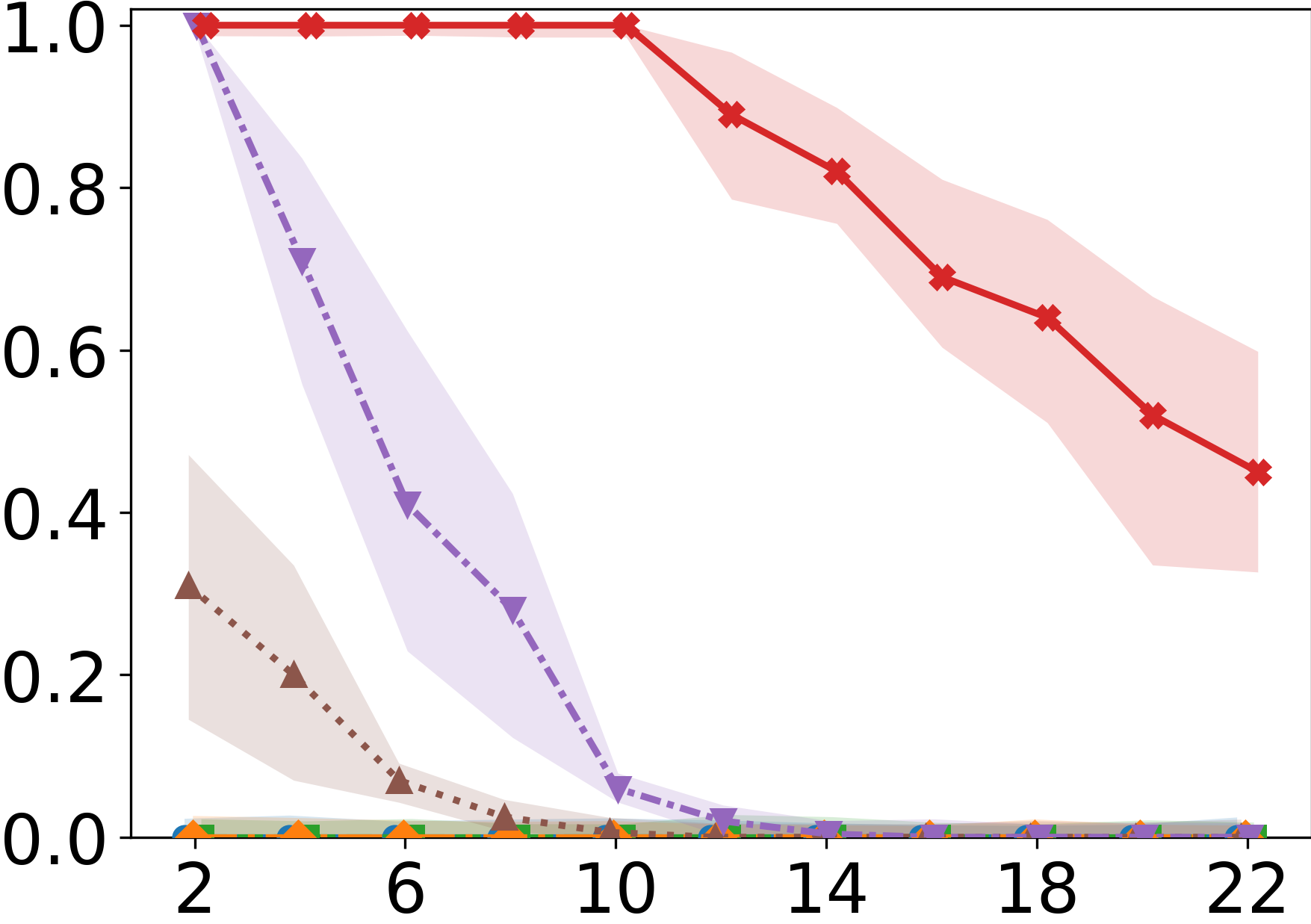}
  \caption{}\label{fig:k3_r2b}
\end{subfigure}

\setcounter{subfigure}{0}
\par\medskip
{\centering\small\textbf{(iii) \textsc{DroneAttitude} $k$-Reachability}\par}
\vspace{0.2em}

\begin{subfigure}[t]{0.38\textwidth}
  \centering
  \includegraphics[width=\linewidth]{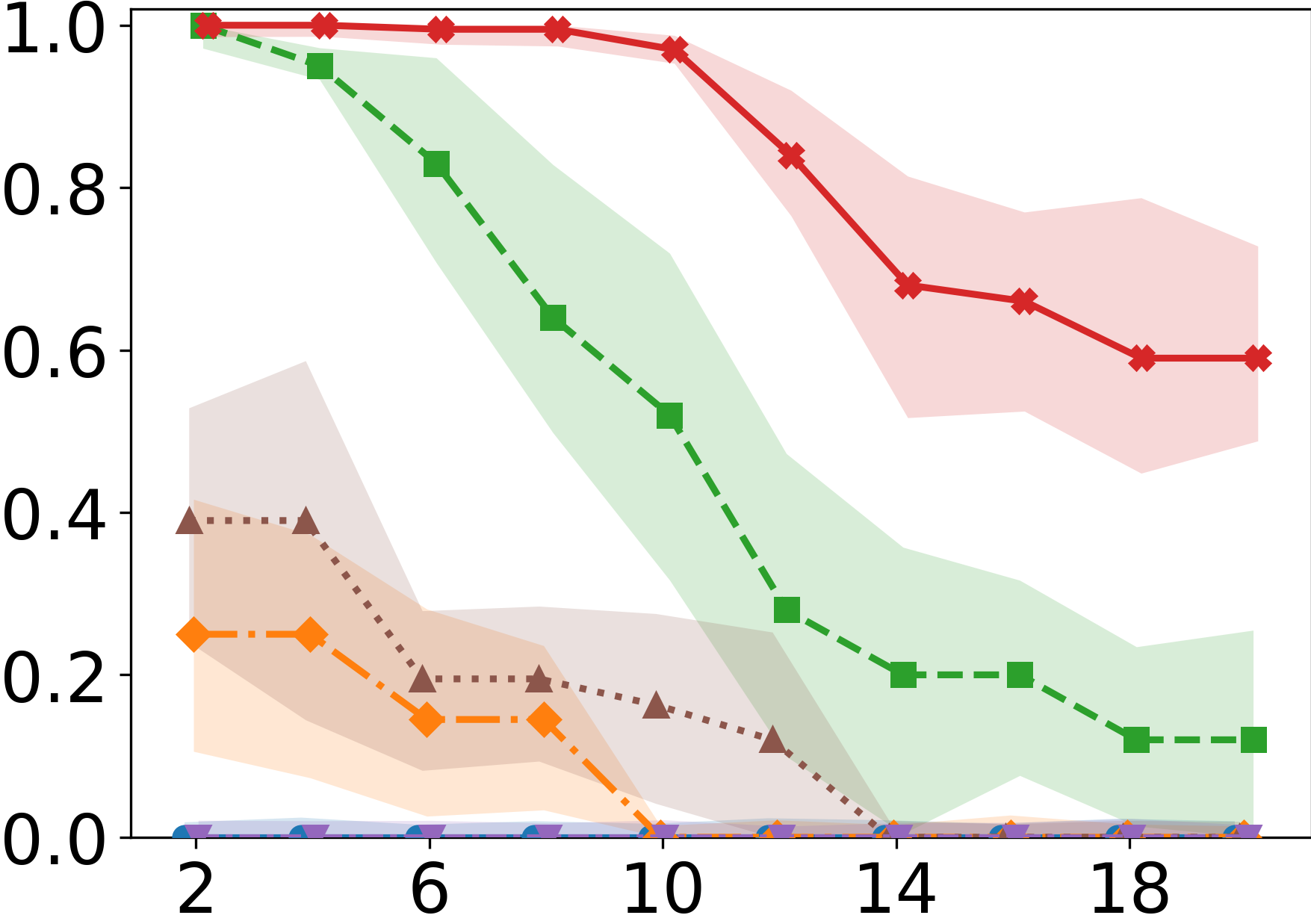}
  \caption{}\label{fig:k3_r3a}
\end{subfigure}\hspace{0.1\textwidth}
\begin{subfigure}[t]{0.38\textwidth}
  \centering
  \includegraphics[width=\linewidth]{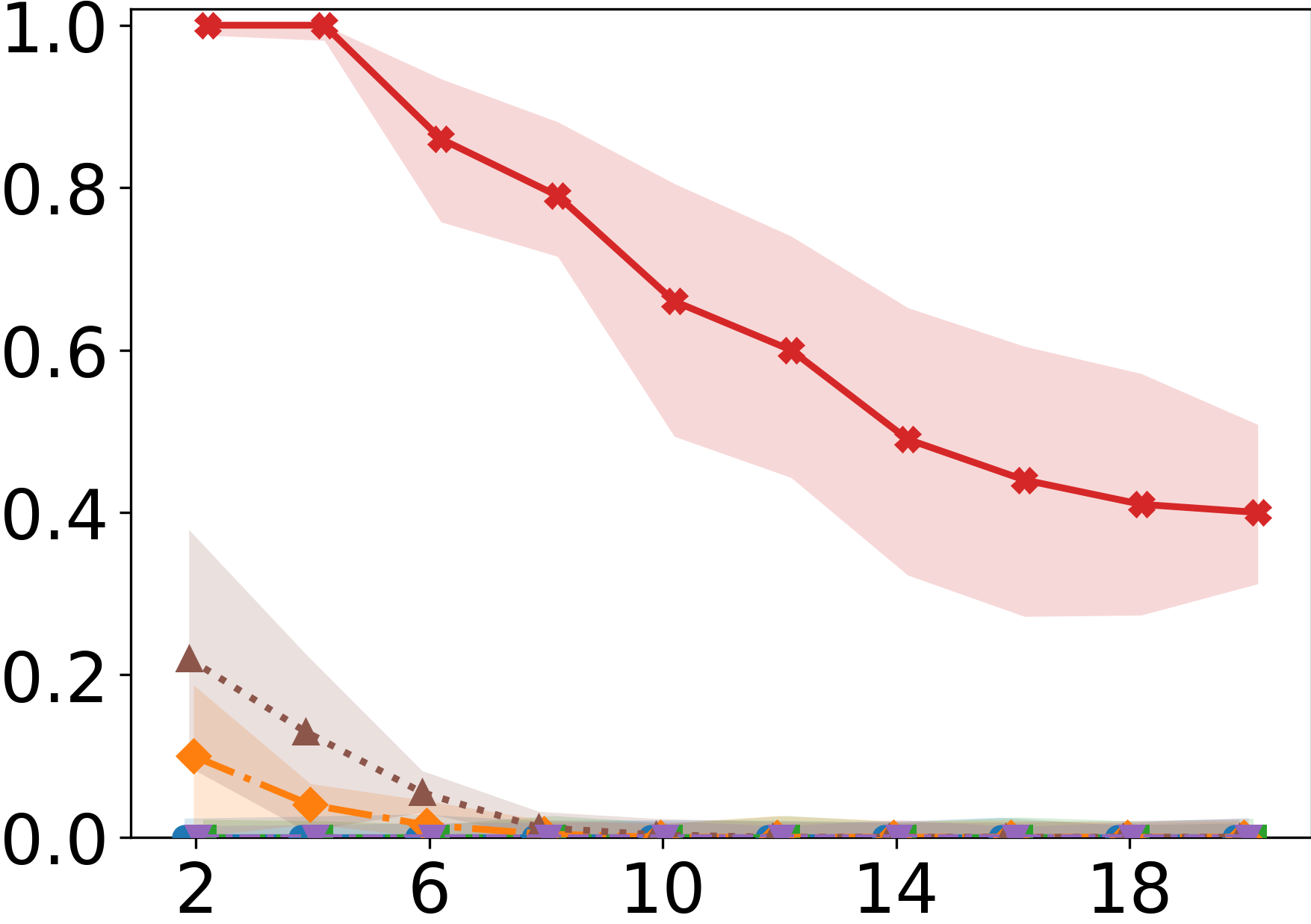}
  \caption{}\label{fig:k3_r3b}
\end{subfigure}

\setcounter{subfigure}{0}
\par\medskip
{\centering\small\textbf{(iv) \textsc{Car2D} $k$-Reachability + obstacles}\par}
\vspace{0.2em}

\begin{subfigure}[t]{0.38\textwidth}
  \centering
  \includegraphics[width=\linewidth]{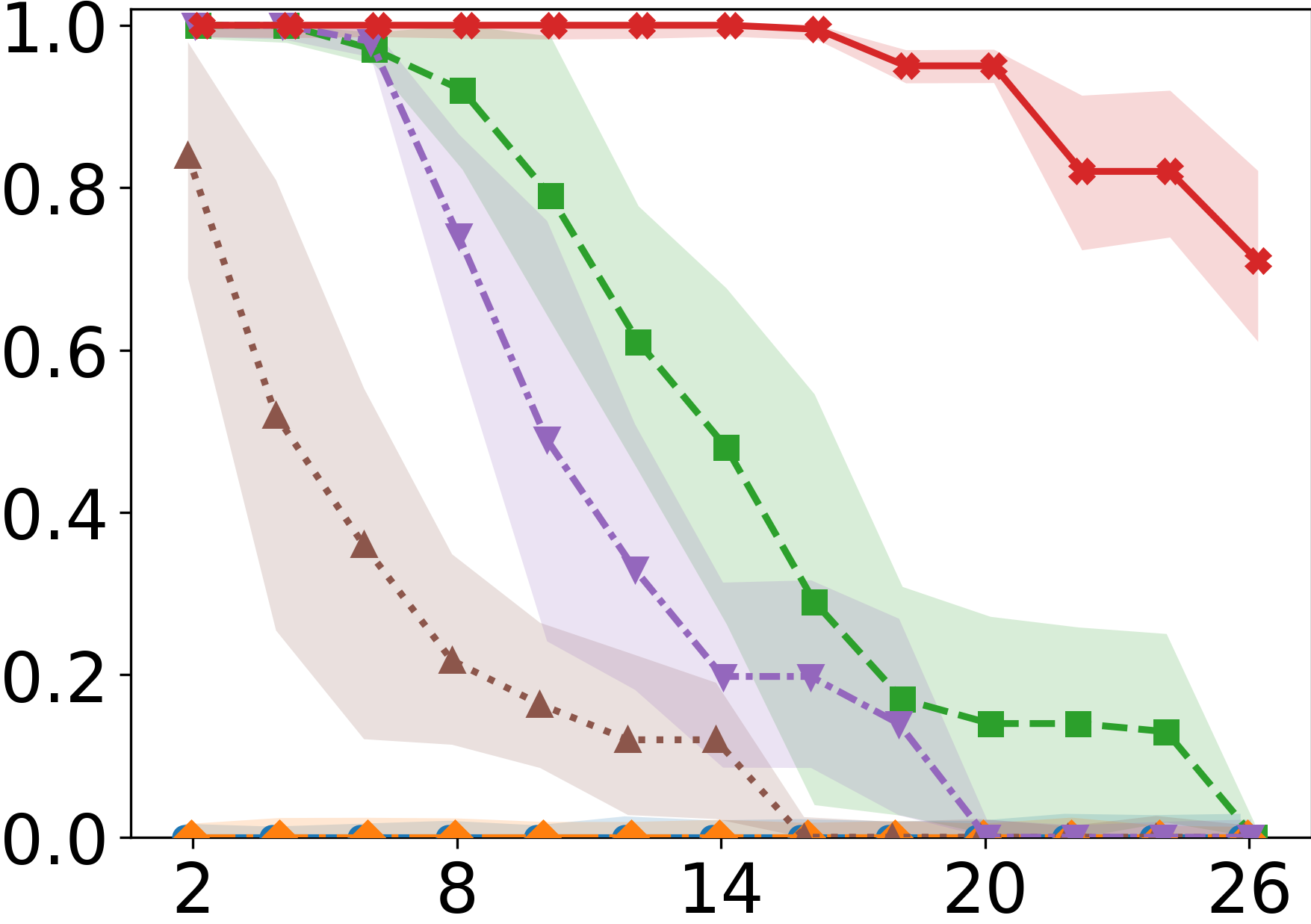}
  \caption{}\label{fig:k3_r4a}
\end{subfigure}\hspace{0.1\textwidth}
\begin{subfigure}[t]{0.38\textwidth}
  \centering
  \includegraphics[width=\linewidth]{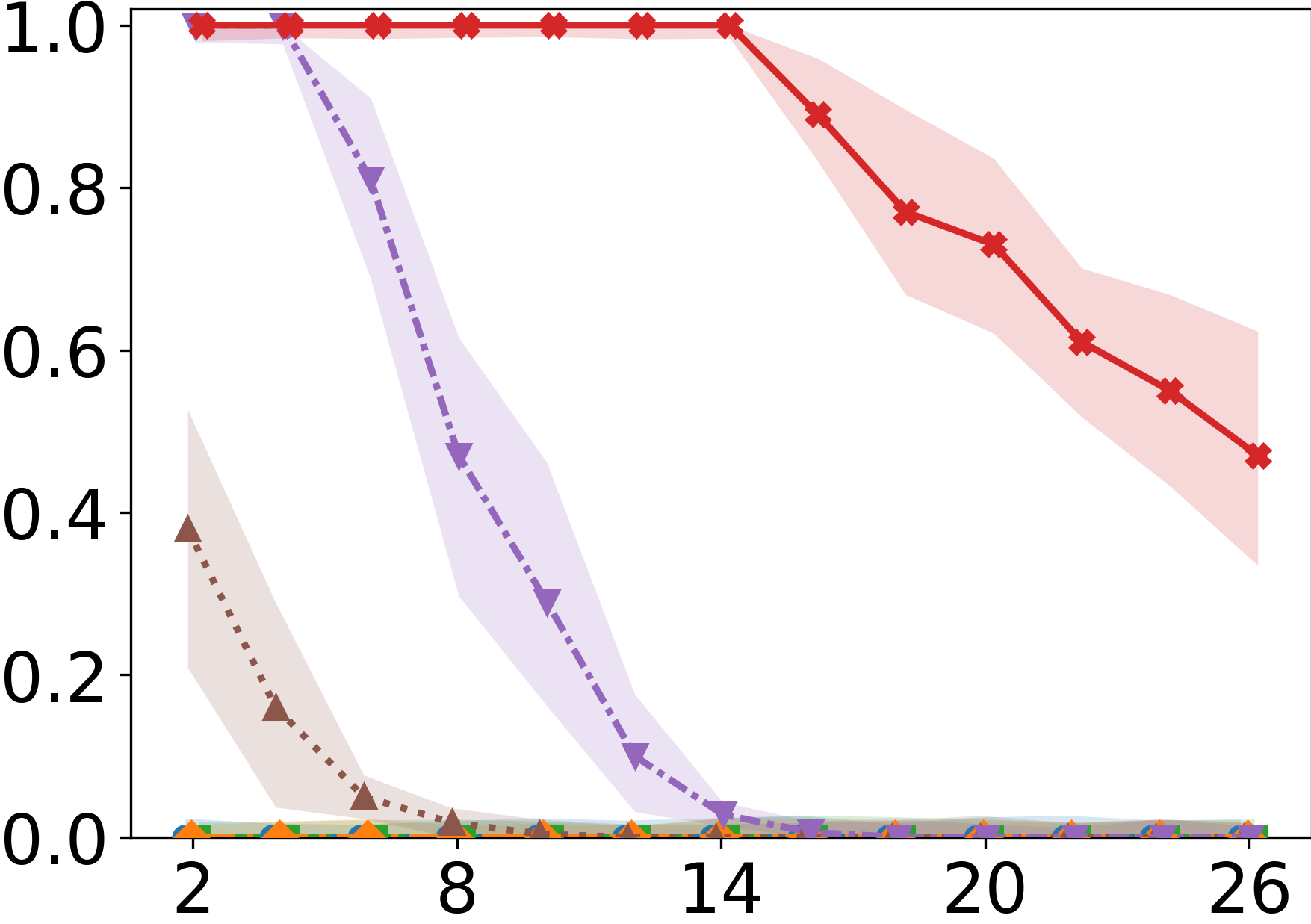}
  \caption{}\label{fig:k3_r4b}
\end{subfigure}

\caption{\textbf{Scalability plots for $k=3$ reachability.}
Each row corresponds to one environment on $k$-reachability tasks.
\textbf{Column (a)} shows the successful training ratio ($y$-axis) versus the number of training tasks $|\Train|$ ($x$-axis) for $k=3$.
\textbf{Column (b)} shows the zero-shot generalization ratio ($y$-axis) versus the number of training tasks $|\Train|$ ($x$-axis) for $k=3$.
We report mean ($\pm$ standard deviation) performance across \textbf{ten} random seeds.}
\label{fig:successvstasks_k3_all}
\end{figure*}

\begin{figure*}[t]
\centering
Legend:
{\small {\color{arsblue}\rule[0.6ex]{1.2em}{1pt}$\,\bullet$} ARS,
{\color{bcgreen}\rule[0.6ex]{1.2em}{1pt}$\,\blacksquare$} BC,
{\color{mamlorange}\rule[0.6ex]{1.2em}{1pt}$\,\blacklozenge$} MAML,
{\color{varibadbrown}\rule[0.6ex]{1.2em}{1pt}$\,\blacktriangle$} VariBAD,
{\color{genrlpurple}\rule[0.6ex]{1.2em}{1pt}$\,\blacktriangledown$} GenRL, and
{\color{dipsred}\rule[0.6ex]{1.2em}{1pt}{\large $\times$}} \drill~(Ours).}

\setlength{\abovecaptionskip}{2pt}
\setlength{\belowcaptionskip}{2pt}

\setcounter{subfigure}{0}
\par\medskip
{\centering\small\textbf{(i) \textsc{Car2D} $k$-Reachability}\par}
\vspace{0.2em}

\begin{subfigure}[t]{0.38\textwidth}
  \centering
  \includegraphics[width=\linewidth]{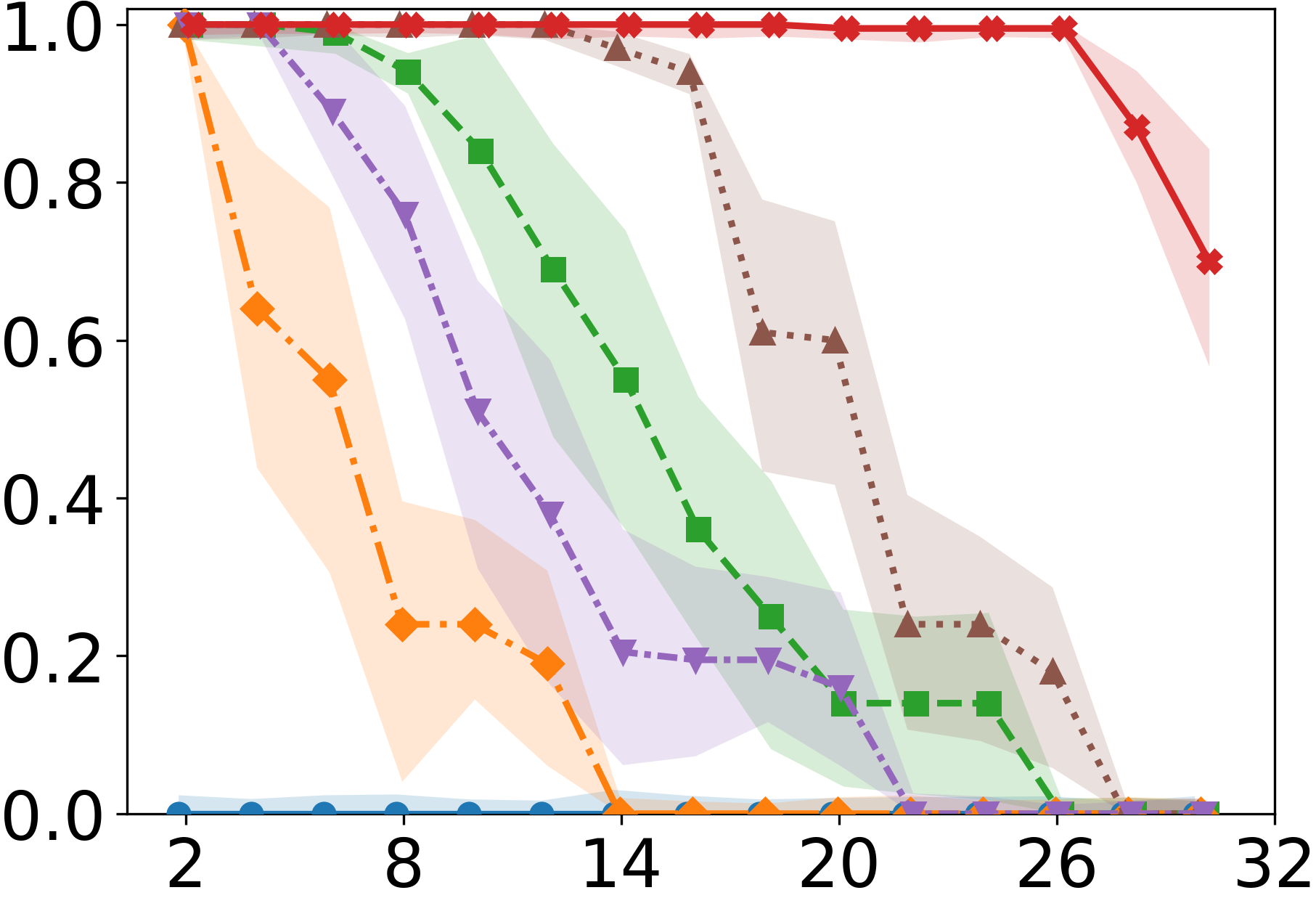}
  \caption{}\label{fig:k2_r1a}
\end{subfigure}\hspace{0.1\textwidth}
\begin{subfigure}[t]{0.38\textwidth}
  \centering
  \includegraphics[width=\linewidth]{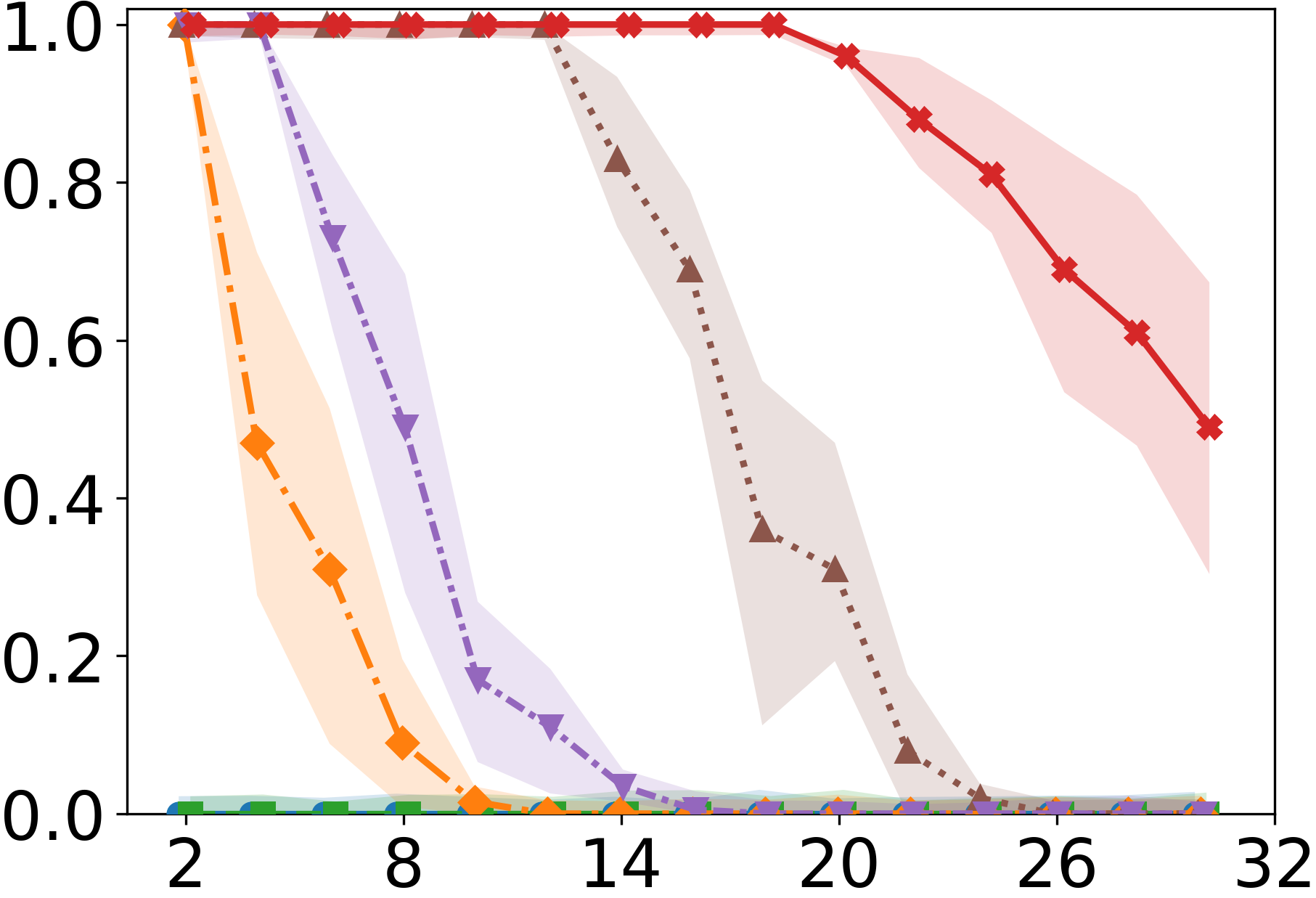}
  \caption{}\label{fig:k2_r1b}
\end{subfigure}

\setcounter{subfigure}{0}
\par\medskip
{\centering\small\textbf{(ii) \textsc{SimpleDrone} $k$-Reachability}\par}
\vspace{0.2em}

\begin{subfigure}[t]{0.38\textwidth}
  \centering
  \includegraphics[width=\linewidth]{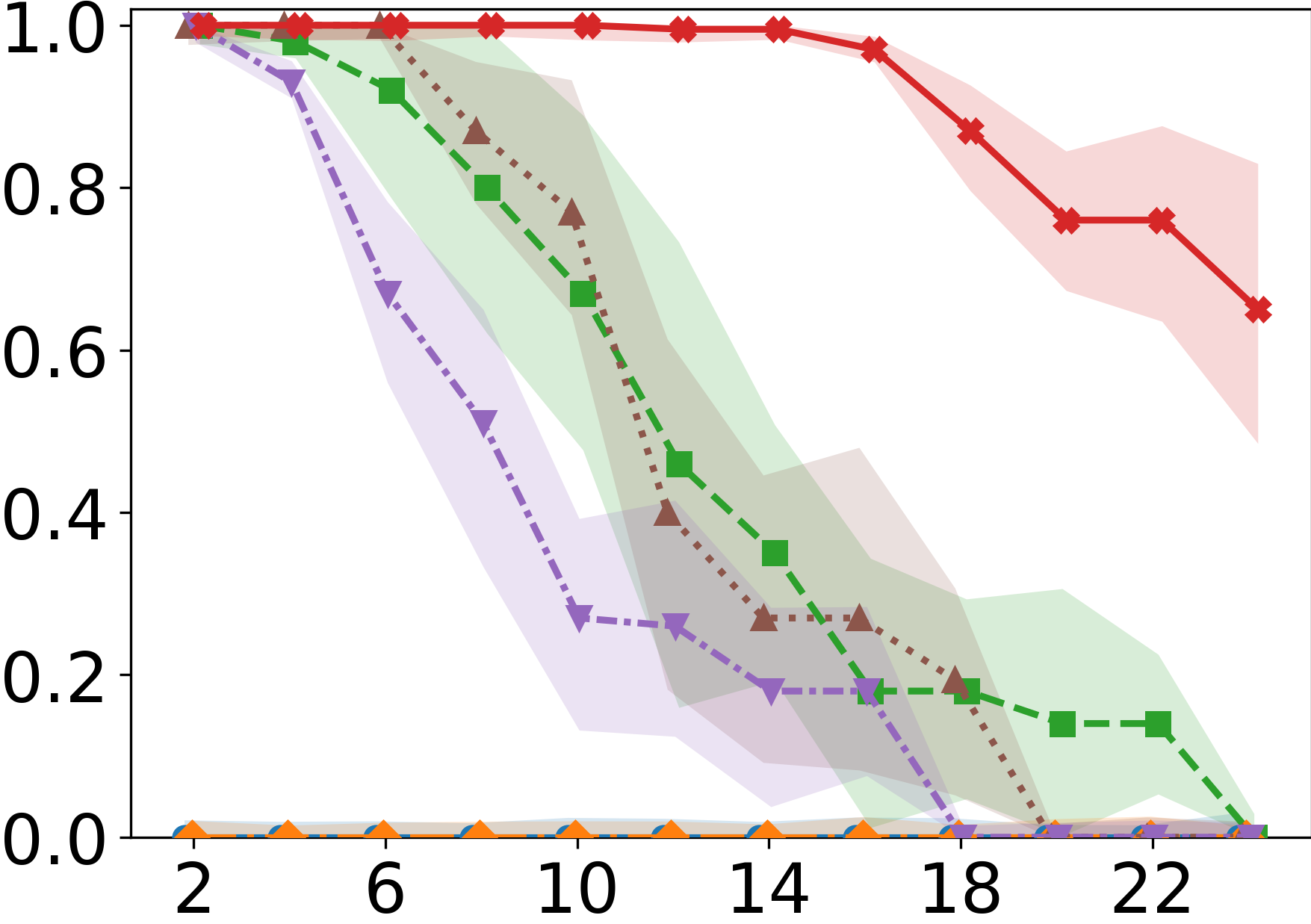}
  \caption{}\label{fig:k2_r2a}
\end{subfigure}\hspace{0.1\textwidth}
\begin{subfigure}[t]{0.38\textwidth}
  \centering
  \includegraphics[width=\linewidth]{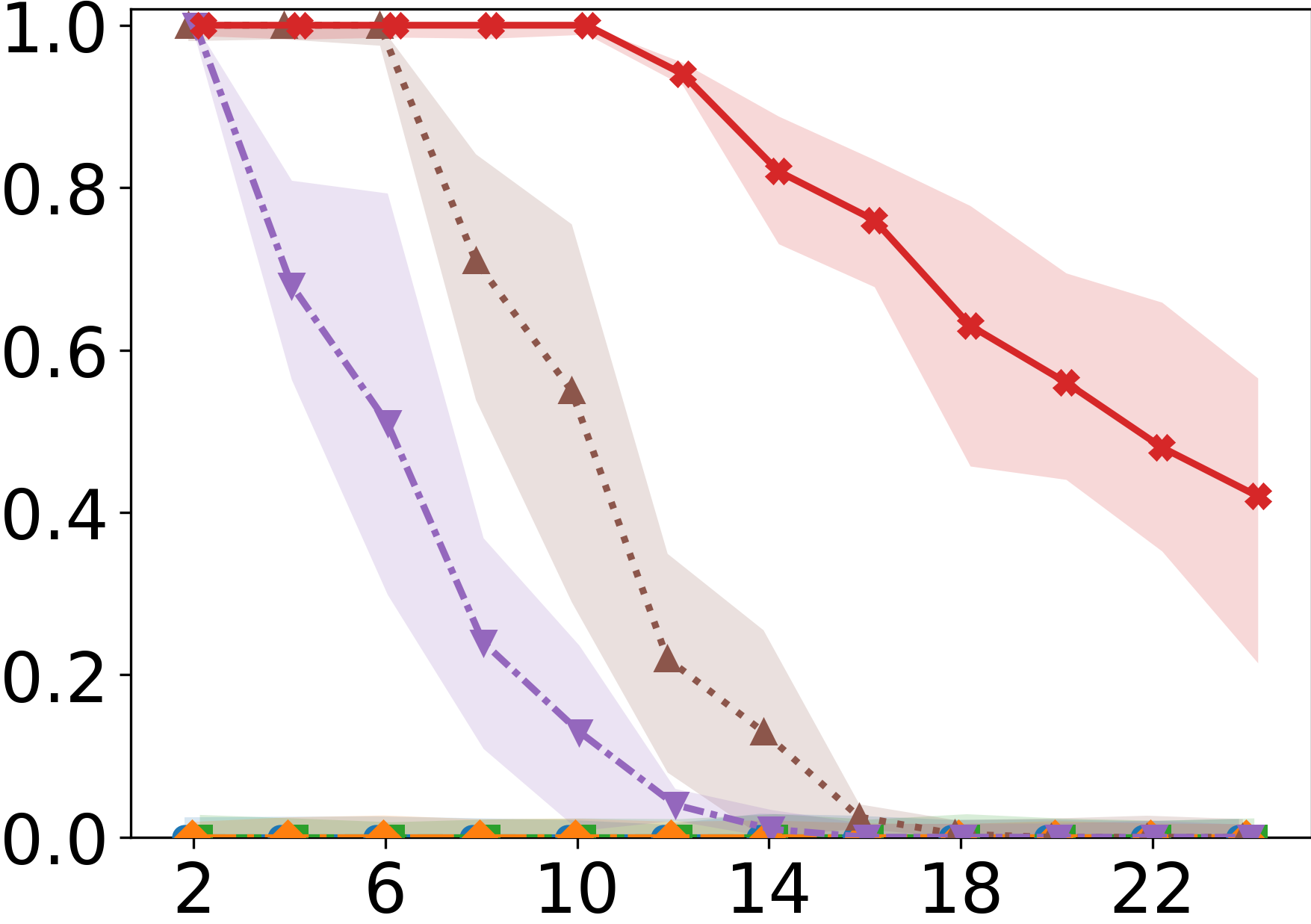}
  \caption{}\label{fig:k2_r2b}
\end{subfigure}

\setcounter{subfigure}{0}
\par\medskip
{\centering\small\textbf{(iii) \textsc{DroneAttitude} $k$-Reachability}\par}
\vspace{0.2em}

\begin{subfigure}[t]{0.38\textwidth}
  \centering
  \includegraphics[width=\linewidth]{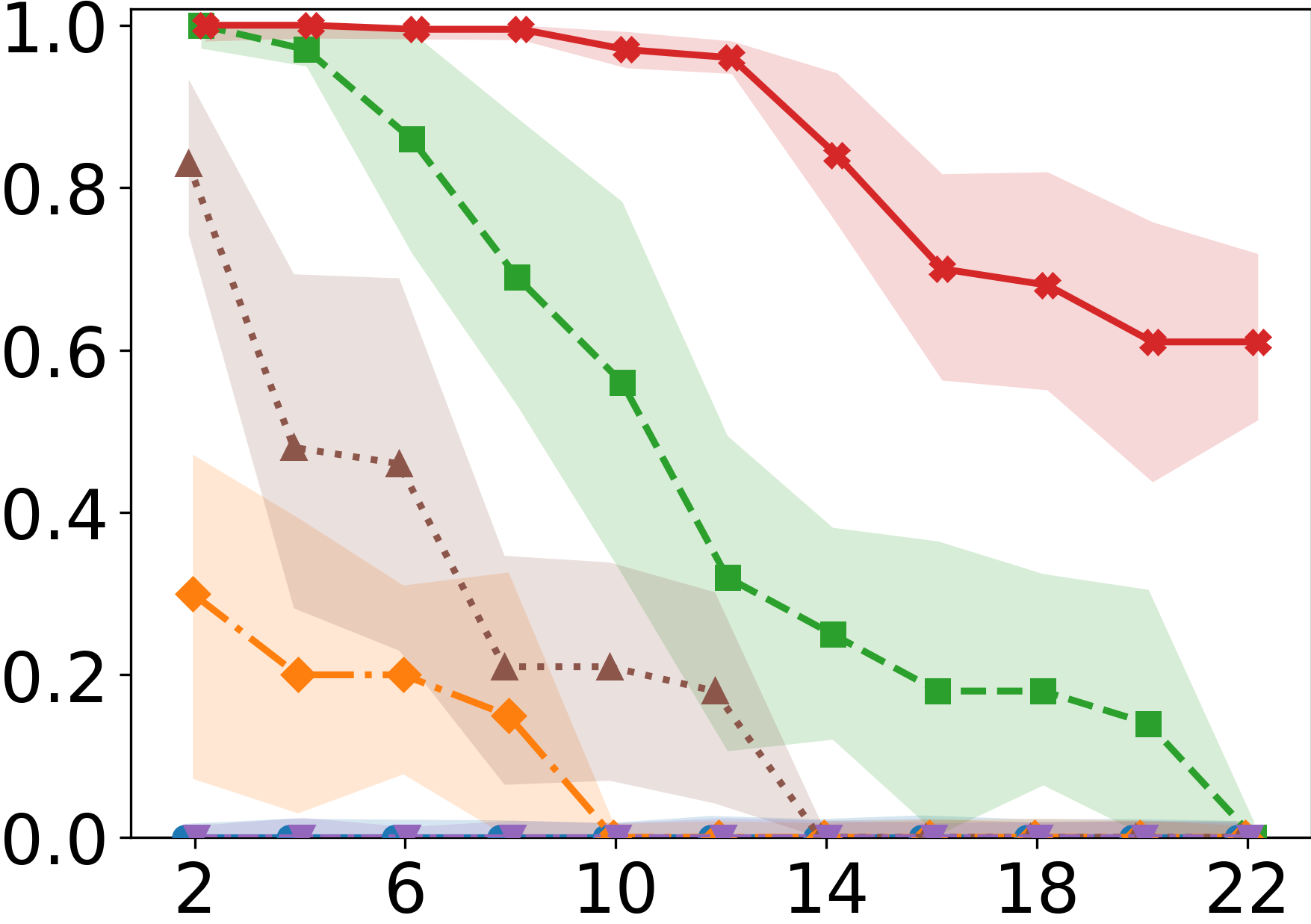}
  \caption{}\label{fig:k2_r3a}
\end{subfigure}\hspace{0.1\textwidth}
\begin{subfigure}[t]{0.38\textwidth}
  \centering
  \includegraphics[width=\linewidth]{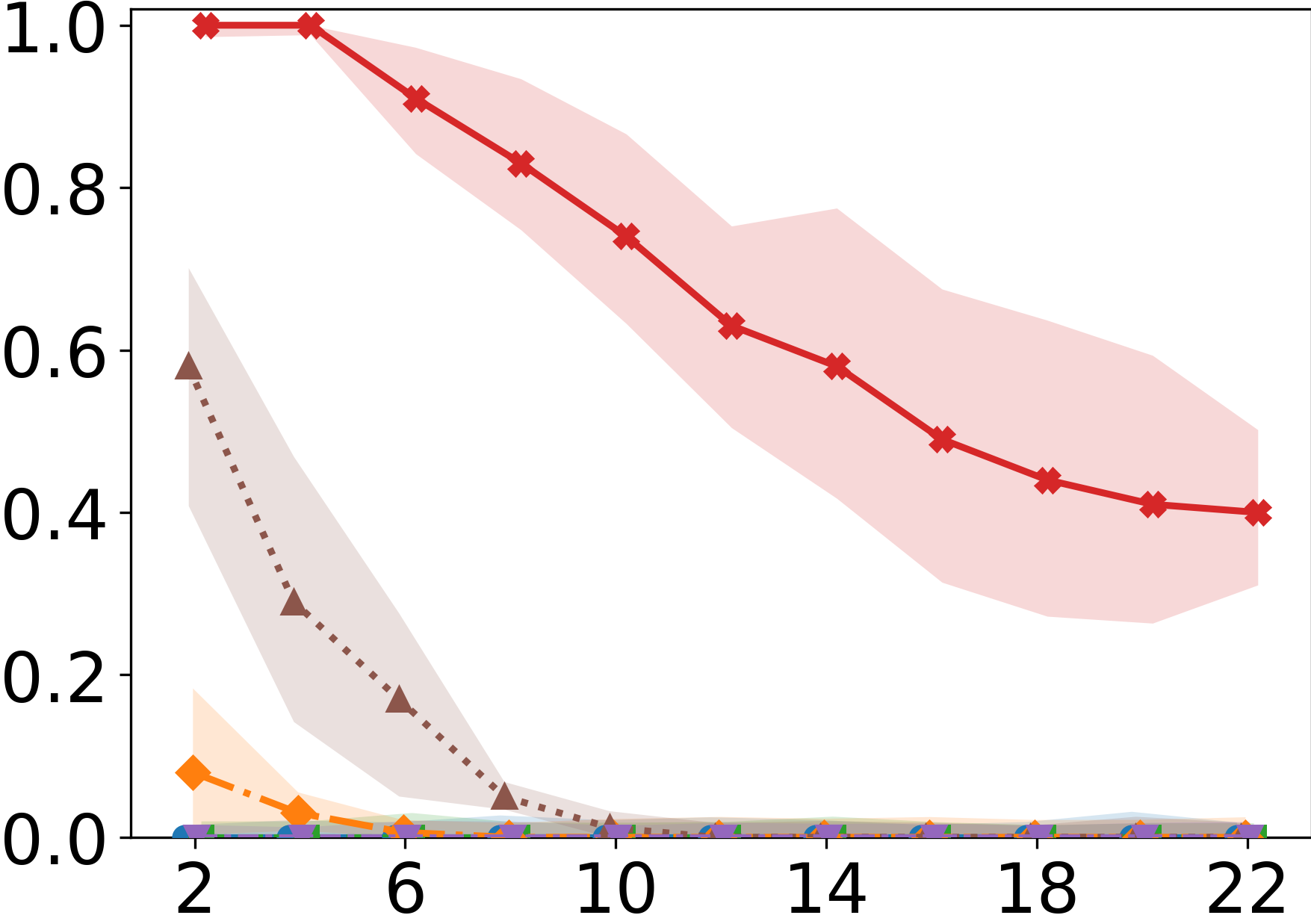}
  \caption{}\label{fig:k2_r3b}
\end{subfigure}

\setcounter{subfigure}{0}
\par\medskip
{\centering\small\textbf{(iv) \textsc{Car2D} $k$-Reachability + obstacles}\par}
\vspace{0.2em}

\begin{subfigure}[t]{0.38\textwidth}
  \centering
  \includegraphics[width=\linewidth]{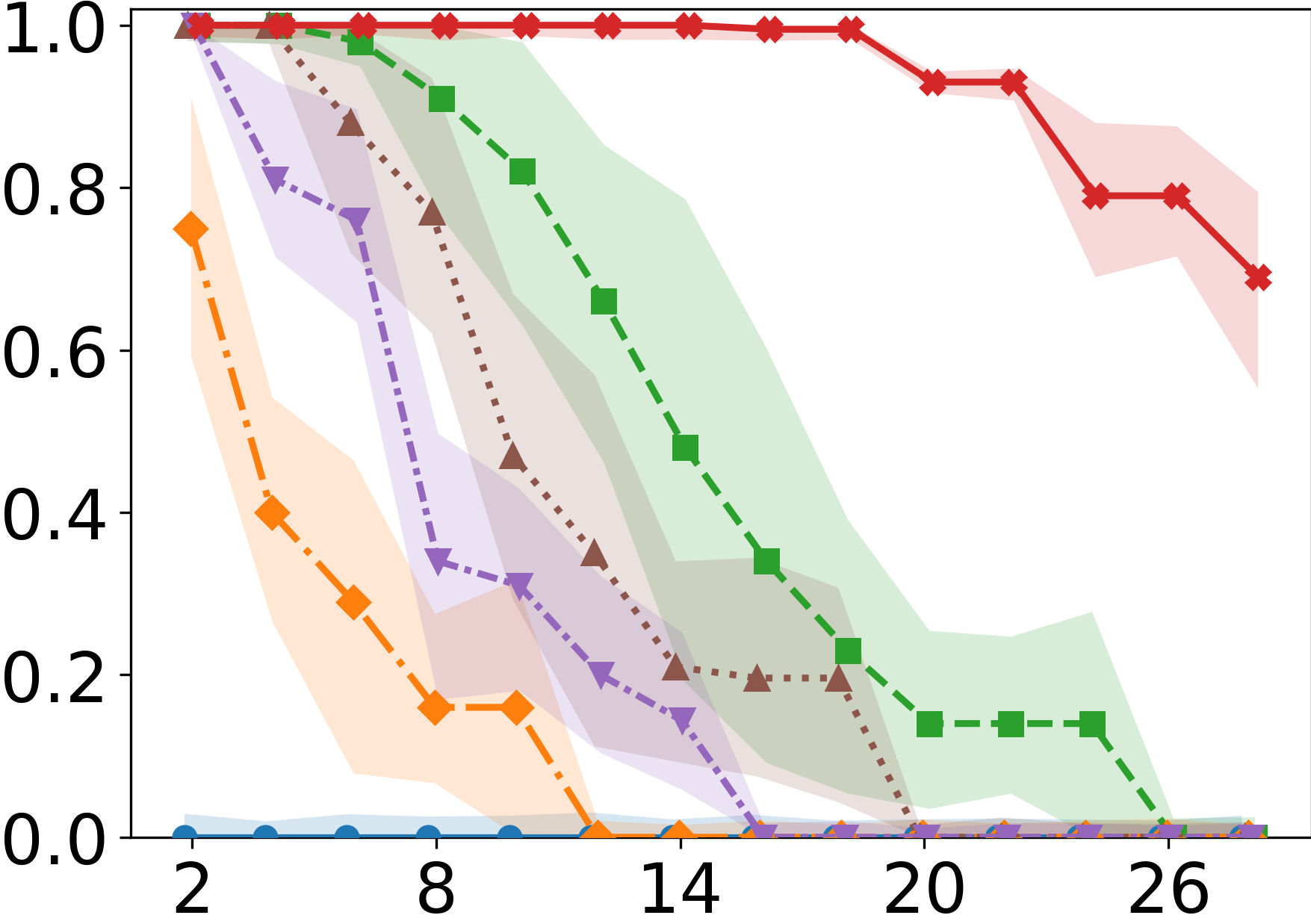}
  \caption{}\label{fig:k2_r4a}
\end{subfigure}\hspace{0.1\textwidth}
\begin{subfigure}[t]{0.38\textwidth}
  \centering
  \includegraphics[width=\linewidth]{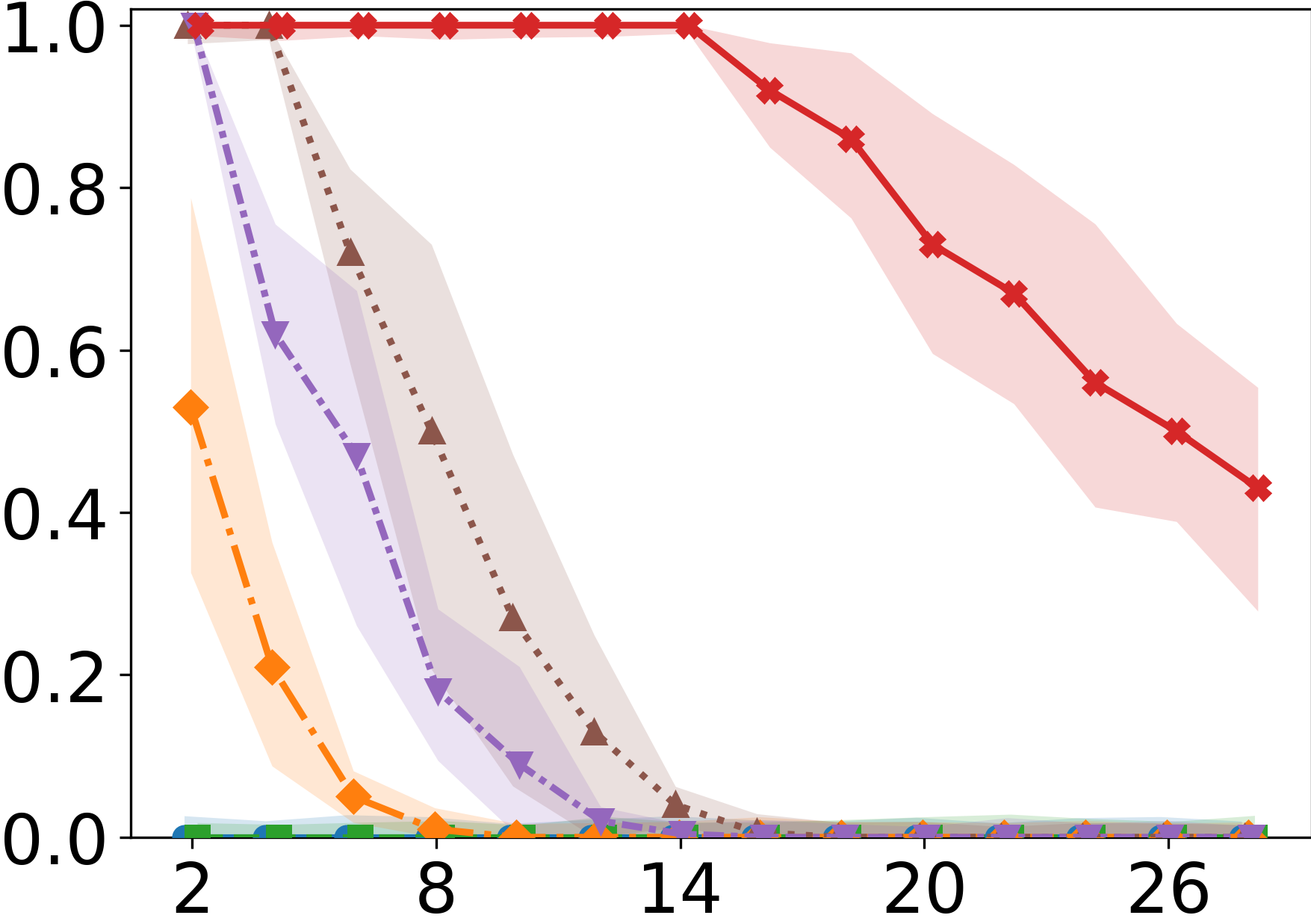}
  \caption{}\label{fig:k2_r4b}
\end{subfigure}

\caption{\textbf{Scalability plots for $k=2$ reachability.}
Each row corresponds to one environment on $k$-reachability tasks.
\textbf{Column (a)} shows the successful training ratio ($y$-axis) versus the number of training tasks $|\Train|$ ($x$-axis) for $k=2$.
\textbf{Column (b)} shows the zero-shot generalization ratio ($y$-axis) versus the number of training tasks $|\Train|$ ($x$-axis) for $k=2$.
We report mean ($\pm$ standard deviation) performance across \textbf{ten} random seeds.}
\label{fig:successvstasks_k2_all}
\end{figure*}

\begin{figure*}[t]
\centering
Legend:
{\small {\color{arsblue}\rule[0.6ex]{1.2em}{1pt}$\,\bullet$} ARS,
{\color{bcgreen}\rule[0.6ex]{1.2em}{1pt}$\,\blacksquare$} BC,
{\color{mamlorange}\rule[0.6ex]{1.2em}{1pt}$\,\blacklozenge$} MAML,
{\color{varibadbrown}\rule[0.6ex]{1.2em}{1pt}$\,\blacktriangle$} VariBAD,
{\color{genrlpurple}\rule[0.6ex]{1.2em}{1pt}$\,\blacktriangledown$} GenRL, and
{\color{dipsred}\rule[0.6ex]{1.2em}{1pt}{\large $\times$}} \drill~(Ours).}

\setlength{\abovecaptionskip}{2pt}
\setlength{\belowcaptionskip}{2pt}

\setcounter{subfigure}{0}
\par\medskip
{\centering\small\textbf{(i) \textsc{Car2D} $k$-Reachability}\par}
\vspace{0.2em}

\begin{subfigure}[t]{0.38\textwidth}
  \centering
  \includegraphics[width=\linewidth]{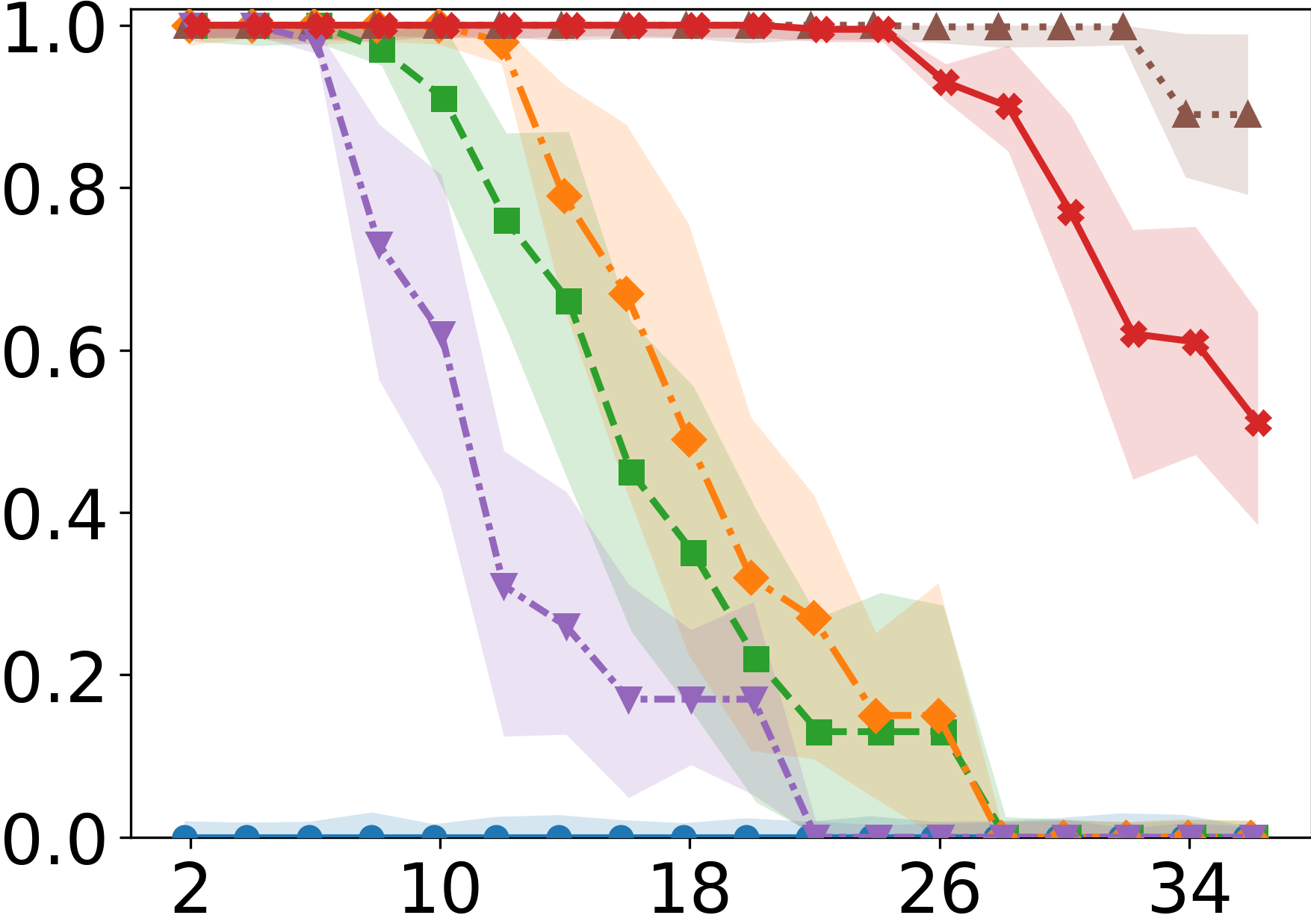}
  \caption{}\label{fig:k1_r1a}
\end{subfigure}\hspace{0.1\textwidth}
\begin{subfigure}[t]{0.38\textwidth}
  \centering
  \includegraphics[width=\linewidth]{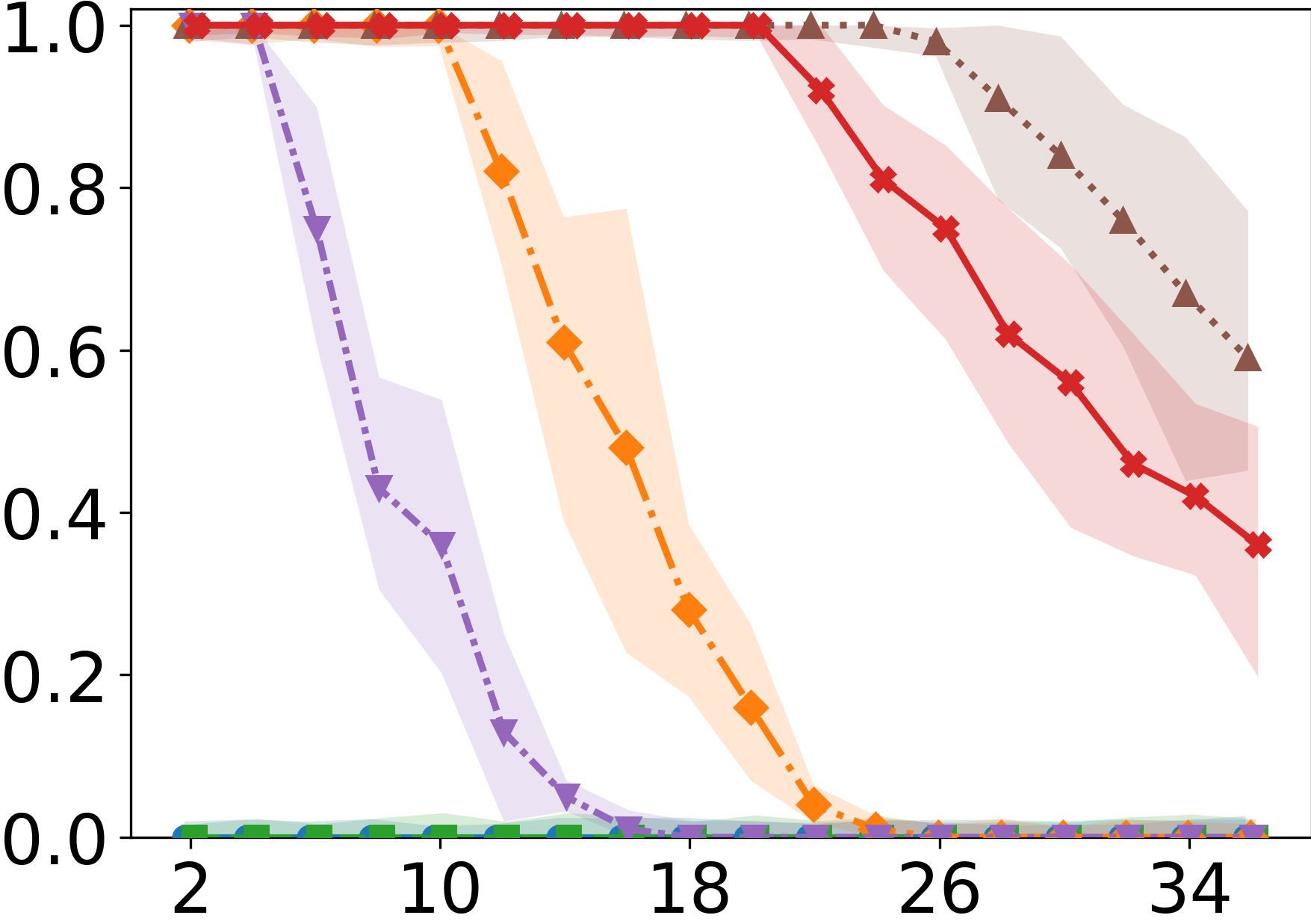}
  \caption{}\label{fig:k1_r1b}
\end{subfigure}

\setcounter{subfigure}{0}
\par\medskip
{\centering\small\textbf{(ii) \textsc{SimpleDrone} $k$-Reachability}\par}
\vspace{0.2em}

\begin{subfigure}[t]{0.38\textwidth}
  \centering
  \includegraphics[width=\linewidth]{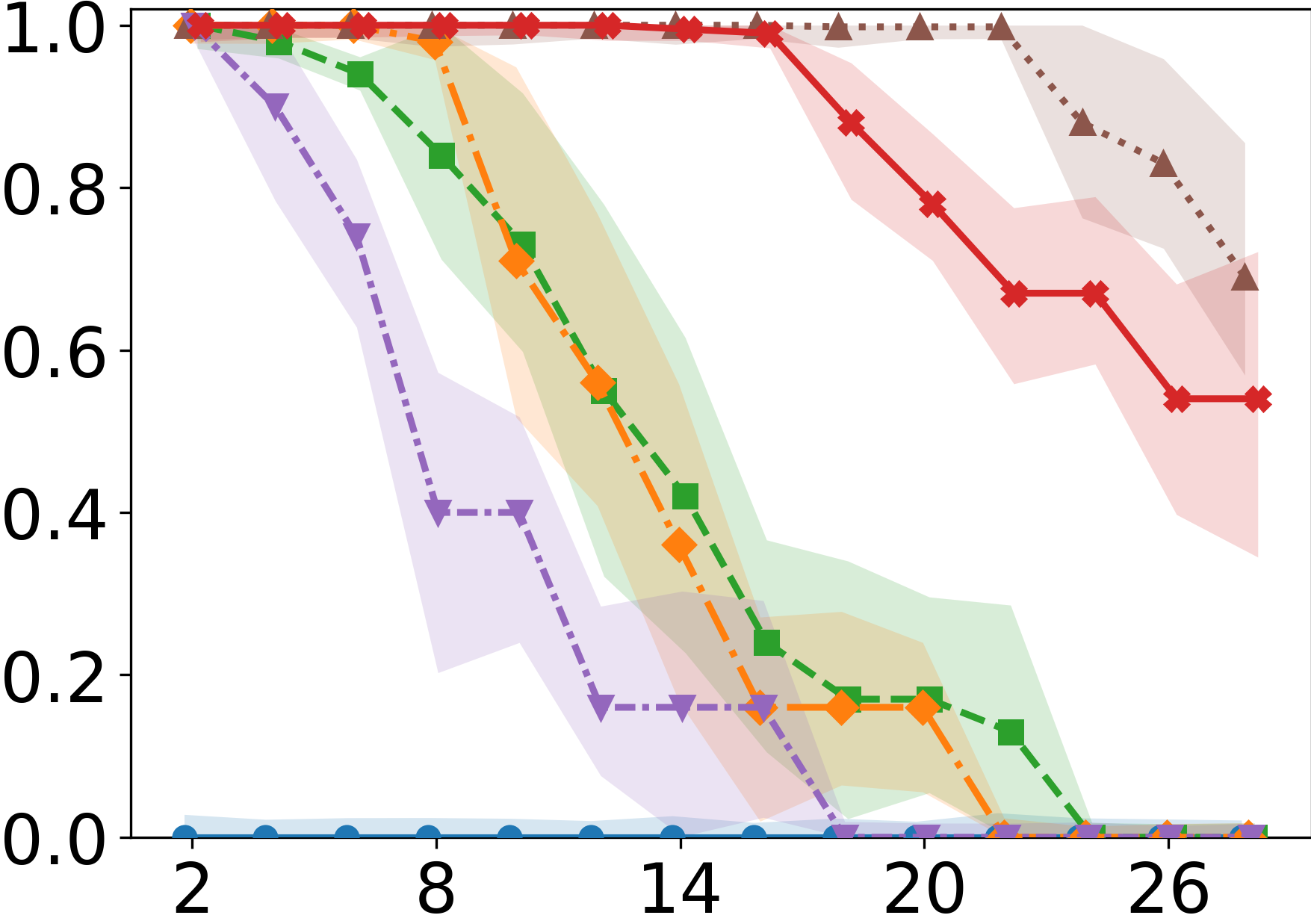}
  \caption{}\label{fig:k1_r2a}
\end{subfigure}\hspace{0.1\textwidth}
\begin{subfigure}[t]{0.38\textwidth}
  \centering
  \includegraphics[width=\linewidth]{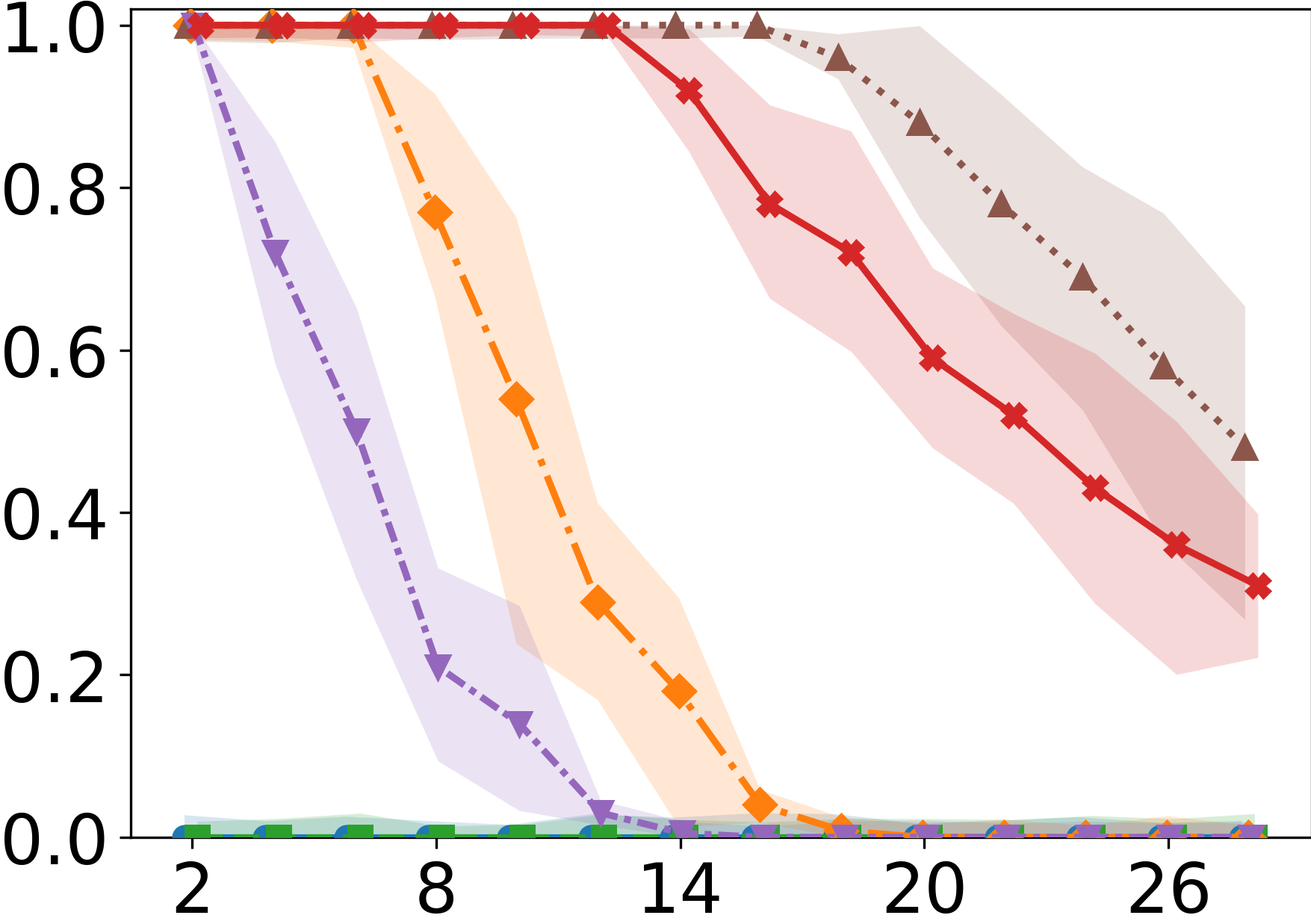}
  \caption{}\label{fig:k1_r2b}
\end{subfigure}

\setcounter{subfigure}{0}
\par\medskip
{\centering\small\textbf{(iii) \textsc{DroneAttitude} $k$-Reachability}\par}
\vspace{0.2em}

\begin{subfigure}[t]{0.38\textwidth}
  \centering
  \includegraphics[width=\linewidth]{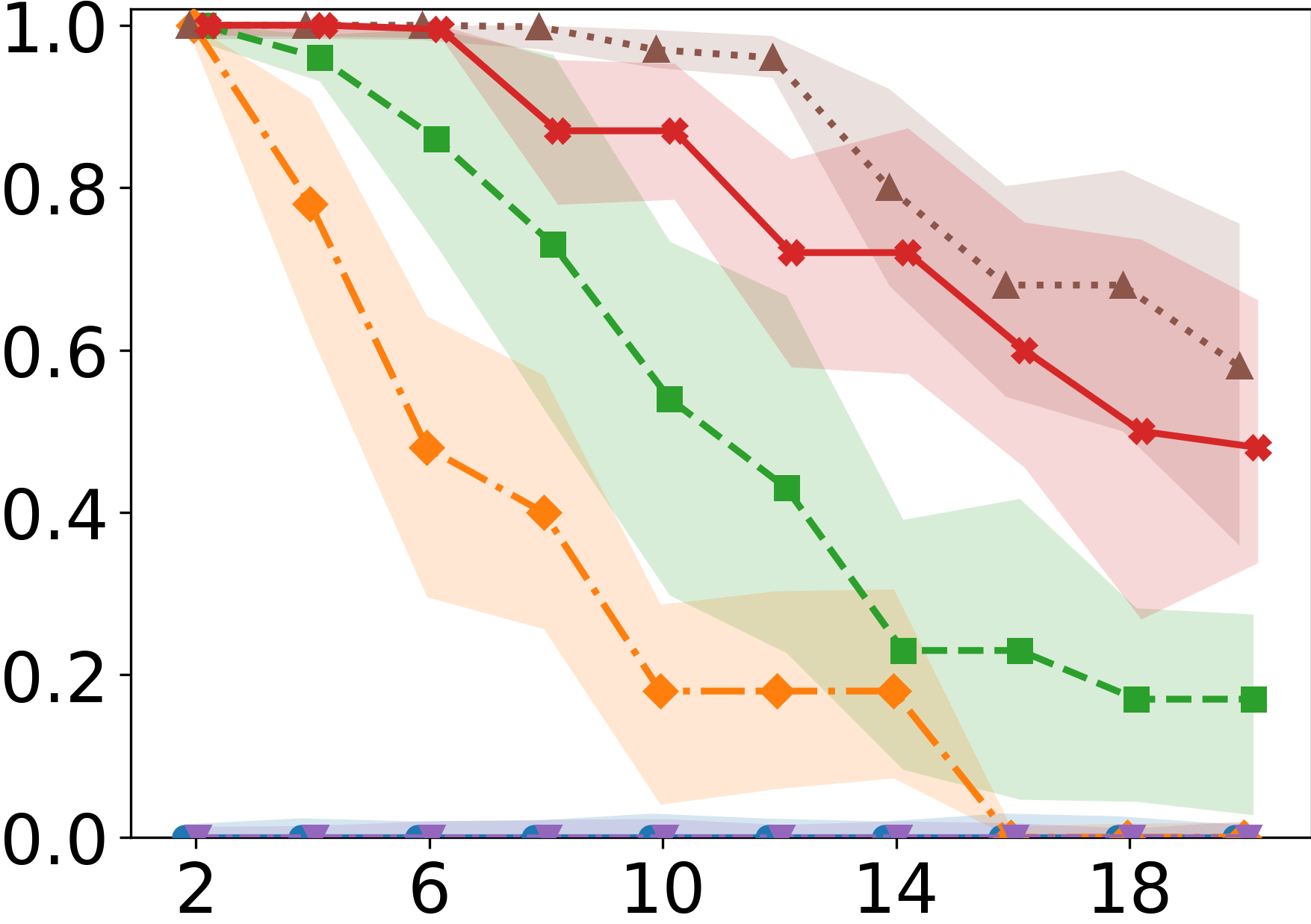}
  \caption{}\label{fig:k1_r3a}
\end{subfigure}\hspace{0.1\textwidth}
\begin{subfigure}[t]{0.38\textwidth}
  \centering
  \includegraphics[width=\linewidth]{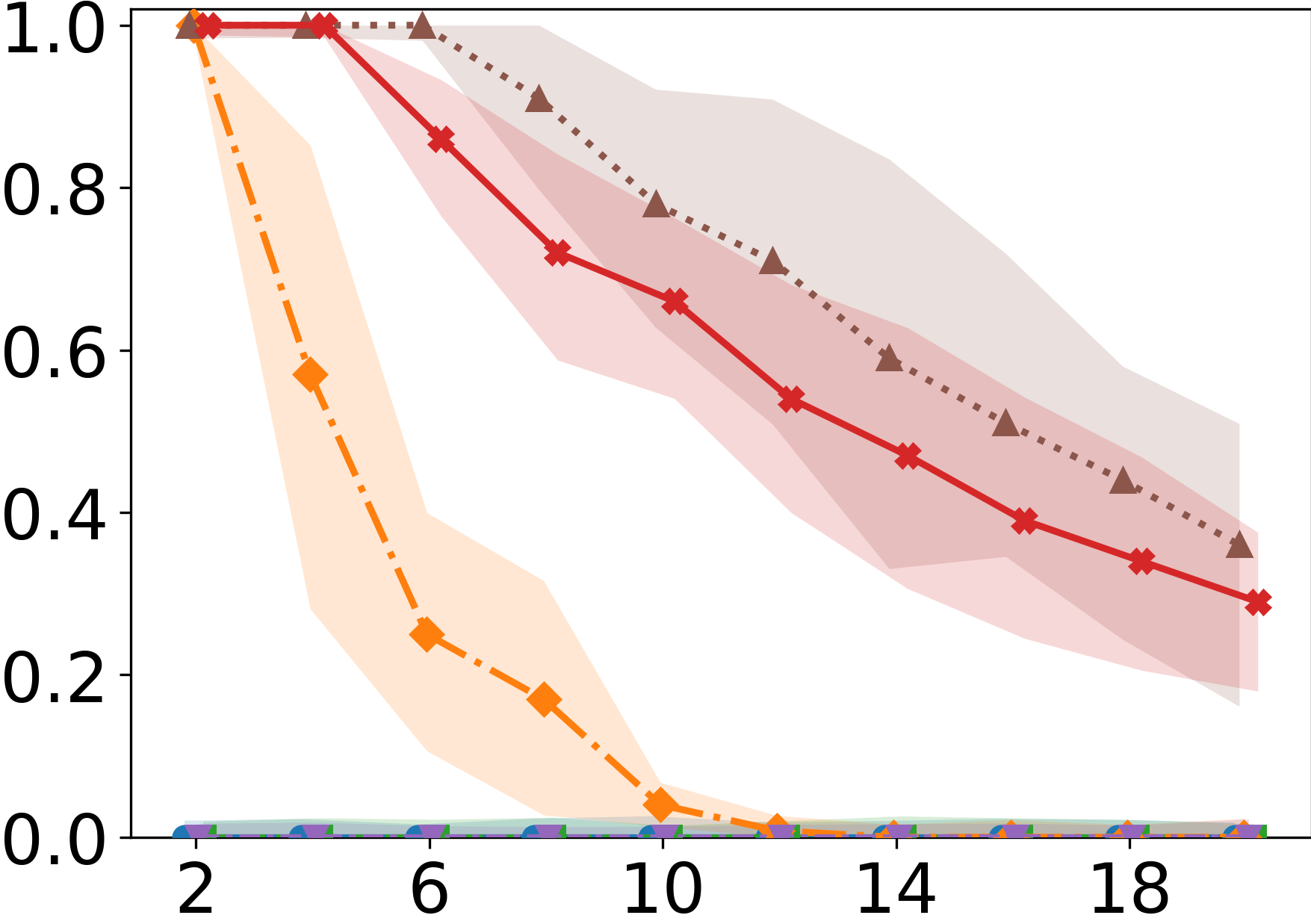}
  \caption{}\label{fig:k1_r3b}
\end{subfigure}

\setcounter{subfigure}{0}
\par\medskip
{\centering\small\textbf{(iv) \textsc{Car2D} $k$-Reachability + obstacles}\par}
\vspace{0.2em}

\begin{subfigure}[t]{0.38\textwidth}
  \centering
  \includegraphics[width=\linewidth]{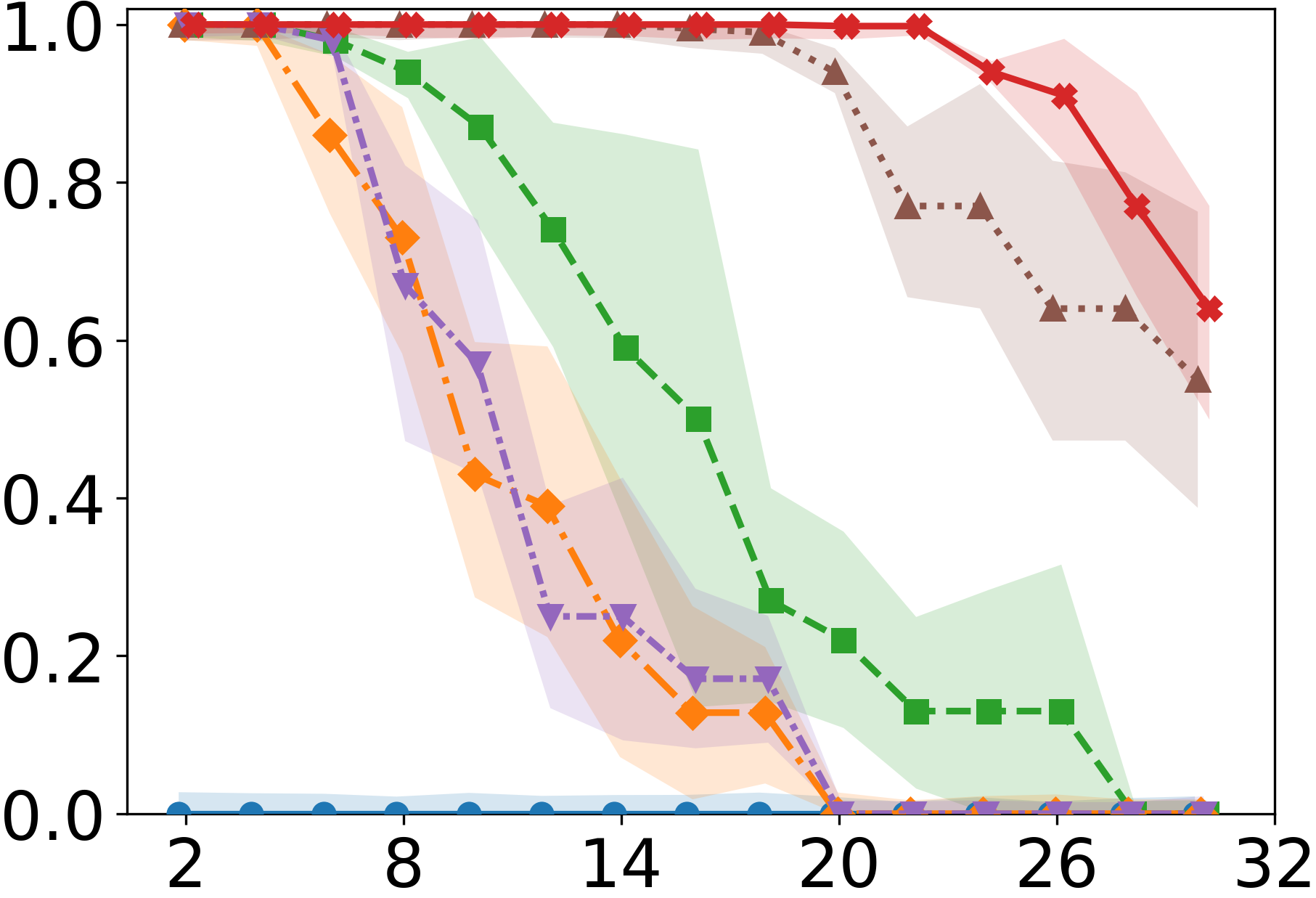}
  \caption{}\label{fig:k1_r4a}
\end{subfigure}\hspace{0.1\textwidth}
\begin{subfigure}[t]{0.38\textwidth}
  \centering
  \includegraphics[width=\linewidth]{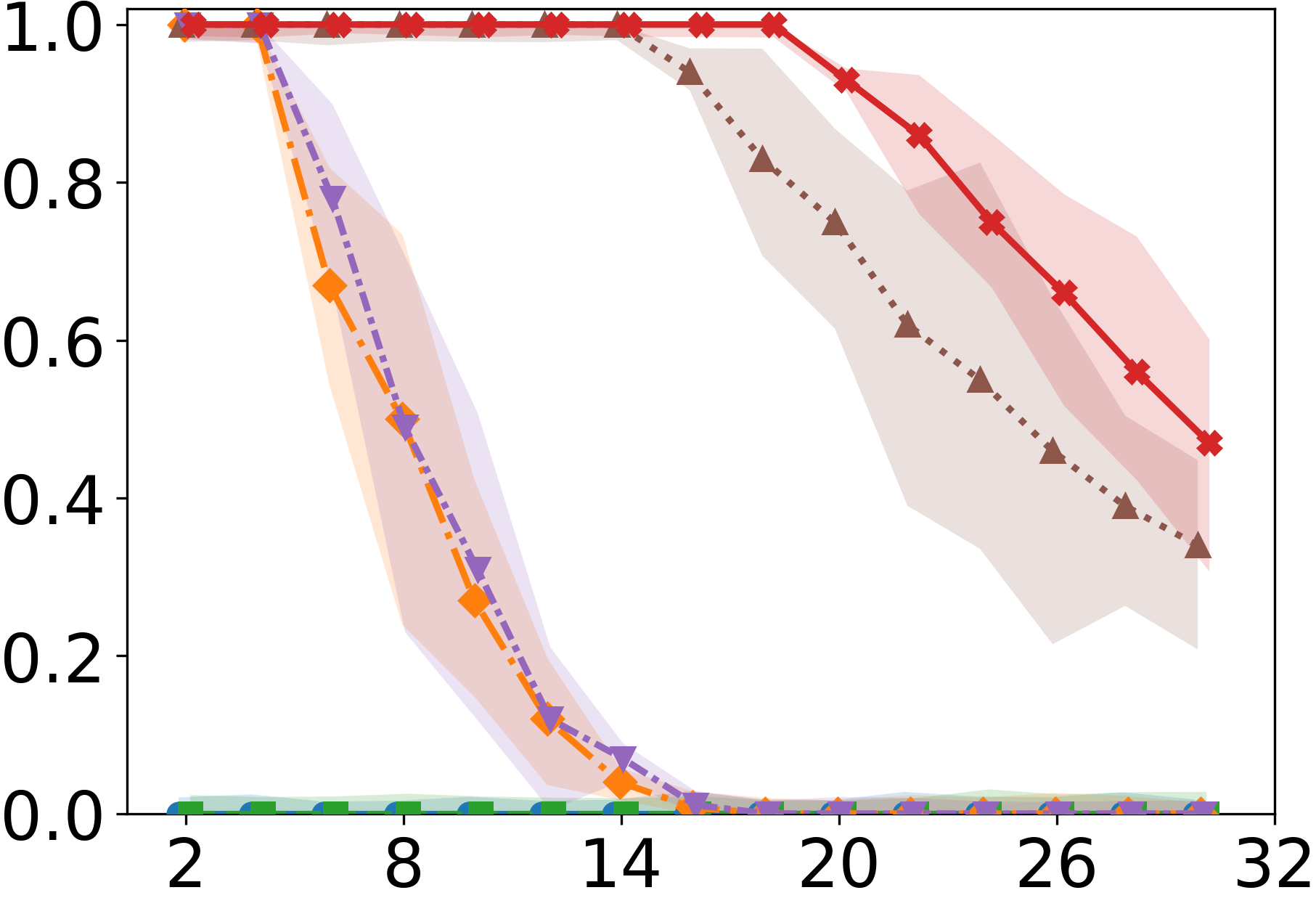}
  \caption{}\label{fig:k1_r4b}
\end{subfigure}

\caption{\textbf{Scalability plots for $k=1$ reachability.}
Each row corresponds to one environment on $k$-reachability tasks.
\textbf{Column (a)} shows the successful training ratio ($y$-axis) versus the number of training tasks $|\Train|$ ($x$-axis) for $k=1$.
\textbf{Column (b)} shows the zero-shot generalization ratio ($y$-axis) versus the number of training tasks $|\Train|$ ($x$-axis) for $k=1$.
We report mean ($\pm$ standard deviation) performance across \textbf{ten} random seeds.}
\label{fig:successvstasks_k1_all}
\end{figure*}

We report the remaining $k$-reachability experiments for $k=4$ down to $k=1$, corresponding to decreasing horizon lengths, across \textsc{Car2D}, \textsc{SimpleDrone}, \textsc{DroneAttitude}, and \textsc{Car2D} with obstacles. These results are shown in Figures~\ref{fig:successvstasks_k4_all}, \ref{fig:successvstasks_k3_all}, \ref{fig:successvstasks_k2_all}, and \ref{fig:successvstasks_k1_all}. We also include the \textsc{Car2D} obstacle experiment for $k=5$ in Figure~\ref{fig:successvstasks_obstacles}.

Across these settings, the same overall trend holds: \drill~maintains higher training scalability and stronger zero-shot generalization as the task horizon varies, while the baselines degrade more rapidly as the number of training or unseen tasks increases. The exception is the $k=1$ setting, where the task is sufficiently short-horizon and does not require long-horizon compositional reasoning. In this easier regime, \textsc{VariBAD} also performs well and slightly outperforms \drill.

\subsection{Additional Reacher results}

\begin{figure}[t]
  \centering
  Legend:
{\small
{\color{arsblue}\rule{0.9em}{0.9em}} ARS,
{\color{bcgreen}\rule{0.9em}{0.9em}} BC,
{\color{mamlorange}\rule{0.9em}{0.9em}} MAML,
{\color{varibadbrown}\rule{0.9em}{0.9em}} VariBAD,
{\color{genrlpurple}\rule{0.9em}{0.9em}} GenRL, and
{\color{dipsred}\rule{0.9em}{0.9em}} \drill~(Ours).}
  \includegraphics[width=\linewidth]{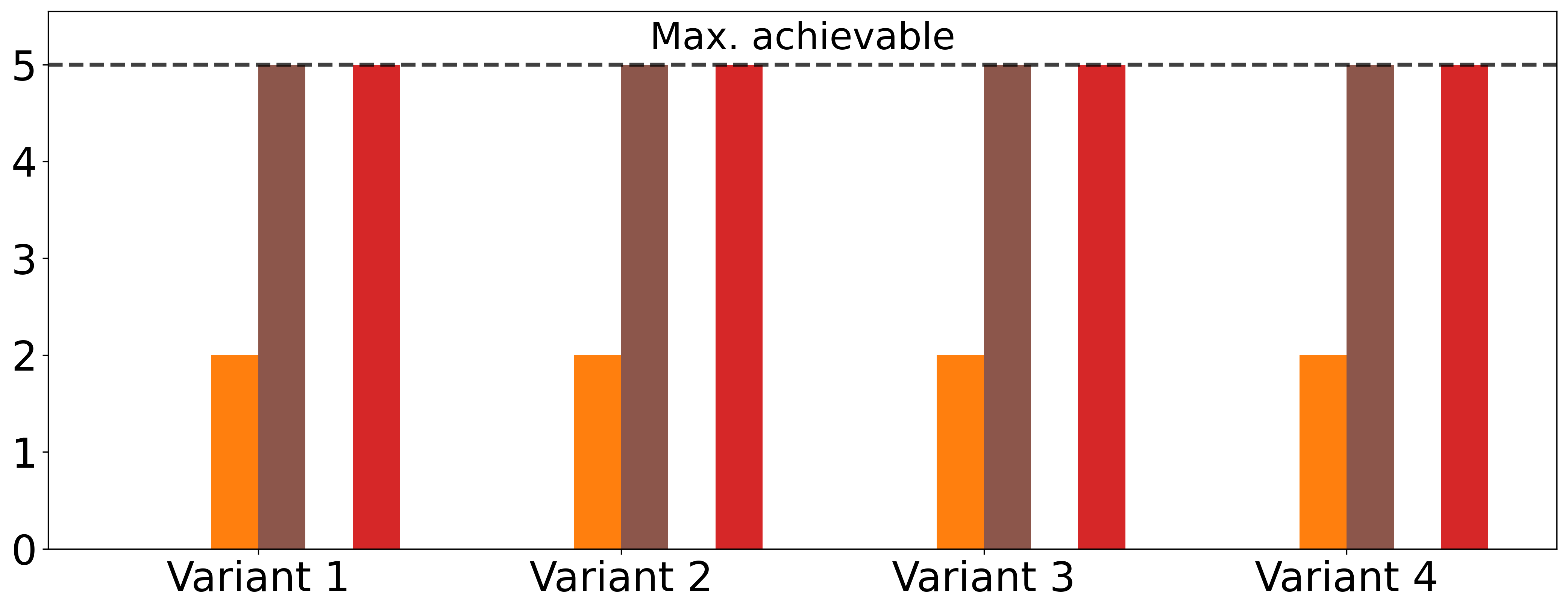}
  \caption{\textsc{Reacher}: 9 blocks}
  \label{fig:reacher_9_blocks}
\end{figure}

Figure~\ref{fig:reacher_9_blocks} reports the \textsc{Reacher} experiment with 9 blocks, complementing the 8- and 10-block settings shown in the main paper. The quantitative trends remain consistent: \drill~scales with the number of training indices and maintains strong zero-shot generalization to unseen configurations, while baselines either degrade in training scalability or fail to generalize as the environment difficulty increases.

\subsection{Choice benchmark results}

\begin{figure*}[t]
\centering

Legend:
{\small {\color{arsblue}\rule[0.6ex]{1.2em}{1pt}$\,\bullet$} ARS,
{\color{bcgreen}\rule[0.6ex]{1.2em}{1pt}$\,\blacksquare$} BC,
{\color{mamlorange}\rule[0.6ex]{1.2em}{1pt}$\,\blacklozenge$} MAML,
{\color{varibadbrown}\rule[0.6ex]{1.2em}{1pt}$\,\blacktriangle$} VariBAD,
{\color{genrlpurple}\rule[0.6ex]{1.2em}{1pt}$\,\blacktriangledown$} GenRL, and
{\color{dipsred}\rule[0.6ex]{1.2em}{1pt}{\large $\times$}} \drill~(Ours).}

\setlength{\abovecaptionskip}{2pt}
\setlength{\belowcaptionskip}{2pt}

\setcounter{subfigure}{0}
{\centering\small\textbf{(i) \textsc{Car2D} \textsc{Choice} benchmark, 1-level branching}\par}
\vspace{0.2em}

\begin{subfigure}[t]{0.33\textwidth}
  \centering
  \includegraphics[width=\linewidth]{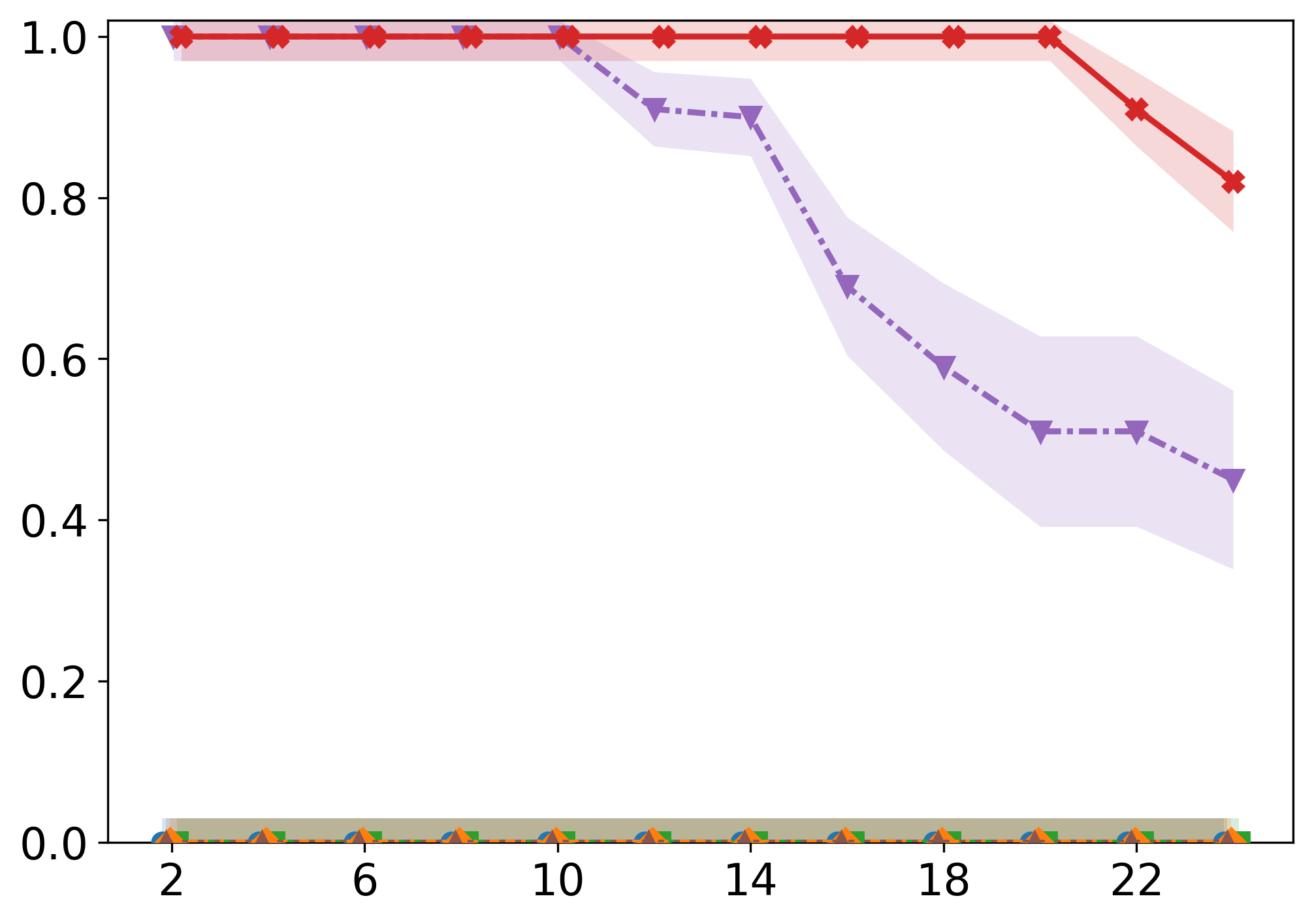}
  \caption{}\label{fig:choice1a}
\end{subfigure}
\begin{subfigure}[t]{0.33\textwidth}
  \centering
  \includegraphics[width=\linewidth]{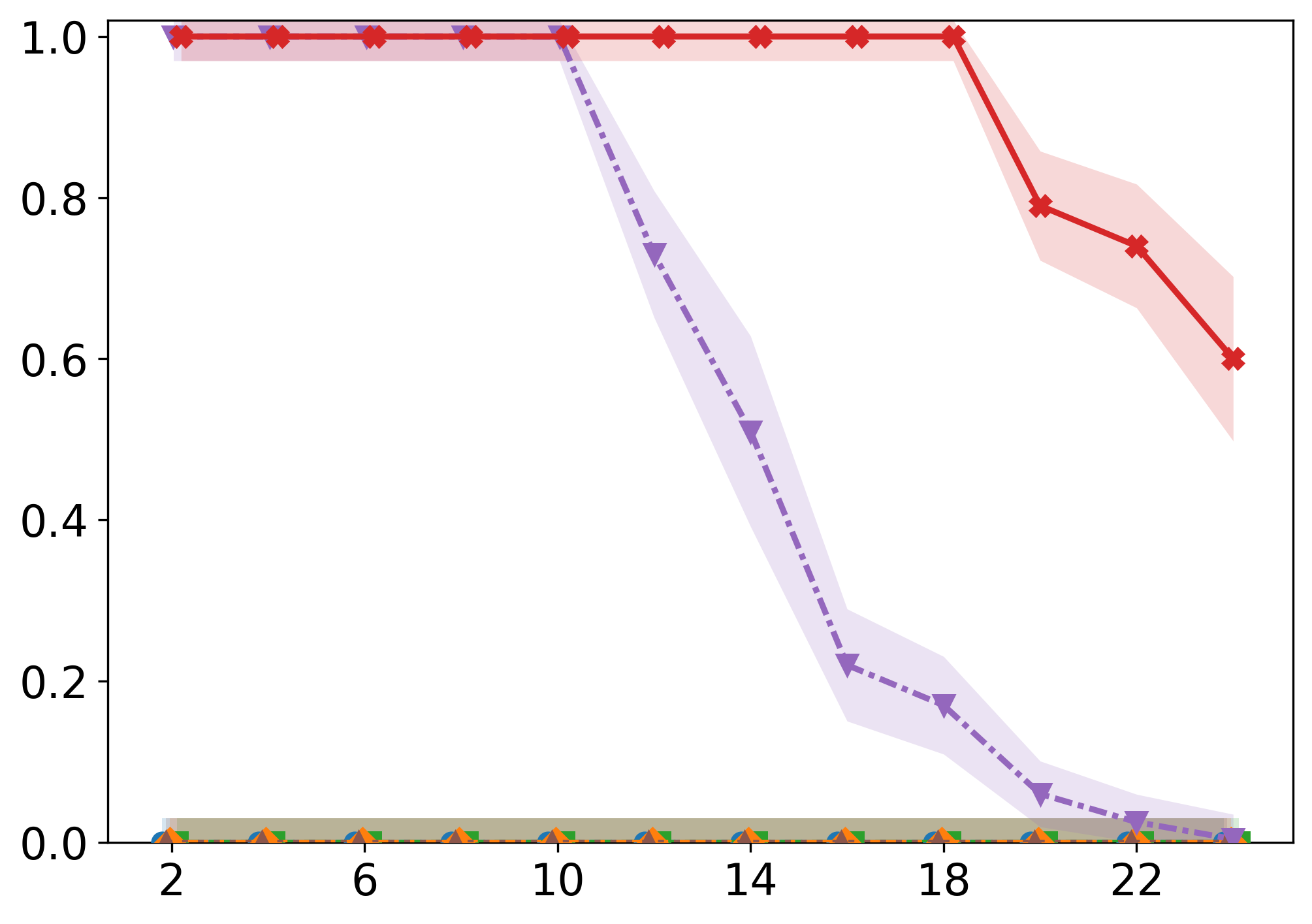}
  \caption{}\label{fig:choice1b}
\end{subfigure}
\begin{subfigure}[t]{0.32\textwidth}
  \centering
  \includegraphics[width=\linewidth]{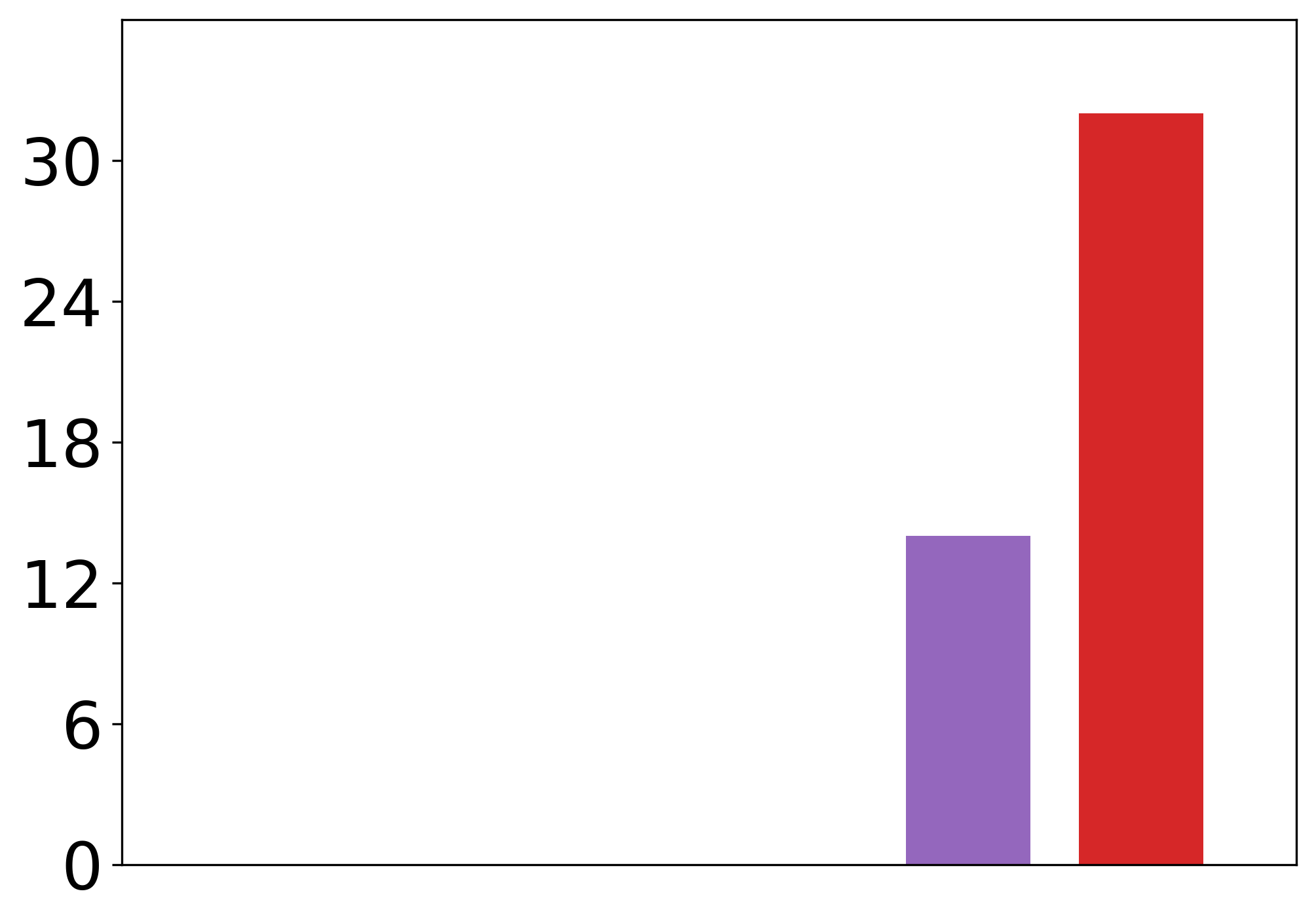}
  \caption{}\label{fig:choice1c}
\end{subfigure}

\vspace{0.6em}

\setcounter{subfigure}{0}
{\centering\small\textbf{(ii) \textsc{Car2D} \textsc{Choice} benchmark, 2-level branching}\par}
\vspace{0.2em}

\begin{subfigure}[t]{0.33\textwidth}
  \centering
  \includegraphics[width=\linewidth]{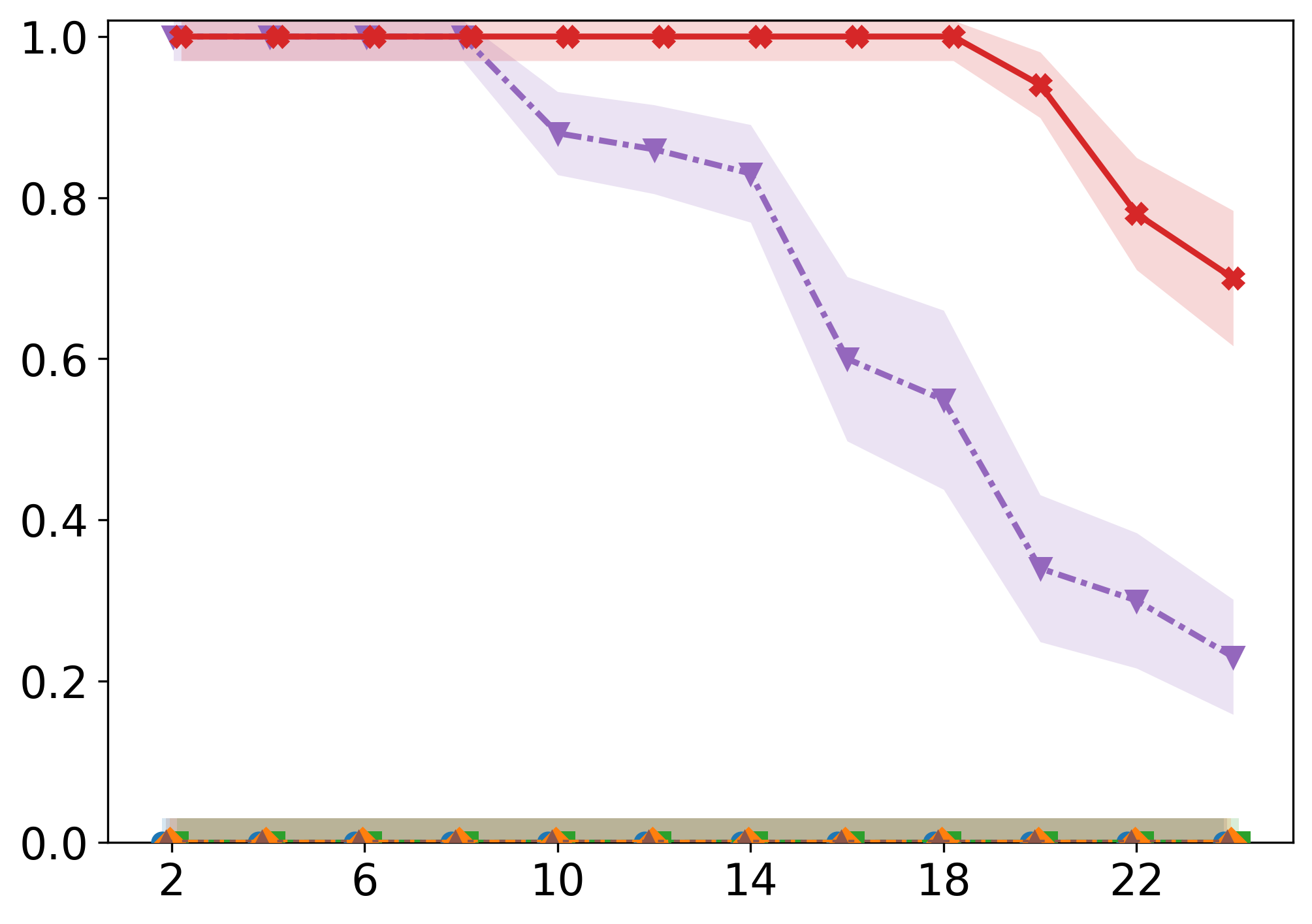}
  \caption{}\label{fig:choice2a}
\end{subfigure}
\begin{subfigure}[t]{0.33\textwidth}
  \centering
  \includegraphics[width=\linewidth]{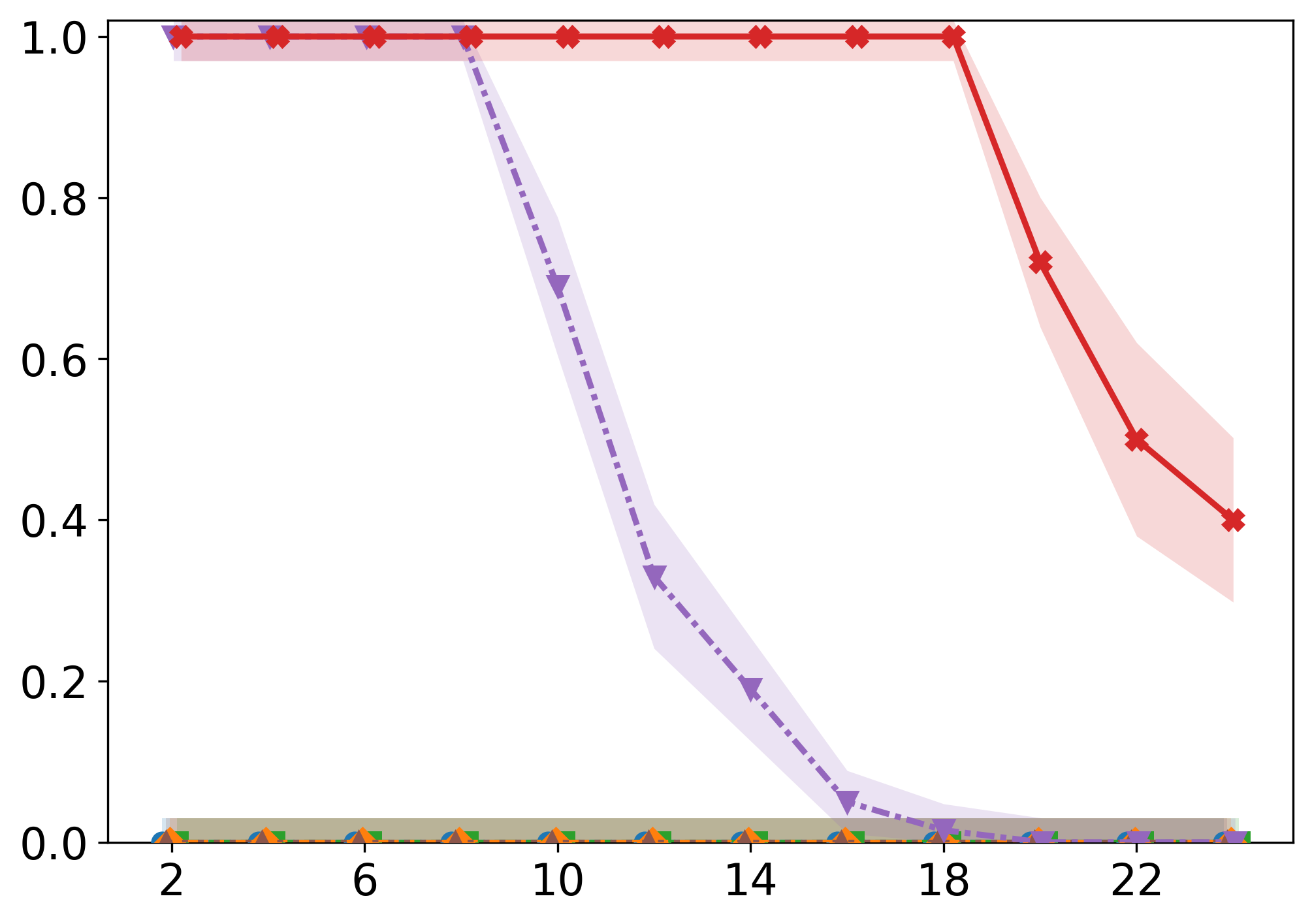}
  \caption{}\label{fig:choice2b}
\end{subfigure}
\begin{subfigure}[t]{0.32\textwidth}
  \centering
  \includegraphics[width=\linewidth]{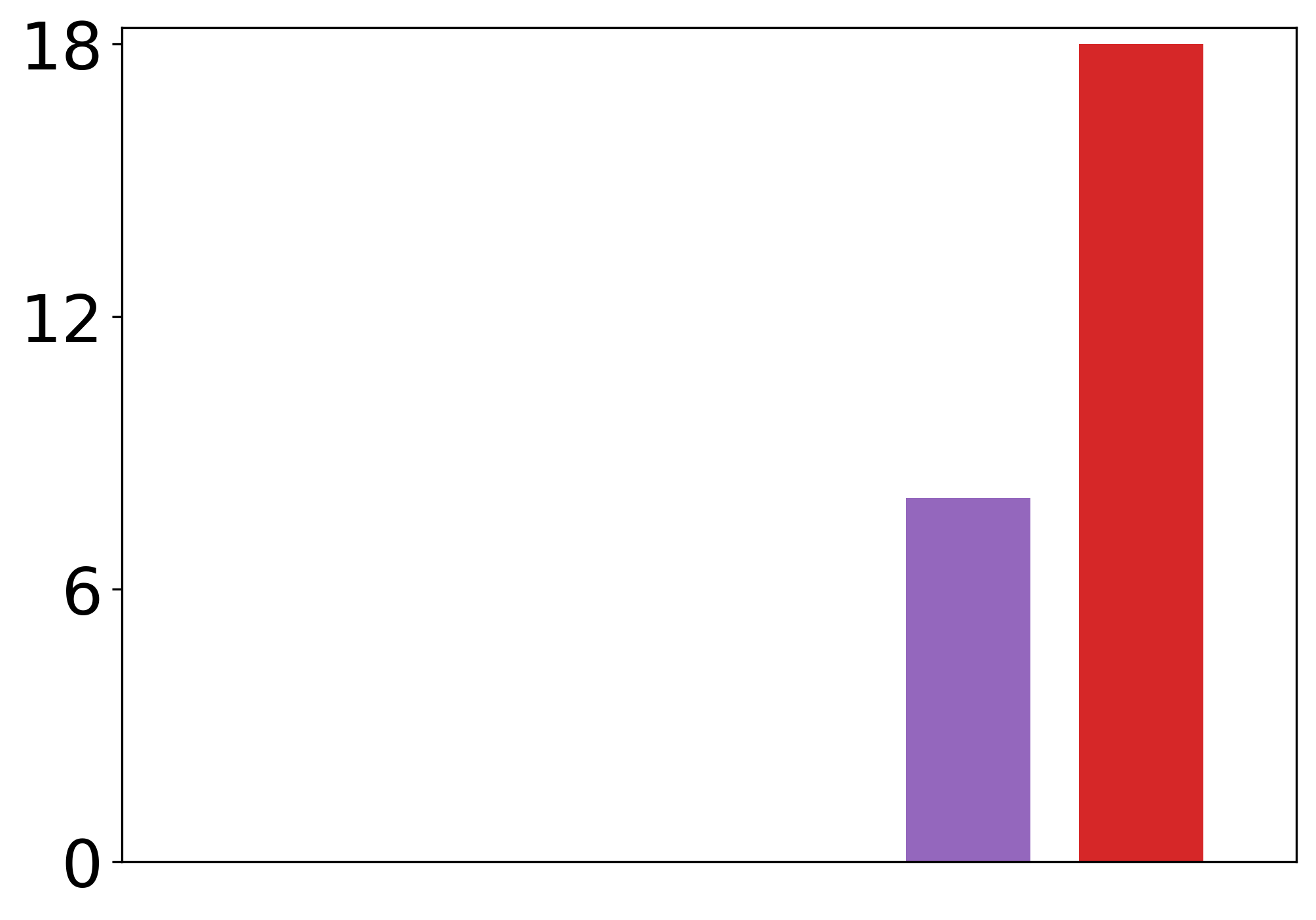}
  \caption{}\label{fig:choice2c}
\end{subfigure}

\caption{\textbf{Scalability plots for the \textsc{Choice} benchmark.}
Each row corresponds to one \textsc{Car2D} \textsc{Choice} benchmark variant. 
\textbf{Column (a)} shows the successful training ratio ($y$-axis) versus the number of training tasks $|\Train|$ ($x$-axis). 
\textbf{Column (b)} shows the zero-shot generalization ratio ($y$-axis) versus the number of training tasks $|\Train|$ ($x$-axis). 
\textbf{Column (c)} shows absolute zero-shot generalization ($y$-axis). 
We report mean ($\pm$ standard deviation) performance across \textbf{ten} random seeds.}

\label{fig:successvstasks_choice}
\end{figure*}

We also include the \textsc{Choice} experiments for both one-level and two-level branching tasks in Figure~\ref{fig:successvstasks_choice}. In these experiments, only \genrl~and \drill~achieve non-trivial performance, since the other baselines do not have an explicit branching mechanism and therefore struggle to solve the structured choice problem. Across both branching levels, \drill~consistently outperforms \genrl, showing that the decoupled imitation-based template learning procedure is more effective for handling branching task structure and generalizing to unseen choice configurations.

\section{Additional ablations}

\subsubsection{Teacher-driven subtask discovery enables branching around obstacles.}
\label{sec:supp:ablation_subtasks}

This ablation studies an obstacle setup in \textsc{Car2D} that violates the
single-template inductive bias of GenRL. Consider a family where the start region shifts
monotonically along the $x$-axis while the goal lies above a central obstacle. For start
indices on the left of the obstacle, the correct reach-avoid strategy routes around the
\emph{left} side; once the start crosses near the obstacle center, the correct strategy
switches to routing around the \emph{right} side. While the specification template is
identical across indices (only the initial distribution is translated), the \emph{policies}
are not smoothly related across all indices, i.e., a contiguous set of indices must follow one
side and the remaining indices must follow the other. GenRL, which fits a single template,
regularizes toward one consistent mode and often collapses to a single route (e.g., always
going left), failing to represent the required switch.

\paragraph{Branched template fitting.}
Our decoupled method enables a simple preprocessing step that leverages teacher behavior
before template fitting. In this ablation, we first train near-optimal per-task teachers
without cross-index regularization (i.e., we do not constrain teachers for different
indices to have similar weights or shared evolution). Because each teacher is optimized
only for its own task instance, the resulting rollouts naturally realize both routing
modes (left-of-obstacle and right-of-obstacle) when those modes are required. We then use
these rollouts as a preprocessing signal for \emph{subtask discovery}. We cluster teacher
trajectories (and thus their corresponding indices) into two groups, yielding a partition
$\Train=\Train_{\mathrm{left}}\cup \Train_{\mathrm{right}}$.

Finally, we fit two separate templates, $\kappa^{\mathrm{left}}$ and
$\kappa^{\mathrm{right}}$, each using teacher-labeled datasets restricted to its cluster.
At inference time, we unroll the corresponding template in index space (as in
Section~\ref{sec:supp:index_unrolling}) to generate policies within that subfamily. Since each
template only needs to model a single coherent routing mode, template fitting becomes
easier and more accurate, and the resulting branched generator recovers the mode switch
around the obstacle.

\paragraph{Result.}
Compared to fitting a single template over all indices, the branched variant avoids mode
collapse and correctly switches routes around the obstacle as the start crosses the
obstacle center, improving specification satisfaction in this setting.

\end{document}